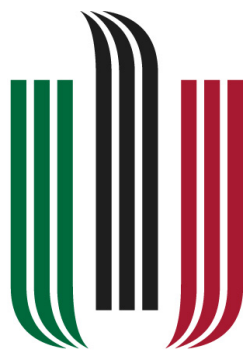

**PRACA DYPLOMOWA MAGISTERSKA**

## *Data adaptation in HANDY economy-ideology model*

*Adaptacja danych w modelu ekonomiczno-ideologicznym HANDY*

Autor:               *Marcin Sendera*
Kierunek studiów:    *Informatyka*
Opiekun pracy:        *prof. dr hab. inż. Witold Dzwinel*

Kraków, 2019

# Declaration

Uprzedzony o odpowiedzialności karnej na podstawie art. 115 ust. 1 i 2 ustawy z dnia 4 lutego 1994 r. o prawie autorskim i prawach pokrewnych (t.j. Dz.U. z 2006 r. Nr 90, poz. 631 z późn. zm.): „Kto przywłaszcza sobie autorstwo albo wprowadza w błąd co do autorstwa całości lub części cudzego utworu albo artystycznego wykonania, podlega grzywnie, karze ograniczenia wolności albo pozbawienia wolności do lat 3. Tej samej karze podlega, kto rozpowszechnia bez podania nazwiska lub pseudonimu twórcy cudzy utwór w wersji oryginalnej albo w postaci opracowania, artystycznego wykonania albo publicznie zniekształca taki utwór, artystyczne wykonanie, fonogram, wideogram lub nadanie.", a także uprzedzony o odpowiedzialności dyscyplinarnej na podstawie art. 211 ust. 1 ustawy z dnia 27 lipca 2005 r. Prawo o szkolnictwie wyższym (t.j. Dz. U. z 2012 r. poz. 572, z późn. zm.): „Za naruszenie przepisów obowiązujących w uczelni oraz za czyny uchybiające godności studenta student ponosi odpowiedzialność dyscyplinarną przed komisją dyscyplinarną albo przed sądem koleżeńskim samorządu studenckiego, zwanym dalej «sądem koleżeńskim».", oświadczam, że niniejszą pracę dyplomową wykonałem(-am) osobiście i samodzielnie i że nie korzystałem(-am) ze źródeł innych niż wymienione w pracy.

......................................................

# Acknowledgements

First of all, I would like to thank my supervisor Prof. dr hab. inż. Witold Dzwinel. He helped me whenever I had a question about my research and steered me and my thesis in the right direction. I would also like to thank Mr. Adrian Kłusek who as an expert in Supermodeling was involved in the workflow and help me with going trough this methodology successfully.

Moreover, I have to acknowledge that this work has been supported by National Science Centre, Poland grant no. 2016/21/B/ST6/01539. In addition, this research was supported in part by PLGrid Infrastructure. I am very grateful to both the agency and the infrastructure for enabling the demanding computations.

Finally, I want to express my sincere gratitude to people without whom this thesis might not have been written. To my parents and members of the family for their endless motivating me to finish my research and writing. To my girlfriend for her involvement and being so supportive. And last but not least, to my friends for showing me that I can always rely on them.

# Abstract

Modelowanie matematyczne jest podejściem szeroko stosowanym w prawie wszystkich dziedzinach współczesnej nauki i inżynierii. Ze względu na potrzebę dokładnego przewidywania zjawisk przyrodniczych, coraz to bardziej skomplikowane modele matematyczne ich opisu są opracowywane. Lecz pominąwszy coraz większą zdolność do przewidywania, pojawia się nieunikniona trudność właściwego dopasowania zebranych danych do wielowymiarowych i wysoce nieliniowych modeli. W celu rozwiązania tego wymagającego problemu, powstała cała, odrębna dyscyplina naukowa - metoda asymilacji danych (*data assimilation*). Wykorzystując model rozwoju społeczno-ekonomicznego HANDY (Human and Nature Dynamics), zaprezentowano w pracy szczegółowe porównanie dwóch metod asymilacji danych: *Approximate Bayesian Computation* (klasycznej metody asymilacji) oraz nowatorskiego podejścia znanego jako *Supermodeling*. Co więcej, na podstawie wykonanej analizy wrażliwości parametrów modelu (*Sensitivity Analysis*), zaproponowano nową technikę redukcji współczynników łączących podmodele (coupling coefficients), która znacząco przyspieszyła zbieżność Supermodelu. Zaprezentowano, że użycie metody ABC z pełną wiedzą o czułości jej parametrów może skutkować zadowalającym przybliżeniem ich początkowych wartości. Jednak - co także udało się przedstawić - powyższe podejście nie prowadzi do lepszych wyników, niż te, które można by uzyskać ze zwyczajnego zastosowania metody ABC. Udowodniono, że synchronizując Supermodel poprzez łączenie ze sobą najbardziej czułych zmiennych dynamicznych, można uzyskać lepsze przewidywanie zachowania zarówno systemów chaotycznych, jak i tych o bardziej stabilnej charakterystyce, niż było to możliwe za pomocą metody ABC. W niniejszej pracy zaproponowano odpowiednie metodologie, mogące przyczynić się do poprawy przewidywań badanego modelu, uwzględniające jego zachowania i charakterystykę.

# Abstract

The concept of *mathematical modeling* is widespread across almost all of the fields of contemporary science and engineering. Because of the existing necessity of predictions the behavior of natural phenomena, the researchers develop more and more complex models. However, despite their ability to better forecasting, the problem of an appropriate fitting ground truth data to those, high-dimensional and nonlinear models seems to be inevitable. In order to deal with this demanding problem, the entire discipline of *data assimilation* has been developed. Basing on the *Human and Nature Dynamics* (HANDY) model, we have presented a detailed and comprehensive comparison of *Approximate Bayesian Computation* (classic data assimilation method) and a novelty approach of *Supermodeling*. Furthermore, with the usage of *Sensitivity Analysis*, we have proposed the methodology to reduce the number of coupling coefficient between submodels and as a consequence to increase the speed of the Supermodel converging. In addition, we have demonstrated that usage of Approximate Bayesian Computation method with the knowledge about parameters' sensitivity could result with satisfactory estimation of the initial parameters. However, we have also presented the mentioned methodology as unable to achieve similar predictions to Approximate Bayesian Computation. Finally, we have proved that Supermodeling with synchronization via the most sensitive variable could effect with the better forecasting for chaotic as well as more stable systems than the Approximate Bayesian Computation. What is more, we have proposed the adequate methodologies.

# Table of contents

# List of figures

# List of tables

# Glossary of Symbols & Abbreviations

**Acronyms / Abbreviations**

ABC   Approximate Bayesian Computation

GDP   Gross Domestic Product

GT      Ground Truth Model

HANDY  Human And Nature DYnamics model

MCMC  Markov chain Monte Carlo

ODE   Ordinary Differential Equation

PCA   Principal Components Analysis

PDE   Partial Differential Equation

RMSE  Root-Mean-Square Error

SA      Sensitivity Analysis

SMC   Sequential Monte Carlo

SUMO  Supermodeling

**Handy Model's Parameters**

$\alpha_M$      Famine (maximum) death rate

$\alpha_m$      Normal (minimum) death rate

$\beta_C$      Commoners birth rate

$\beta_E$      Elites birth rate

$\delta$        Depletion (production) factor

$\gamma$        Regeneration rate of nature

$\kappa$        Inequality factor

$\lambda$        Nature carrying capacity

$\rho$        Threshold wealth per capita

$s$        Subsistence salary per capita

$w$        Accumulated wealth

$x_C$        Commoners population

$x_E$        Elites population

$y$        Nature

**Mathematical Symbols**

$\binom{n}{k}$        Binomial coefficient

$\boldsymbol{A}$        Matrix A

$\boldsymbol{A}_{\boldsymbol{B}}^{(i)}$        Matrix that have all columns from $\boldsymbol{A}$, but $i$-th column from $\boldsymbol{B}$

$\boldsymbol{A}_{\sim i}$        Matrix A without $i$-th factor

$\gamma(x, y)$        Joint distribution

$\mathbb{P}$        Probability distribution

$\mathscr{N}$        Gaussian distribution

$\bar{x}$        Mean value for x

$\Pi(\mathbb{P}_r, \mathbb{P}_g)$        Set of all joint distributions $\gamma(x, y)$ with marginals $\mathbb{P}_r$ and $\mathbb{P}_g$

$\mathbf{C} = \{C_{\mu\nu}^i\}$        Tensor $\mathbf{C}$ composed of coupling nonnegative coefficients $C_{\mu\nu}^i$ between models $\mu$ and $\nu$ for $ith$ variable $x^i$

$d(\bullet, \bullet)$        Metric $d$

$E_X(\cdot)$        Mean taken over $X$

$K_t$    Perturbation kernel

$p(\theta)$    Prior distribution

$S_i$    First order sensitivity coefficient

$S_{Ti}$    First order sensitivity coefficient

$Tr(\boldsymbol{A})$    Trace of matrix $\boldsymbol{A}$

$V_X(\cdot)$    Variance taken over $X$

$x_{std}$    Standard deviation for x

**Other Symbols**

$\mu$    Complexity measure

$\overline{f2w_{std}}$    Standard deviation for both ways forecasting error, $f2w$

$\overline{f2w}$    Mean both ways forecasting error, $f2w$

$\overline{fb}$    Mean backward forecasting error, $fb$

$\overline{ff_{std}}$    Standard deviation for forward forecasting error, $f2w$

$\overline{ff}$    Mean forward forecasting error, $ff$

$\overline{t_{norm}}$    Mean value of normalized learning time

$\overline{t_{sumo}}$    Mean value of Supermodel's learning time

$\bar{t}$    Mean learning time, $t$

$f$    Frequency

$f2w$    Both ways (backward and forward) forecasting error

$fb$    Backward forecasting error

$ff$    Forward forecasting error

$H_p$    The number of HANDY's initial parameters

$S_{cc}$    The number of Supermodel's coupling coefficients

$t$    Learning time

$t_{norm}$     Normalized learning time

$t_{std}$     Standard deviation for learning time, $t$

$t_{sumo}$     Supermodel's learning time

# Chapter 1

# Introduction

## 1.1 Preface

During the nineteenth century, the natural sciences undeniably accomplished the greatest development in their history. From observations of natural phenomena to develop mathematical methods and construct more and more sophisticated experiments [50].

The most influential field of knowledge had seemed to be classical mechanics that was build upon the concept of *determinism* - which was a conviction that the knowledge of current state of a mechanical systems completely determines its future as shown in Reich & Cotter [50].

Formalisation of the mentioned concept was presented for the first time by Pierre Simon Laplace. The French mathematician proposed an intellect (now known as *Laplace's demon*) which has three properties: exact knowledge of the laws of nature; complete knowledge of the state of the universe at a particular point in time; the ability to solve any form of mathematical equation exactly.

Obviously, despite rare cases, the mentioned properties would not appear in practice, because mathematical models are simplified version of the phenomenon. Moreover, each real-world measurement are limited and provides measurement errors and at the end, the solutions of mathematical models are most often approximated and not exact.

Taking under consideration the unquestionable results of physicist in the classical mechanics, we should not be surprised that *scientism* (that came from determinism) in a century became the most popular ideology in the amongst scientists and elites. With the introduction of contemporary physics - quantum mechanics and theory of relativity, the change of the paradigm became a fact.

## 1.2    Ensemble prediction & data assimilation

Due to all limitations that were mentioned, one could find mathematical modeling pointless. However, even not so exact computational models are really useful in many very demanding problems like weather predictions [58]. Combination of predictions based on complex computational models and analysis of outputs compared with observations, make the usage possible [50].

The mentioned combination may be understand as extrapolating limited knowledge of the present state of the model (ground truth data) to the future (forecasting). It is easy to observe that both part of the process: extrapolation and model adjustment (based on data) are sources of the inevitable errors - forecast uncertainties. Due to the necessity of limit these uncertainties, two approaches - *ensemble prediction* and *data assimilation* were proposed within the scientific community as shown in [50].

Ensemble prediction could be seen in a way of represent forecast uncertainties as random variables. In a fact, as a result of ensemble prediction the wide sets of forecast are created. Each of them is viewed as connected with specific random variable.

On the other hand, data assimilation is used as a common term for vary methodologies that allow to improve forecasting by using both ground truth data and artificially generated forecasts. These algorithms are combining large observations sets with computational models, so it reduce errors of both forecasting and model parameters. The main aim of data assimilation approach is to estimate forecast much better than by using only ground truth data or computational models. Due to the possible pros of data assimilation, the wide applications in different fields (from biology to engineering) should not be unpredictable.

Across variety of possible data assimilation approaches, the Bayesian approach which identify it as a Bayesian estimation problem [4]. It should be mentioned not only because of its popularity but also due to the fact that algorithms based on Bayes theorem will be used in this thesis. Moreover, data assimilation algorithms are needed for high-dimensional and nonlinear computational models, especially in geoscience applications. For such purposes, classical smoothing and filtering algorithms are not sufficient [50, 75].

## 1.3    Problem statement

The main reasons for using the data assimilation are high-dimensionality and nonlinearity of the computational models. Moreover, in the real-world applications the small amount of data with uncertainties leads to the necessity of usage the data assimilation methodologies. However, most of the applications (e.g. weather forecasting [58] or tumor growth prediction [20, 19, 47]) are based on complex chaotic systems, which further behaviour has to be forecast with a high accuracy in a small amount of time.

On the other hand, the commonly used methods could either have a tendency to stay in the local minima or it takes a lot of computational time to receive good enough results. Because of the above, it seems to be reasonable to try lower the problem dimensionality with respect to improve the computational time that is necessary to get satisfactory forecasting uncertainties. These implications lead to formulate purposes of the thesis.

This work has two aims:

1. Human and Nature Dynamics (HANDY) model [43] analysis and an attempt to the simplification of it's complexity.

2. However, the main aim of this thesis is to study the capabilities of the data assimilation methods used for such nontrivial model. Particularly interesting is the comparison between novelty concept of the Supermodeling [58, 75] and the Approximate Bayesian Computation [6, 64] - classic method of the data assimilation.

The answer for the second aim of this thesis could have strong impact on the science community in each field that is connected to computational models. Besides geoscience and climatology [58, 75], the mathematical modelling is widely used in any fields of biological [76, 21] and environmental sciences [43] as well as engineering or physics. The positive answer for any of these problems could in fact improve forecasting or time that is needed to do it correctly. What is more, if the results would shown the cases when the Supermodeling is noticeably better than the classical methods, it would push this methodology from the community of climate scentists into classical data assimilation and machine learning fields of interests.

## 1.4   Thesis structure

The structure of this thesis is prepared to appropriate lead a reader for three main parts: from theoretical foundations by specific application of Sensitivity Analysis, Approximate Bayesian Computation and Supermodeling to conclusions of the main topic of this thesis.

The following chapters contain:

• **Chapter 2** - Introduction to computational model - Human and Nature Dynamics (HANDY) that will be studied.

• **Chapter 3** - Theoretical foundations of the classical data assimilation algorithm - Approximate Bayesian Computation (ABC) with the examples of real-world usage.

- **Chapter 4** - Theoretical foundations of the novelty methodology from climatology and chaotic systems community - Supermodeling (SUMO) with the examples of real-world usage.

- **Chapter 5** - Overview of the sensitivity analysis method to find the most sensitive parameters.

- **Chapter 6** - Proposed methodologies with results and their improvements that could answer to the main questions of this thesis.

- **Chapter 7** - Quick review of the used computational methods.

- **Chapter 8** - Conclusions and overview of the methodology.

- **References** - References that were necessary during writing this thesis.

- **Appendix A** - Prediction results at each step of the process of development the proposed methodology.

# Chapter 2

# Human and nature dynamics model (HANDY)

Lots of processes and phenomena which are well-known from the observations of the nature are much more complicated when it comes to describe. In such cases, people most often use the methods known as a *mathematical modelling*. The choice appears to be natural or fundamental not only because of the very precise language of mathematics but also due to the ability of numerical calculations. Obviously, most often models are not creating to be an intellectual toy but to predict meaningful information.

However, some phenomena (e.g. turbulent flow in fluid dynamics) do not seem to be complicated, usually have nontrivial description in a language of differential equations (e.g. ODEs and PDEs).

Nowadays, computers are able to run more and more demanding computational processess in a *resonable* amount of time. Because of the mentioned above development of resources, the mathematical modelling is now being used in a great variety of reasearch fields, e.g. epigenetics, cell-biology [1, 11] or stock market prediction [35].

## 2.1 Major advantages

One of the fields in which mathematical modelling could be applied is prediction of human society's behavior. There are many questions and concerns that the sociologists are trying to answer like about the consequences of today's political decisions and social structures. For example, there are many warnings about the why how people exploit the nature and the significant amount of time that is needed to restore the natural resources. What is more, it was recently predicted by Steffen et al. [63] that no changes in use of the natural resources could cause the irreversible collapse of nature by constant growth of world average temperature. On the other hand, model was selected because of the its enough complexity as based on 4 PDEs and intelligibility.

## 2.2 Class of the predator-prey models

Model HANDY [43] was inpired by the predator-prey model. The very simple predator-prey model was created by two mathematicians independently (Lotka [39], Volterra [73]) and shows relation between numbers of individuals of two species - for example lions (predator) and antelopes (prey):

$$\begin{cases} \frac{\partial x}{\partial t} = (\alpha y)x - \beta x \\ \frac{\partial y}{\partial t} = \gamma y - (\delta x)y \end{cases} \tag{2.1}$$

where:

x - predator population

y - prey population

$\alpha$ - predator's birth rate (increases with the size of prey population)

$\beta$ - predator's death rate

$\gamma$ - prey's birth rate

$\delta$ - predation rate (connected with the size of predator population)

Presented model don't achieve the stability (equilibrium), but populations are periodically increasing and decreasing. The periodic movement near the equilibrium values:

$$\begin{cases} x_R = \frac{\gamma}{\delta} \\ y_R = \frac{\beta}{\alpha} \end{cases} \tag{2.2}$$

The typical solution of a simple predator-prey model could be observed and analysed in the Fig. 2.1.

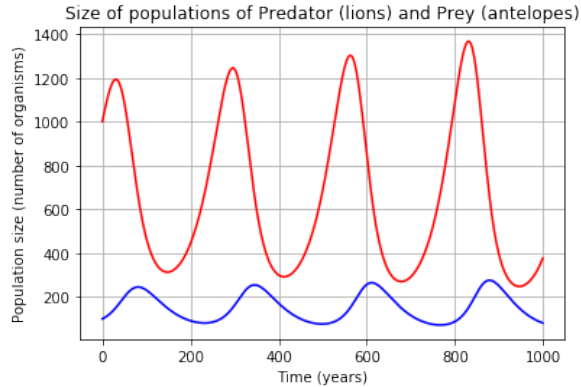

Fig. 2.1 Most common results of a simple predator-prey model. Population of Predator (lions) is a blue plot, poulation of Prey (antelopes) is red one. Parameters: $\alpha = 3.0 \times 10^{-5}$ (antelopes*years)$^{-1}$,$\beta = 2.0 \times 10^{-2}$ years$^{-1}$,$\gamma = 3.0 \times 10^{-2}$ years$^{-1}$,$\delta = 2.0 \times 10^{-4}$ (lions*years)$^{-1}$.
The initial state of a system is: $x(0) = 1.0 \times 10^2$ lions and $y(0) = 1.0 \times 10^3$ antelopes.

## 2.3 Mathematical statement

The analysed HANDY model was created by Motesharrei et al. [43] and mostly based on predator and prey model. However, authors did not simply mark people as a predators that overuse Nature (in a work of Motesharrei et al. Nature means natural potential e.g. natural resources), but also noticed that in human societies exists accumulation of surpluses (in a model known as a Wealth). What is more, model tries to cope with two main preassumptions, which came from real societies:

1. Wealth is not uniformly distributed over all humans. Rather, surpluses are controlled by elites.

2. Commoners (common people) almost only produce the wealth from nature, but do not accumulate it.

   In order to both mentioned preassumptions and also historical knowledge, authors proposed to divide society between Elites (predator) and Commoners (prey) and create four main ODEs to predict the evolution of rich (Elites) and poor (Commoners), Wealth of the society and Nature (state of natural resources in a place where the humans live):

$$\begin{cases} \frac{\partial x_C}{\partial t} = \beta_C x_C - \alpha_C x_C \\ \frac{\partial x_E}{\partial t} = \beta_E x_E - \alpha_E x_E \\ \frac{\partial y}{\partial t} = \gamma y (\lambda - y) - \delta x_C y \\ \frac{\partial w}{\partial t} = \delta x_C y - C_C - C_E \end{cases} \tag{2.3}$$

where:

$x_C$ - Commoners population

$x_E$ - Elites population

$y$ - Nature

$w$ - Wealth

$\beta_C$ - Commoners birth rate

$\beta_E$ - Elites birth rate

$\gamma$ - Regeneration rate of nature

$\delta$ - Depletion factor

$\lambda$ - Nature carrying capacity

$C_C$, $C_E$, $\alpha_C$ (Commoners death rate) and $\alpha_E$ (Elites death rate) are functions of dynamic variables ($x_C$, $x_E$, $w$) and will be decribe below.

For now, it is simple a variation of predator and prey model. Populations of both parts of society (elites and commoners) grow simultaneously with a birth rates $\beta$ and mitigate according to death rates $\alpha$. Authors proposed to take under consideration that death rates $\alpha$ should depend on wealth. The equation that describe nature is divided between two factors: first is connected with nature self-regeneration, the second is connected with depletation **made by commoners**. The last ODE is for wealth and depends on production $\delta x_C y$ and consumption of both commoners $C_C$ and elites $C_E$.

Another part of Handy's equations are modeling the average consumption within two parts of society's population - respectively $C_C$ and $C_E$:

$$\begin{cases} C_C = min(1, \frac{w}{w_{th}}) s x_C \\ C_E = min(1, \frac{w}{w_{th}}) \kappa s x_E \end{cases} \tag{2.4}$$

$$w_{th} = \rho x_C + \kappa \rho x_E \tag{2.5}$$

where:

$w_{th}$ - Wealth threshold

$s$ - Subsistence salary per capita

$\kappa$ - Inequality factor

$\rho$ - Threshold wealth per capita ("minimum consumption per capita")

Differences in consumption came from the simple observation that elites have much more wealth, which could be spend. The value of wealth threshold is minimum for the population to avoid the famine, because is a product of minimum consumption per capita and size of population with respect to inequality two groups. People consume either the smallest value of produced wealth or everything - in a case of indequate production.

The last two equations simulate the increase of death rate based on a knowledge of a consumption. If the values of consumption appear to be larger than the minimum, the death rates do not change. However, in a situation of a famine, death rates will increse.

$$\begin{cases} \alpha_C = \alpha_m + max(0, 1 - \frac{C_C}{sx_C})(\alpha_M - \alpha_m) \\ \alpha_E = \alpha_m + max(0, 1 - \frac{C_E}{sx_E})(\alpha_M - \alpha_m) \end{cases} \tag{2.6}$$

where:

$\alpha_m$ - Normal (minimum) death rate

$\alpha_M$ - Famine (maximum) death rate

Note that the factors of consumption $sx_E$ and $sx_C$ are identical with respect to populations. It suggests the equality of people in poor societies.

To summarize, the Handy model is a predator-prey class model adapted to cope with several preassumptions about human societies. Handy model consists of:

- 4 Ordinary Differential Equations (ODEs)

- 5 Linear Equations (dependent on situation of population)

- 4 state variables

- 10 parameters

Typical values and description of all variables and parameters were attached in Table 2.1.

## 2.4 Different scenarios predictions

Handy model enables simulations on different types of human societies which could be recognized as a modern as well as historical. Due to the fact that model exist in a space of a distinction on commoners or elites, only three major types can be modeled:

Table 2.1 Parameters and variables of Handy model with description and values that the model behavior is known.

| Parameter | Description | Typical value |
|-----------|-------------|---------------|
| $\alpha_m$ | Normal (minimum) death rate | $1.0 \times 10^{-2}$ |
| $\alpha_M$ | Famine (maximum) death rate | $7.0 \times 10^{-2}$ |
| $\beta_C$ | Commoners birth rate | $3.0 \times 10^{-2}$ |
| $\beta_E$ | Elites birth rate | $3.0 \times 10^{-2}$ |
| $s$ | Subsistence salary per capita | $5.0 \times 10^{-4}$ |
| $\rho$ | Threshold wealth per capita | $5.0 \times 10^{-3}$ |
| $\gamma$ | Regeneration rate of nature | $1.0 \times 10^{-2}$ |
| $\lambda$ | Nature carrying capacity | $1.0 \times 10^{2}$ |
| $\kappa$ | Inequality factor | $1, 10, 100$ |
| $\delta$ | Depletion (production) factor | None |

| Variable | Description | Typical initial value |
|----------|-------------|------------------------|
| $x_C$ | Commoners population | $1.0 \times 10^{2}$ |
| $x_E$ | Elites population | $0, 1, 25$ |
| $y$ | Nature | $\lambda$ |
| $w$ | Accumulated wealth | $0$ |

1. Egalitarian society (without elites): $x_E = 0$

2. Equitable society (without elites, but assuming that not everyone produces wealth): could be modeled with $\kappa = 1$ and $x_E \geq 0$

3. Unequal society: $x_E \geq 0$ and $\kappa > 1$

According to the results of original paper [43] on each of the society types can be modeled following scenarios:

- Soft approach to equilibrium

- Oscillatory approach to equilibrium

- Periodical states of prosperity and collapses

- Irreversible collapse

Obviously, the simulations of scenarios were repeated in order get more knowledge about Handy behavior and check the results from the original paper [43]. Futhermore, experience with more complex results enabled finding the suitable test set for futher research. On the following pages, plots of the most interesting Handy scenarios are shown.

### 2.4.1 Egalitarian society

In the scenario of egalitarian society, inequality factor $\kappa$ has no impact on the behaviour of society - it is caused by setting $x_E = 0$ during whole simulation. In fact, it can be easily observe that the population of elites (in the Fig. 2.2, Fig. 2.3 and Fig. 2.4 marked with green line) does not change during simulation.

Initial parameters for the following models came from Table 2.1. Only a depletion modifier, $\delta$ has been changed on the following models. Due to the fact, that $\delta$ has strong impact on production (red) and nature (yellow), it can be notice that too small $\delta$ could effect with not enough wealth to feed population, but too large leads to exhaustion of natural resources.

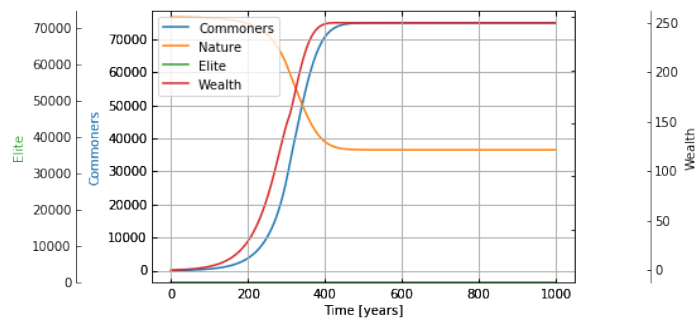

Fig. 2.2 Figure of a scenario with egalitarian society that achieve equilibrium. Model parameters: $x_E(0) = 0$ and $\delta = 1$.

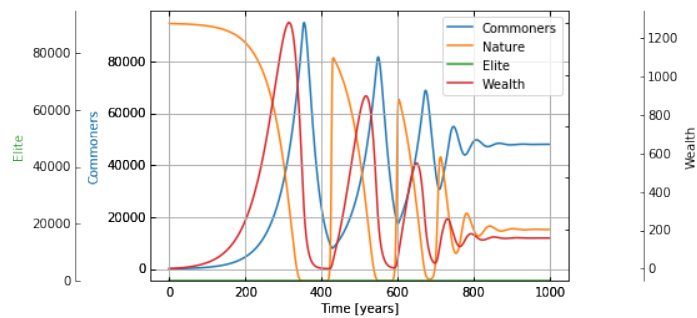

Fig. 2.3 Figure of a scenario with egalitarian society that achieve equilibrium after oscillations. Model parameters: $x_E(0) = 0$ and $\delta = 2.5$.

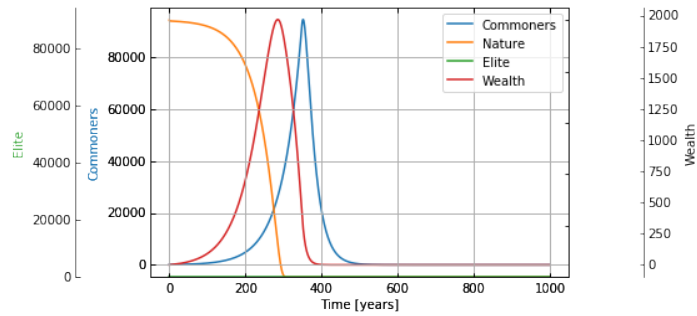

Fig. 2.4 Figure of a scenario with a collapse of egalitarian society. Model parameters: $x_E(0) = 0$ and $\delta = 5.5$.

The plot in the Fig. 2.2 presents the egalitarian society with depletion modifier $\delta = 1$, which allows to achieve the system equilibrium with soft landing. However, the depletion modifier was increased in the consecutive models - $\delta = 2.5$ in Fig. 2.3 and $\delta = 5.5$ in Fig. 2.4. As a result the society behaviour came from soft approach to the optimal equilibrium, by oscillatory approach to equilibrium to full collapse of a society.

### 2.4.2 Equitable society

Equitable society scenario with its initial parameters from Table 2.1 and inequality factor, $\kappa = 1$ seems to be pretty similar to egalitarian society. However, in this scenario elites initial population $x_E > 0$ and allow to simulate a situation when there are Workers (commoners) and Non-Workers (elites) in the society.

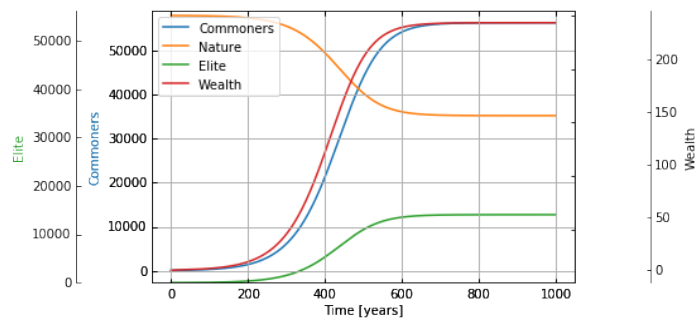

Fig. 2.5 Figure of a scenario with equitable society that achieve equilibrium. Model parameters: $x_E(0) = 25$ and $\delta = 1$.

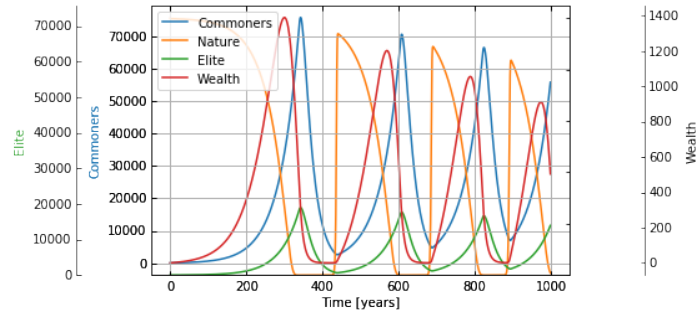

Fig. 2.6 Figure of a scenario with egalitarian society that come across cycles of times of prosperity and collapse. Model parameters: $x_E(0) = 25$ and $\delta = 3.46$.

Two above plots are different in the value of depletion modifier, $\delta$. First of them (Fig. 2.5) with $\delta = 1$ provides the society that can achieve an equilibrium after a certain time. Moreover, the population of Workers (blue) and Non-Workers (green) are growing as well as the wealth (red) of the society. The another one (Fig. 2.6) has depletion modifier, $\delta = 3.46$ that seems to be too large to ensure the approaching to equilibrium, but small enough to protect society from the irreversible collapse. There are several cycles of times of prosperity and collapse which give nature a time to restore its resources.

### 2.4.3   Unequal society

The following scenario that has a proper elites - a group of people which not only do not work but also consume ($\kappa > 1$) much more than the rest of the society. It can be easily observe that with initial parameters from Table 2.1 and inequality factor, $\kappa = 10$ ($\delta = 1$, $x_E(0) = 0.001$) no equilibrium could be achieved. What is more, another type of collapse come into play - the one associated with large growth of commoners and elites populations and their fall down (Fig. 2.7). It seems to be caused by scarcity of labour.

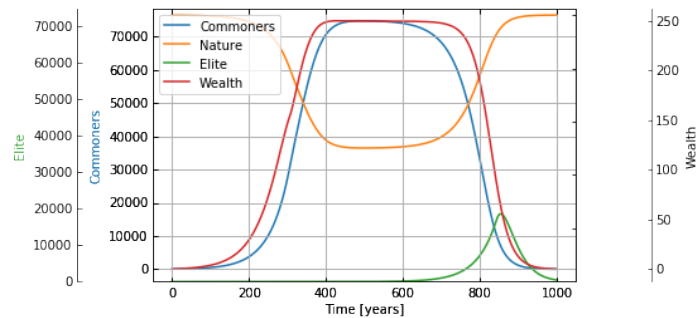

Fig. 2.7 Figure of a scenario with unequal society that collapse. Model parameters: $x_E(0) = 0.001$, $x_C(0) = 100$, $\kappa = 10$, $\delta = 1$.

In the following plots, commoners birth rate and elites birth rate were changed to respectively $\beta_C = 0.065$, $\beta_E = 0.02$.

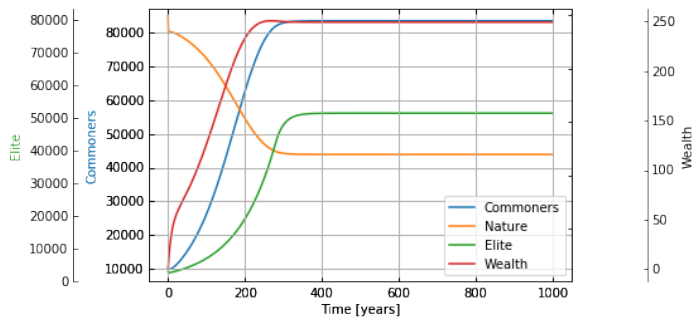

Fig. 2.8 Figure of a scenario with unequal society that achieve equilibrium. Model parameters: $x_E(0) = 3000$, $x_C(0) = 10000$, $\kappa = 10$, $\delta = 1$, $\beta_E = 0.02$, $\beta_C = 0.065$.

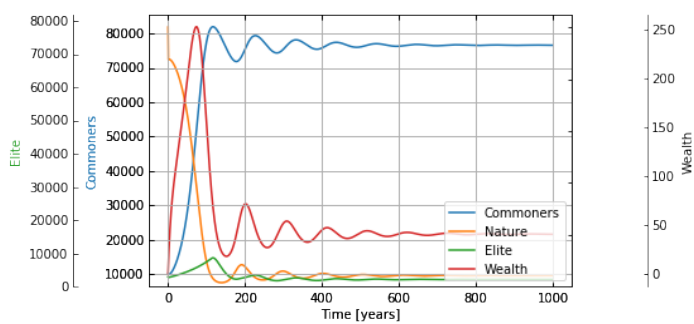

Fig. 2.9 Figure of a scenario with unequal society that achieve equilibrium after oscillations. Model parameters: $x_E(0) = 3000$, $x_C(0) = 10000$, $\kappa = 10$, $\delta = 2$, $\beta_E = 0.02$, $\beta_C = 0.065$.

Each of the above models simulates societies that achieve equilibrium. Whereas, in first (Fig. 2.8) it happen after a soft landing, in the second (Fig. 2.9) with an oscillatory approach. These scenarios show the necessary of birth policies to achieve equilibrium in unequal societies.

## 2.5 Discussion

Comparison of the results that were presented in the previous section (**2.4**) with the originals convinced to the correctness of implementation. However, after longer consideration it seems to be doubtful that model reflects reality of human societies as properly as authors suggest. Some of the Handy preassumptions (both evident as well as hidden) appear not to be consistent with human experience. Because of that, list of possible extensions of Handy will be presented below:

1. Introduction of a middle-class - a working population that have enough resources to keep part of it.

2. Change the differences in consumption between elites and commoners, beacuse elites consumption does not change even during sociiety collapse.

3. Enable the elites to participate in wealth production. Part of a elites' fund can be invest.

Obviously, the list could be much longer than mentioned 3 points, but Handy still have to be a simplicity view on the society behavior within centuries and probably will never simulate a real life.

## 2.6   Simplified Handy models

In most of Bayesian approximation methods, researchers can only find the approximated initial parameters of a well-known model. However, one of the most significant aims of the Supermodeling is the ability of predict model attractor even in a scenario when the ground-truth phenomena is not exactly known. Such a situation is called a *imperfect model class scenario* (e.g. Wiegerinck & Selten, [75]).

Because one of the major aims of this thesis is to propose the best way of approximation the imperfect models, simplifing the Handy model has been necessary. During the process of analysing the model behaviour, following simplifications were proposed:

**A** Assumption that both Commoners as well as Elites produce the same amount of wealth. It also means that Elites start to generate wealth. Two of the PDEs from (eq. (2.3)) have to be changed into:

$$
\begin{cases}
\frac{\partial y}{\partial t} = \gamma y (\lambda - y) - \delta (x_C + x_E) y \\
\frac{\partial w}{\partial t} = \delta (x_C + x_E) y - C_C - C_E
\end{cases}
\tag{2.7}
$$

**B** Set the consumption of both classes (Elites and Commoners) as equal. It resulted with change of the equations (eq. (2.4)):

$$
\begin{cases}
C_C = min(1, \frac{w}{w_{th}}) s x_C \\
C_E = min(1, \frac{w}{w_{th}}) s x_E
\end{cases}
\tag{2.8}
$$

**C** Deletion the inequality factor impact on wealth threshold (eq. (2.5)):

$$
w_{th} = \rho (x_C + x_E)
\tag{2.9}
$$

Three of the simplified models have been created by using different two of the mentioned changes. Moreover, the fourth model has been prepared with all possible simplifications. Models structures could be find with the help of Table 2.2.

Table 2.2 Proposed simplified models with the changes.

| | Proposed models | | | |
| --- | --- | --- | --- | --- |
| Change | Handy1 | Handy2 | Handy3 | Handy4 |
| **A** | ✓ | | ✓ | ✓ |
| **B** | | ✓ | ✓ | ✓ |
| **C** | ✓ | ✓ | | ✓ |

As it could be easily noticed, proposed models minimize the assumption of society as a predator-prey system and model called *Handy4* removes it entirely.

# Chapter 3

# Approximate Bayesian Computation

## 3.1 Motivation

Recently, more and more mathematical models are created in such a different fields of research as e.g. biology, epigenetics or meteorology. Because of that, researches most often come accross of selection the most appropriate model. Many of this complex real-world models can be succesfully analised with a standard approaches of model selection, but only if probabilistic models exists. However, either if the model likelihood functions are computationally intractable (too complicated) or the probablility functions do not exist, then selection of the most suitable model has to be based mostly on the agreement between observed data and simulation. In the worst case, not only the best model is not known but also the model parameters. Nevertheless, the same problems can be resolve efectivaly with the Approximate Bayesian Computation algorithms as shown in [64, 67, 66].

Approximate Bayesian Computation (ABC) is not a single algorithm, the ABC are rather a very wide class of algorithms and methods that are finding Bayesian inference [14, 22]. However, the main novelty of this methods is the fact that the model likelihood functions are being approximated when they are analytically intractable. The variety of advanatages comes from that possibility, but the most significant is the fact that ABC algorithms can be evaluate to run on datasets generated from computational models as shown in [14].

Because of that, the class of ABC methods are very popular and widely used in almost every field of mathematical modelling researches, especially in biological sciences as epigenetics, genetics and epidemiology (e.g. [41]).

## 3.2    Mathematical statement

In ABC methods, the functions of likelihood are not computed, but the likelihood is approximated by comparison of observed and simulated data.

Let us assume, that $\theta \in \mathbb{R}^n, n \geq 1$ is a vector of $n$ parameters and $p(\theta)$ is a prior distribution, then the goal is to approximate the posterior distribution $p(\theta|D)$ where the $D$ is a set of given data [6, 2].

The posterior distribution is approximated in the presented way:

$$p(\theta|D) \propto f(D|\theta)p(\theta),\tag{3.1}$$

where $f(D|\theta)$ is the function of likelihood of $\theta$ given the set of $D$ as shown in [64]. In general, exist the common form of all ABC algorithms:

1. Choose a candidate parameter vector $\widetilde{\theta}$ from the prior distribution $p(\theta)$;

2. Simulate a dataset $\widetilde{D}$ from the model given by probability distribution $f(D|\widetilde{\theta})$;

3. Check the distance in metric $d(\bullet, \bullet)$ between simulated dataset, $\widetilde{D}$, and the ground truth data $D$. If $d(D,\widetilde{D}) \leq \varepsilon$, accept $\widetilde{\theta}$. The tolerance $\varepsilon \geq 0$ is acceptable level of agreement between both simulated and ground truth datasets.

As an output of this basic ABC method, researcher can get a vector of parameters from the distribution $p(\theta|d(D,\widetilde{D}) \leq \varepsilon)$. Obviously, the smaller tolerance of agreement $\varepsilon$, the better approximation of the posterior ditribution $p(\theta|D)$ as shown in [67].

### 3.2.1   ABC rejection algorithm

The simplest version of the ABC method is the *ABC rejection algorithm* as shown in [49]:

---
**Algorithm 1:** ABC rejection algorithm

---
**while do**
> Choose $\widetilde{\theta}$ from distribution $p(\theta)$;
> Simulate a dataset $\widetilde{D}$ from $f(D|\widetilde{\theta})$;
> **if** $d(D,\widetilde{D}) \leq \varepsilon$ **then**
> > accept $\widetilde{\theta}$;
>
> **else**
> > reject;
>
> **end**

**end**

---

Although, *ABC rejection algorithm* is very simple, its major disadvantage is the fact that rate of acceptance is very low (especially with prior and posterior distribution that are not similar). There are two main classes of different methods created to deal with this problem:

- ABC based on Markov chain Monte Carlo (ABC-MCMC) [40]

- ABC based on sequential Monte Carlo techniques (ABC-SMC) [67, 5]

Throughout the next part of this section, mentioned methods will be introduced.

### 3.2.2   ABC-MCMC algorithm

The first approach, called ABC-MCMC uses Markov chain Monte Carlo method to resolve a problem that was presented above. As a result of the Algorithm 2, one is getting a Markov chain with $p(\theta|d(D,\widetilde{D}) \leq \varepsilon)$, which is the stationary distribution as shown in Marjoram et al. [40]. Despite the fact, that ABC-MCMC always converge to the given approximate posterior distribution, there are probable cons:

- probable low acceptance with the samples that are strongly correlated can cause very long Markov chains;

- chains could be localized in parts of space of low probablility for a long time.

---

**Algorithm 2:** ABC-MCMC algorithm

Initialize $\theta_i$, set i=0;

**while do**

    Select $\widetilde{\theta}$ according to distribution $g(\theta|\theta_i)$;

    Simulate a dataset $\widetilde{D}$ from $f(D|\widetilde{\theta})$;

    **if** $d(D,\widetilde{D}) \leq \varepsilon$ **then**

        Set $\theta_{i+1} = \widetilde{\theta}$ with probability

        $\alpha = min(1, \frac{p(\widetilde{\theta})g(\theta_i|\widetilde{\theta})}{p(\theta_i)g(\widetilde{\theta}|\theta_i)})$;

        Set $\theta_{i+1} = \theta_i$ with probability $1 - \alpha$;

    **else**

        $\theta_{i+1} = \theta_i$;

    **end**

    Set $i = i + 1$

**end**

---

### 3.2.3    ABC-SMC algorithm

To come across potential cons of ABC-MCMC algorithm, the method ABC-SMC that use sequential Monte Carlo has been developed (ABC-SMC algorithm's pseudocode is presented at the end of the subsection). The major difference between these two approaches is caused by come into play a set of particles - $\theta^{(1),...,\theta^{(N)}}$ (parameter values sampled from prior distribution - $p(\theta)$) that are used in a sequence of intermediate distributions $p(\theta|d(D,\widetilde{D}) \leq \varepsilon_i)$, where $i = 1,...,T-1$ as shown in Toni et al. [67]

The particles' propagation stops when good represantation of target distribution ($p(\theta|d(D,\widetilde{D}) \leq \varepsilon_T)$) is achieved. Tolerances are chosen to be a decreasing sequence $\varepsilon_1 > \cdots > \varepsilon_T \geq 0$, which in fact can approve convergence to target distribution. What is more, adequatelly large set of particles is able to neutralize opportunity of chain localization in regions of low probability.

In the following algorithm, $\widetilde{\theta}$ means a particles sampled from previous distribution, whereas these particles after perturbation will be indicated in the way: $\widetilde{\widetilde{\theta}}$. $K_t$ is a perturbation kernel, that will be a random walk, either Gaussian or uniform.

### 3.2.4   ABC for model selection

The above algorithms for parameter estimation can be used only if the model is known and well defined. However, in most of the real-world reasearches we find model selection problem, where not only parameters are not known, but also there is more than one candidate for an appropriate model. Model selection problem is closely related to the problem of parameter estimation under assumption of considering Bayesian methods.

In this case, under consideration are marginal probablilities of the models [66, 64]:

$$p(m|D) = \frac{p(D|m)p(m)}{p(D)}, \tag{3.2}$$

where: $p(D|m)$ - marginal probablility, $p(m)$ - prior probability of the model $m$.

The most demanding part is efficient evaluation of the marginal likelihood, which can be define e.g. as a result of operation: $p(D|m) = \int f(D|m, \theta)p(\theta|m)d\theta$, where parameter prior distribution for model $m$, $p(\theta|m)$.

Due to this disadvantage, both SMC and MCMC approaches are being widely used to efficient estimate the marginal distribution of a model. However, the usage of the ABC algorithm in this thesis was limited to the first case - well-defined models. Because of that, the broader introduction to the variety of approaches to model selection problem in ABC methods would not be included in this thesis.

---

**Algorithm 3:** ABC-SMC algorithm

Initialize $\varepsilon_1, \ldots, \varepsilon_T$;

Set the population indicator t=0;

**while** $t \leq T$ **do**

    Set the particle indicator i=1;

    **while** $i \leq N$ **do**

        **A**;

        **if** *t=0* **then**

            Sample $\widetilde{\widetilde{\theta}}$ independently from $p(\theta)$;

        **else**

            Sample $\widetilde{\theta}$ from revious population $\theta_{t-1}^{(i)}$ with weights $w_{t-1}$;

            Peturb the particle to obtain $\widetilde{\widetilde{\theta}} \sim K_t(\theta|\widetilde{\theta})$, ($K_t$ - peturbation kernel);

        **end**

        **if** $p(\widetilde{\widetilde{\theta}}) = 0$ **then**

            Return to **A**;

        **end**

        Simulate a dataset $\widetilde{D} \sim f(D|\widetilde{\widetilde{\theta}})$;

        **if** $d(\widetilde{D}, D) \geq \varepsilon_t$ **then**

            Return to **A**;

        **end**

        Set $\theta_t^{(i)} = \widetilde{\widetilde{\theta}}$, calculate weight for particle $\theta^{(i)}$:

$$w_t^{(i)} = \begin{cases} 1, & \text{if } t = 0, \\ \frac{p(\theta^{(i)})}{\sum_{j=1}^{N} w_{t-1}^{(j)} K_t(\theta_{t-1}^{(j)}, \theta_t^{(i)})}, & \text{if } t > 0. \end{cases}$$

        Set $i = i + 1$;

    **end**

    Normalize the weights;

    Set $t = t + 1$;

**end**

---

## 3.3    Real-world examples of the use of ABC algorithms

As it was mentioned in a previous sections, the ABC method is widely use in almost all fields of mathematical modelling. Such a great popularity has its roots in the fact that a large number of research models has well-known statistical features. Moreover, from the introduction the basics concepts that are behind ABC methods, the development is a work of statisticians and biologists. Because of that, such a large number of usecases in the biological fields. To mention just a few:

- Bacterial infections in child care center (as shown by Marttinen et al. in [41])

- The behavior of the Sonic Hedgehog (Shh) transcription factor in Drosophila melanogaster (as shown by Sunnaker et al. in [64])

- Inferring the parameters of the neutral theory of biodiversity (as shown by Csilléry et al. in [12])

- JAK-STAT signalling pathway model selection (as shown by Toni & Stumpf in [66])

- Chemical reaction kinetics (as shown by Toni & Stumpf in [66])

Nowadays, ABC methods and basics approaches are still considered (e.g. [23, 8, 77]) and used to develope newer machine learning frameworks, like: Expectation propagation - EP [71] or Hierarchical Implicit Models - HIMs and Likelihood-Free Variational Inference - LFVI as shown in Tran et al. [68].

# Chapter 4

# Supermodeling methodology

## 4.1 Motivation

It is commonly believed that many real-world phenomena have complex structure which means that their models are not precisely known. In the most common simulation experiments, exist an assumed ground truth and a set of good but not perfect models (e.g. in weather prediction, climatology or cancer growth models) [59]. The other possible scenario accepts situation when the imperfect models and ground truth have even miscellaneous structures as shown by Wiegerinck & Selten [75].

In both scenarios, the supermodel is created by coupling a set of (inaccurate) submodels. The most meaningful advantage in relation to simple ensemble learning is the fact that only the coupling coefficient have to be learned, which could implies a massive reduction of problem dimensionality (number of learned parameters). With assumption of varied initial values of parameters in every submodel, the coupling coefficients can fit dynamics to ground truth data via the process of *synchronization* [19, 20, 17, 78].

Synchronization is the phenomenon, which allows mutually coupled, oscillating systems to be coordinated. The synchronization mechanism as a data assimilation method, firstly arised from a field of climatology (e.g. Duane et al. [17], Yang et al. [78] or van der Berge [7]). Although, supermodeling have very natural and uncomplicated assumption of coupling imperfect models, the studies (e.g. Duane [16]) shows developing a supermodel as a method to increase significantly modeling efficiency.

It was shown that with fitted coupling coefficients, the supermodel dynamics can synchronise. What is more, several coefficients learning methods that were proposed, allow the supermodel to provide a better representation attractor behaviour than each of the submodels.

## 4.2 Mathematical statement

Supermodeling is a method of connection even up to several submodels that are believed to be a simplified version of real-world phenomena.

Let us assume, that exists observable vector $\mathbf{x}_{true}(t)$ (called *ground truth*) that is driven by a chaotic dynamic system. What is more, there will be $M$ different models labeled by $\mu$, but each of them is imperfect. The state vector $x_\mu(t)$ for each of these models could be set in the following way [75]:

$$\dot{x_\mu^i} = f_\mu^i(\mathbf{x}_\mu), \qquad (4.1)$$

where $i$ represents vector projections and was used the well-known Newton's notation for differentiation over time. According to this statement, $x_\mu(t)$ is just a result of ODE.

Let us define the tensor $\mathbf{C} = \{C_{\mu\nu}^i\}$ composed of coupling nonnegative coefficients $C_{\mu\nu}^i$ between models $\mu$ and $\nu$ for $ith$ variable $x^i$. The Supermodel (SUMO) is a combination of the imperfect models $\mu$ by inserting connections according to tensor $\mathbf{C}$ as shown by Dzwinel et al. [19]:

$$\dot{x_\mu^i} = f_\mu^i(\mathbf{x}_\mu) + \sum_\nu C_{\mu\nu}^i(x_\nu^i - x_\mu^i) \qquad (4.2)$$

The solution of defined SUMO can be expressed as an ensemble average over all $M$ models:

$$\mathbf{x}_s(t, \mathbf{C}) \equiv \frac{1}{M} \sum_\mu \mathbf{x}_\mu(t, \mathbf{C}). \qquad (4.3)$$

Obviously, during the process of learning, the tensor of coefficients $\mathbf{C}$ will be deriving from a training set (observable data). Following such a procedure, the supermodel could converge to an attractor similar to the ground truth [75].

In the first approach to the Supermodeling, we can go further than determine the tensor $\mathbf{C}$, but also set a connections between variables $x_\mu^i$ and ground truth with a coefficients $K^i > 0$ (example of *nudging*) [75]. Moreover, connections between imperfect models will be dependant both from synchronisation error between submodels and synchronisation error between ground truth and SUMO. Because of additional connections, the definition of SUMO will be slightly different:

$$\dot{x_\mu^i} = f_\mu^i(\mathbf{x}_\mu) + \sum_{\nu \neq \mu} C_{\mu\nu}^i(x_\nu^i - x_\mu^i) + K^i(x_{true}^i - x_\mu^i),\qquad(4.4)$$

$$C_{\mu\nu}^i = a(x_\nu^i - x_\mu^i)(x_{true}^i - \frac{1}{M}\sum_\mu x_\mu^i) - \frac{\varepsilon}{(C_{\mu\nu}^i - C_{max})^2} + \frac{\varepsilon}{(C_{\mu\nu}^i - \delta)^2},\qquad(4.5)$$

where expressions with $\varepsilon$ allow to keep coefficients $C_{\mu\nu}^i$ in an interval $(\delta, C_{max})$. The method converges according to Duane et al. [18].

The newer approach is based on minimization an error between ground truth and SUMO states, defined as an average of prediction errors $(K)$ that are measure on an interval of time $\Delta T$ [7, 69, 57].

$$E(\mathbf{C}) = \frac{1}{K\Delta T}\sum_{i=1}^{K}\int_{t_i}^{t_i+\Delta T}|\mathbf{x}_s(t,\mathbf{C}) - \mathbf{x}_{true}(t)|^2\gamma^t\,dt,\qquad(4.6)$$

where $t_i$ indicates specific time (initial state) and factor $\gamma^t \in (0,1)$ was set to reduce the impact of squared distances in further observations. Which in fact could lead to lower the influence of internal error growth as shown in [7].
In the case of discrete set of ground truth observation states, the error is a simple variation on Root-mean-square error. The error becomes a weighted squared Euclidean distance.

The second, function $E(\mathbf{C})$ minimization approach has been presented several times as a method of effective convergence into attractors of low dimensional models. Studies have shown that not only *Lorenz systems* (Lorenz 63 [37], Lorenz 84 [38]) but also real-world low dimensional tumor growth models' attractors (e.g. Dzwinel et al. [19]) can be successfully achieved with supermodels.

Although, SUMO achieves satisfying results even with two or three submodels, it was presented by Wiegerinck et al. [74] that with large number of connections, supermodel starts to proceed in a similar way to linear combination of its submodels. As an example, vector of state $\mathbf{x}$ becomes to behave as a weighted sum of imperfect submodels behaviors:

$$\dot{x}^i = \sum_\mu w_\mu^i f_\mu^i(\mathbf{x}),\qquad(4.7)$$

where $\{w^i_\mu\}$ is a normalized set of non-negative weights. What is more, these weights can be interpreted as unique eigenvectors of $L^i$ - Laplacian matrices of coefficients $C^i_{\mu\nu}$ of each component labeled by $i$.

## 4.3   Novelty approach on imperfect class model scenario

During the process of modeling it is possible to create some simplified models. However, it is strongly believed that many real-world phenomena are so complex with dozens of interactions and degrees of freedom, that finding the correct model would be really challenging. Climatology - the field that is resposible for rising up the Supermodeling method will be one of the best examples of such phenomena. Beacause of the facts that real atmosephere has lots of degrees of freedom and equations describing its behaviour could not be known or even not exist, causes the novelty approach to Supermodeling which is called *imperfect class model scenario* [75, 59].
The assumption of this approach is that ground truth model is much more complex than the known submodels and only a couple of real variables can be observed.

Due to the high dimensionality of analysing ground truth data, it is neccesary to replace visual comparison of attractors with an objective measure. The most appropriate would be a direct measure of distance between probablility densitity functions corresponding to both attractors.
To fulfill the mentioned characteristic, the Wasserstein metric (sometimes called the earth mover's distance) firstly comes to mind. Wasserstein distance which arises from the idea of *optimal transport* is becoming more and more popular in Statistics and Machine Learning. Intuitively, joint distribution $\gamma(x,y)$ denotes how much "earth" have to be moved from $x$ to $y$ [51] in order to transform one distribution $\mathbb{P}_r$ into the other $\mathbb{P}_g$:

$$W(\mathbb{P}_r, \mathbb{P}_g) = \inf_{\gamma \in \Pi(\mathbb{P}_r, \mathbb{P}_g)} \mathbb{E}_{(x,y) \sim \gamma}[||x - y||], \qquad (4.8)$$

where $\Pi(\mathbb{P}_r, \mathbb{P}_g)$ is a set of all joint distributions $\gamma(x,y)$ with marginals $\mathbb{P}_r$ and $\mathbb{P}_g$ as shown in [3, 72].

The main disadvantage of Wasserstein metric - number of problem that is exponentially large in number of dimensions (and request of exponentially large dataset), makes it completely impractical for high-dimensional problems.
Because of that, three less and less demanding distance functions have been proposed. Each of them is considering only the state means and (co)variances of the attractors.

Firstly, it was assumed that most often the distributions could be approximate by multivariate Gaussians which means and covariances are equal to original data. Then, the Wasserstein distance for two Gaussian distributions $\mathcal{N}_0$ and $\mathcal{N}_1$ is simplified and given by an expression:

$$W^2 = |\mu_1 - \mu_0|^2 + Tr(\Sigma_0 + \Sigma_1 - 2(\Sigma_0^{1/2}\Sigma_1\Sigma_0^{1/2})^{1/2}), \qquad (4.9)$$

where $(\mu_0, \Sigma_0)$ and $(\mu_1, \Sigma_1)$ are means and covariances of $\mathcal{N}_0$ and $\mathcal{N}_1$ respectively [15, 24, 75].

It is worth to mentioned, that the distance $W$ will be zero for all pairs of distributions with the same means and covariances.

Whenever one is more interested in finding shape of an model attractor (model variability) than in its average state, the previous distance measure could be restricted only to the terms connected with the covariances:

$$V^2 = Tr(\Sigma_0 + \Sigma_1 - 2(\Sigma_0^{1/2}\Sigma_1\Sigma_0^{1/2})^{1/2}) \qquad (4.10)$$

Obviously, the error could be restricted even more and analyse only the variances $\sigma^2$, as follows:

$$U^2 = |\sigma_1 - \sigma_0|^2 \qquad (4.11)$$

The last, $U$, measure could be useful especially in high dimensional systems as proposed by Wiegerinck et al. [75].

Unlike in the previous scenario (with well-known model and not known parameters), minimizing errors in short term prediction will not necessary result with SUMOs which have satisfactory attractors behaviour. What is more, it would not lead to better attractors than improving imperfect models.

Some studies during recent years analysed the problem of SUMOs training with respect to commonly used short-term predictions error and proposed several methods of so called *attractor learning*. Bayesian optimization is often mentioned as the method of choice for *attractor learning* and allows to minimize traditional short-term predictions error as well as introduced measures ($W$, $V$, $U$) and leads to favourable attractors behaviour [75, 60, 61].

# Chapter 5

# Sensitivity Analysis

## 5.1 Motivation

During the proccess of a recognition of computational model properties, Sensitivity Analysis [55] seems to be an essential step. Classification of parameters not only enabled for finding the most important features, but also the less important could be ignored in futher research.

Sensitivity analysis (SA) is an approach that allows one to identify subset of initial parameters which impact on the model outputs is the greatest. Amongst a large variety of this method applications, the insight into relationships between model input and output seems to be the most common use.

Due to the different complexity of investigated models, there exist two main group of sensitivity analysis methods [80, 29]:

- **local SA:** Evaluates changes in outputs with respect to only one initial parameter. Could be use only in the situations of linear relation between output and parameters near a specific value (in fact, only to simple models).

- **global SA:** These methods simultanously evaluate not only the contributions of each parameter to the output but also interactions between pairs of parameters. There are several main groups of global SA approaches like multiparametric sensitivity analysis [79], Method of Morris [9], Fourier amplitude sensitivity analysis (FAST) [13, 56] and Variance-based sensitivity analysis (sometimes known as Sobol method) [62, 52, 54, 53].

As it could be noticed, Handy model cannot be describe with linear functions. Moreover, the main purpose of using the sensitivity analysis is to identify the parameters of the strongest influence on model's output. Because of the above facts, we have decided to use one of the Sobol methods as a good sensitivity indicator.

## 5.2 Sobol sensitivity analysis method

Sobol method is one of the methods based on variance decomposition, which are known to be versatile and effective among other techniques. It can deal with nonlinear and nonmonotonic models or functions [54].

### 5.2.1 Sensitivity indices

Let the model be represented by the function:

$$Y = f(\boldsymbol{X}) = f(X_1, \ldots, X_k) \tag{5.1}$$

where $Y$ is a model output (scalar) and $\boldsymbol{X} = (X_1, \ldots, X_k)$ is a parameter set that consists on factors $X_i$ ($i$-th factor) as shown in [62].

A variance based first order effect for a factor $X_i$ (sometimes called *main effect* of $X_i$ on $Y$) is given by:

$$V_{X_i}(E_{\boldsymbol{X}_{\sim i}}(Y|X_i)) \tag{5.2}$$

where $\boldsymbol{X}_{\sim i}$ denotes the matrix of all factors but $X_i$ and $V_{X_i}(\cdot)$, $E_{X_i}(\cdot)$ are respectively variance and mean taken over $X_i$. It could be understand as follows that the expectation operator is for the mean of $Y$ taken over all possible values of $\boldsymbol{X}_{\sim i}$ with fixed $X_i$. While variance is taken over all values that are possible for the specific factor $X_i$.

The related sensitivity measure - first order sensitivity coefficient could be written as follows:

$$S_i = \frac{V_{X_i}(E_{\boldsymbol{X}_{\sim i}}(Y|X_i))}{V(Y)} \tag{5.3}$$

Thanks to the known relation:

$$V_{X_i}(E_{\boldsymbol{X}_{\sim i}}(Y|X_i)) + E_{X_i}(V_{\boldsymbol{X}_{\sim i}}(Y|X_i)) = V(Y) \tag{5.4}$$

recently introduced $S_i$ is normalized. That is because $V_{X_i}(E_{\boldsymbol{X}_{\sim i}}(Y|X_i))$ varies in range from 0 to $V(Y)$. While first component measures the first order effect of $X_i$, then $E_{X_i}(V_{\boldsymbol{X}_{\sim i}}(Y|X_i))$ is called the *residual* as suggested in [54].

Besides previously introduced, there are also another variance based measures. The total effect index can be defined as:

$$S_{Ti} = \frac{E_{\boldsymbol{X}_{\sim i}}(V_{X_i}(Y|\boldsymbol{X}_{\sim i}))}{V(Y)} = 1 - \frac{V_{\boldsymbol{X}_{\sim i}}(E_{X_i}(Y|\boldsymbol{X}_{\sim i}))}{V(Y)} \tag{5.5}$$

$S_{Ti}$ measures first and even higher order effects of factor $X_i$. Higher order effects can be understand as interactions between different factors. While $V_{\boldsymbol{X}_{\sim i}}(E_{X_i}(Y|\boldsymbol{X}_{\sim i}))$ is the first order effect of $\boldsymbol{X}_{\sim i}$ then $V(Y) - V_{\boldsymbol{X}_{\sim i}}(E_{X_i}(Y|\boldsymbol{X}_{\sim i}))$ have to be the contribution of all variance decomposition terms that contains $X_i$.

The variance-based decomposition applies to previously introduced function, $Y = f(X_1, \ldots, X_k)$ which is square integrable and defined over space $\Omega$ ($k$-dimensional unit hypercube):

$$\Omega = (X|0 \leq x_i \leq 1; i = 1, \ldots, k). \tag{5.6}$$

Note that the parameters possible values can always be scale into interval $[0, 1]$.

For the futher part - the introduction to the process of variance-based framework, let all the $X_i$ be uniformly distributed in the unit interval $[0, 1]$. What is more, without less of generality, it could be assume that each factor is defined in $\Omega$ and the mapping between $\Omega$ and distribution of $X_i$ is defined in $f$ [54]. The subsequent steps of the framework are:

- Functional decomposition of the function $f$ into summands with increasing dimensionality. The following scheme is known as Hoeffding decomposition [54]:

$$f = f_0 + \sum_i f_i + \sum_i \sum_{j>i} f_{ij} + \cdots + f_{12\ldots k} \tag{5.7}$$

where $f_i = f_i(X_i)$ and $f_{ij} = f(X_i, X_j)$, etc. The total number of summands is equal to $2^k$ and each of them is square integrable over a space $\Omega$. Eq. (5.7) has a unique solution due to the fact:

$$\int_0^1 f_{i_1, i_2, \ldots, i_s}(x_{i_1}, x_{i_2}, \ldots, x_{i_s}) dx_{i_w} = 0 \tag{5.8}$$

such that $1 \leq i_1 < i_2 < \cdots < i_s \leq k$ and $i_w \in \{i_1, i_2, \ldots, i_s\}$ as presented in [54]. Moreover, functions $f_{i_1, i_2, \ldots, i_s}$ are taken via following process:

$$\begin{aligned} f_0 &= E(Y), \\ f_i &= E_{\boldsymbol{X}_{\sim i}}(Y|X_i) - E(Y), \\ f_{ij} &= E_{\boldsymbol{X}_{\sim ij}}(Y|X_i, X_j) - f_j - f_j - E(Y). \end{aligned} \tag{5.9}$$

The higher order functions can be obtained in the same way.

- Connections between partial variances and considered functions, $f_{i_1,i_2,\dots,i_s}$, can be fined with following steps:

$$
\begin{aligned}
V_i =& V(f_i(X_i)) = V_{X_i}(E_{\boldsymbol{X}_{\sim i}}(Y|X_i)), \\
V_{ij} =& V(f_{ij}(X_i, X_j)) \\
=& V_{X_i X_j}(E_{\boldsymbol{X}_{\sim ij}}(Y|X_i, X_j)) - V_{X_i}(E_{\boldsymbol{X}_{\sim i}}(Y|X_i)) - V_{X_j}(E_{\boldsymbol{X}_{\sim j}}(Y|X_j)).
\end{aligned}
\tag{5.10}
$$

As well as in the previous equation (Eq. (5.9)), the higher order terms can be find in a similar way. Due to the assumption that parameters are orthogonal to each other, the variances are linked by the formula:

$$
V(Y) = \sum_i V_i + \sum_i \sum_{j>i} V_{ij} + \cdots + V_{12\dots k}.
\tag{5.11}
$$

Dividing sides by factor $V(Y)$ leads to relation between sensitivity measures:

$$
1 = \sum_i S_i + \sum_i \sum_{j>i} S_{ij} + \cdots + S_{12\dots k}.
\tag{5.12}
$$

Note that the assumption factors independence let the relations for second and higher order terms (Eq. (5.11)) to be true. Moreover, if the factors be dependent then even Eq. (5.10) will be false.

Sensitivity measures $S_i$, $S_{Ti}$ can be interpreted respectively as expected reduction in variance when $X_i$ is set and expected variance that would left if only $X_i$ is not set.

### 5.2.2 Computation of sensitivity indices $S_i$ and $S_{Ti}$

Previously introduced method seems to be straightforward and well-defined. However, most of te real-world models are nonlinear and too complex that variances could be compute in an analytical way. Due to that fact, two techniques are widely used to improve variance-based sensitivity analysis method:

- **Monte Carlo integrals** instead of trying to integrate with analytical methods,

- **Sobol' quasi-random sampling** that adds samples away from the previously sampled points to fulfill space more uniformly.

There are several slightly different estimators (because of using Monte Carlo integrals) to compute all measures $S_i$ and $S_{Ti}$ in a single set of simulations.

Suppose that $\boldsymbol{A}$ and $\boldsymbol{B}$ are independent input matrices with elements respectively $a_{ji}$ and $b_{ji}$. Size of each matrix will be $N \times k$ where $N$ is a sample size and $k$ - number of parameters, then $1 \leq j \leq N$ and $1 \leq i \leq k$.

Let define $\boldsymbol{A}_{\boldsymbol{B}}^{(i)}$ as a matrix which has all columns from $\boldsymbol{A}$ but one, $i$-th column that comes from $\boldsymbol{B}$. Matrix $\boldsymbol{B}_{\boldsymbol{A}}^{(i)}$ will be define analogously.

Then, first order effect $S_i$ can be computed as:

$$V_{X_i}(E_{\boldsymbol{X}_{\sim i}}(Y|X_i)) = \frac{1}{N} \sum_{j=1}^{N} f(\boldsymbol{A})_j f(\boldsymbol{B}_{\boldsymbol{A}}^{(i)})_j - f_0^2 \qquad (5.13)$$

where $(\boldsymbol{B})_j$ means $j$-th row of $\boldsymbol{B}$.

While $S_{Ti}$ can be obtain from an Eq. (5.5) by computing $V_{\boldsymbol{X}_{\sim i}}(E_{X_i}(Y|\boldsymbol{X}_{\sim i})$ as follows (as shown in [27]):

$$V_{\boldsymbol{X}_{\sim i}}(E_{X_i}(Y|\boldsymbol{X}_{\sim i})) = \frac{1}{N} \sum_{j=1}^{N} f(\boldsymbol{A})_j f(\boldsymbol{A}_{\boldsymbol{B}}^{(i)})_j - f_0^2. \qquad (5.14)$$

Both of the above formulas was developed by applying relation $V(Y) = E(Y^2) - E^2(Y)$ to the left-hand sides and rewriting remaining terms as integrals.

Note that simple computation of first and total order sensitivity measure, $S_i$ and $S_{Ti}$ requires cost of at least $N(2k+1)$ model evaluations.

However, using the above method proposed by Saltelli in [52] sensitivity indices could be computed with only $N(k+2)$ model runs which is almost two times fewer than in a straight way. Futhermore, the method of estimation $S_i$, $S_{Ti}$ and second order sensitivity indices, $S_{ij}$ with $N(2k+2)$ was also developed by him.

Although, recently presented estimators allow to significantly outperform the simple ones, but it does not mean that these formulas should be use. As it has been shown in the research of Saltelli et al. [54], the Jansen's estimators (introduced by Jansen in [30]) are numerically more efficient.

Let introduced the proposed by Jansen, alternative forms for $S_i$ and $S_{Ti}$ estimators. While $V_{X_i}(E_{\boldsymbol{X}_{\sim i}}(Y|X_i))$ that is needed to compute first order sensitivity indices is given via:

$$V_{X_i}(E_{\boldsymbol{X}_{\sim i}}(Y|X_i)) = V(Y) - \frac{1}{2N} \sum_{j=1}^{N} (f(\boldsymbol{B})_j - f(\boldsymbol{A}_{\boldsymbol{B}}^{(i)})_j)^2. \qquad (5.15)$$

$S_{Ti}$ indices are offered to be rather proceed using $E_{\boldsymbol{X}_{\sim i}}(V_{X_i}(Y|\boldsymbol{X}_{\sim i}))$ than $V_{\boldsymbol{X}_{\sim i}}(E_{X_i}(Y|\boldsymbol{X}_{\sim i}))$:

$$E_{\boldsymbol{X}_{\sim i}}(V_{X_i}(Y|\boldsymbol{X}_{\sim i})) = \frac{1}{2N} \sum_{j=1}^{N} (f(\boldsymbol{A})_j - f(\boldsymbol{A}_{\boldsymbol{B}}^{(i)})_j)^2. \qquad (5.16)$$

To improve other estimators, Saltelli et al. in [54] proposed to proceed second order indices, $S_{ij}$ by generalization of the Jansen's formula (Eq. (5.16)):

$$E_{\boldsymbol{X}_{\sim ij}}(V_{X_i X_j}(Y|\boldsymbol{X}_{\sim ij})) = \frac{1}{2N}\sum_{w=1}^{N}(f(\boldsymbol{A}_{\boldsymbol{B}}^{(i)})_w - f(\boldsymbol{A}_{\boldsymbol{B}}^{(j)})_w)^2.$$ (5.17)

All of the introduced estimators are currently used and reconized as one of the best methods to efficient calculate the variance based senstivity analysis indices. Moreover, during the last 25 years many studies were presented in this field, but the constantly increasing popularity of sensitivity analysis has a strong impact on new one.

# Chapter 6

# Proposed methodology

## 6.1   General idea

The main aim of the thesis is a comprehensive study on the Supermodeling [58, 75] and the ABC method [6, 64]. Due to such requirement, we have decided to propose the methodology that could improve predictions that are based on computational models. We have understood improvement both in the terms of decreasing learning time and lower forecasting errors. Firstly, let us note that the speed of minimalization the error during learning process is strongly coupled with the problem's dimensionality. Because of that, the redcution of the number of initial parameters seemed to be an obvious choice in such case. To achieve these assumptions, we decided to perform a Sensitivity Analysis [62, 54, 26] of the set of initial parameters. The mentioned methodology results with a list of the parameters and their sensitivity indices. These informations are crucial during the selection of the most sensitive variables and parameters. Moreover, the parameter's sensitivity indicates if it could be reconised for giving the strongest impact on the model behaviour.

## 6.2   Selected model behaviour

The problem of the selection an appropriate ground truth model's behaviour seems to be essential to demonstrate differences in the considered methodologies. However, we have agreed that the model behaviour should fulfill several conditions:

- **complexity** - the model's behaviour should be complex and each of the model's variables should have frequent and not smooth changes

- **dimensionality** - the number of parameters which *have* a non-negligible impact on the model's behaviour have to be as large as possible; the larger number of the sensitive parameters the easier to observe the *curse of dimensionality*

- **different behaviours** - perfectly if the test attractor could be divided into two parts that act in various ways, e.g. more chaotic behaviour and more stable or even periodic

- **learning part position** should allow to forecast behaviours in two ways: forward and backward; in other words: the ground truth part of the selected model could not be at the very beginning of the model's attractor

It could be easy to apprehend that such model will allow to compare different methods on the prediction's capabilities on both more and less demanding behaviour of the selected model.

We have found that the above conditions could be fulfilled only by the Handy model in the most complex version - the unequal society (Sec. 2.4.3). Because many of the presented scenarios have very smooth attractors, we decided to use the scenario when an unequal society achieve equilibrium after oscillations (Fig. 2.9). Moreover, we slightly peturbed it's initial parameters. The complete list of the ground truth model's parameters will be presented below in the Tab. 6.1. What is more, the value of the inequality factor, $\kappa$ is related to the optimal value for the unequal society as shown in Motesharrei et al. [43].

Table 6.1 Parameters and variables' initial values for the selected behaviour of the Handy model.

| Parameter | Description | Value |
|-----------|-------------|-------|
| $\alpha_m$ | Normal (minimum) death rate | $1.0 \times 10^{-2}$ |
| $\alpha_M$ | Famine (maximum) death rate | $7.0 \times 10^{-2}$ |
| $\beta_C$ | Commoners birth rate | $6.5 \times 10^{-2}$ |
| $\beta_E$ | Elites birth rate | $2.0 \times 10^{-2}$ |
| $s$ | Subsistence salary per capita | $5.0 \times 10^{-4}$ |
| $\rho$ | Threshold wealth per capita | $5.0 \times 10^{-3}$ |
| $\gamma$ | Regeneration rate of nature | $1.0 \times 10^{-2}$ |
| $\lambda$ | Nature carrying capacity | $1.0 \times 10^{2}$ |
| $\kappa$ | Inequality factor | $1.0 \times 10^{1}$ |
| $\delta$ | Depletion (production) factor | $2.04 \times 10^{0}$ |
| $\mu$ | Elites to commoners equilibrium | $6.5 \times 10^{-1}$ |

| Variable | Description | Initial value |
|----------|-------------|---------------|
| $x_C$ | Commoners population | $1.0 \times 10^{4}$ |
| $x_E$ | Elites population | $3.0 \times 10^{3}$ |
| $y$ | Nature | $1.0 \times 10^{2}$ |
| $w$ | Accumulated wealth | $1.0 \times 10^{2}$ |

Furthermore, we also presented the selected model's attractor in the Fig. 6.1 to prove that the model meets the requested conditions. The graph is divided into three same-length intervals by the vertical lines. These parts could be recognised as forecasting backward (years: 0-150), learning part (years: 150-300) and forecasting forward (years: 300-450) respectively. In addition, it could be observed that the part for the forward forecasting is much more stable and periodic than the others (in fact, it is oscillatory behaviour), whereas the first part is an example of the chaotic system.

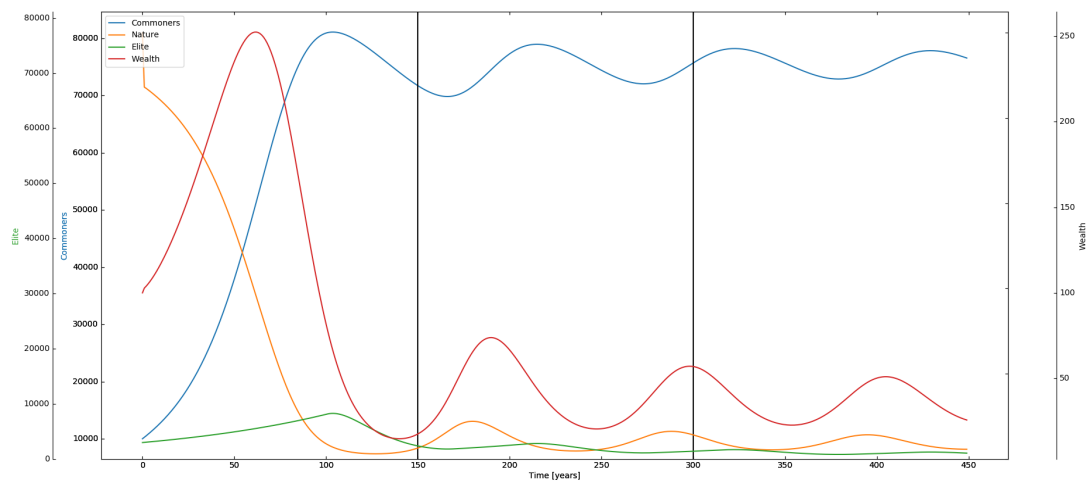

Fig. 6.1 The selected Handy model split into three parts: the ground truth (in the middle) and the parts for predictions - forward and backward.

Due to the fact that the further behaviour of the selected model seems to be oscillatory, we demonstrate the more complete model's attractor (longer time - 1000 years). The specific graph is presented in the Fig. 6.2.

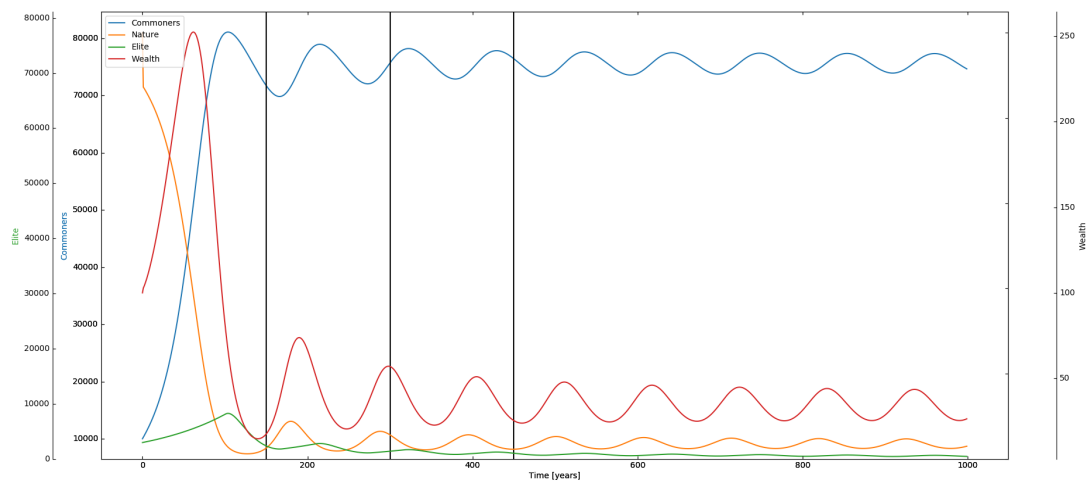

Fig. 6.2 The selected Handy model that is visualized for 1000 years to the better demonstration of the oscillatory behaviour. The attractor is divided into four parts: forecasting backward (years: 0-150), the ground truth (years: 150-300), forecasting forward (years: 300-450) and the further oscillatory behaviour (years: 450-1000).

### 6.2.1 Model sampling

The process of analysing the selected model could be divided into several steps. On of the most demanding is the decision about an adequate number of the chosen points for the learning part as well as the forecasting parts. This decision is crucial for the rest of the research, because it means the number of the points which will be fixed. However, it is hard to answer the question if the proposed number of points will cause with an effective prediction.

What is more, various numbers of points could be considered as the different forecasting cases. Anyway, let us consider the following proposition: if the larger number of points will effect with better forecasting results or at least with better results for the nearest past/future. Moreover, in the real-world applications, many times data are being collected with the fixed (lower or greater) frequency and the scientists will predict the further phenomenon's behaviour with the same frequency.

Due to the above considerations, we have decided to study four various sampling frequencies, select the most adequate frequencies and treat them like separable problems. While the frequency $f$ is understand as the number of same-length intervals that the part of the model's attractor is divided into. After this procedure, the points (values) have been selected from the beginning and the end of each interval. Let us notice that the sampling with the frequency $f$ means the selection of $f + 1$ points from each part of the model. Consequently, we have decided that each of the forecasting parts have to be sampled with the same frequency.

To have better understanding of the model's sampling, we presented samplings with four frequencies $f_1 = 5$, $f_2 = 10$, $f_3 = 15$ and $f_4 = 50$ respectively in the Fig. 6.3, 6.4, 6.5 and 6.6.

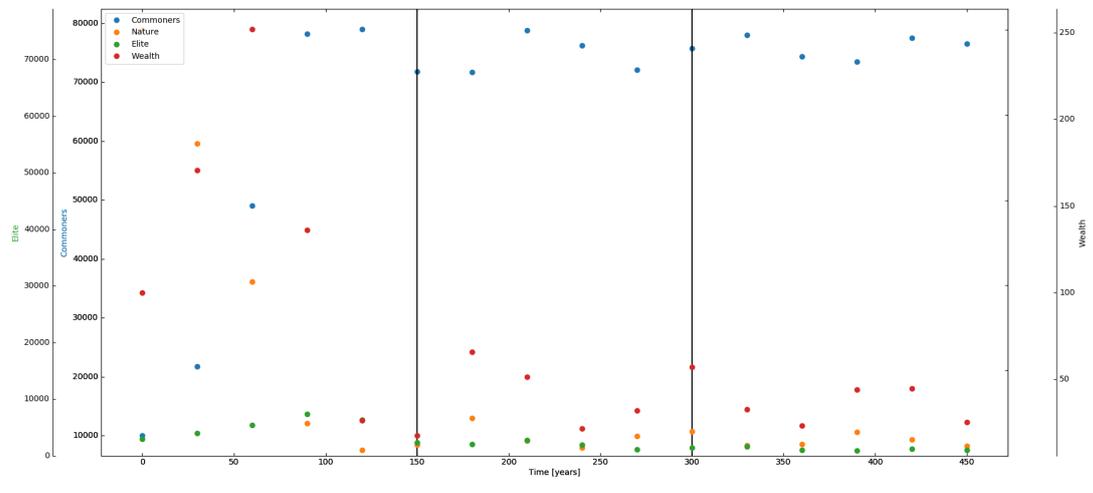

Fig. 6.3 Sampling of the selected model with the frequency $f_1 = 5$ - each of the presented intervals is sampled in the same way.

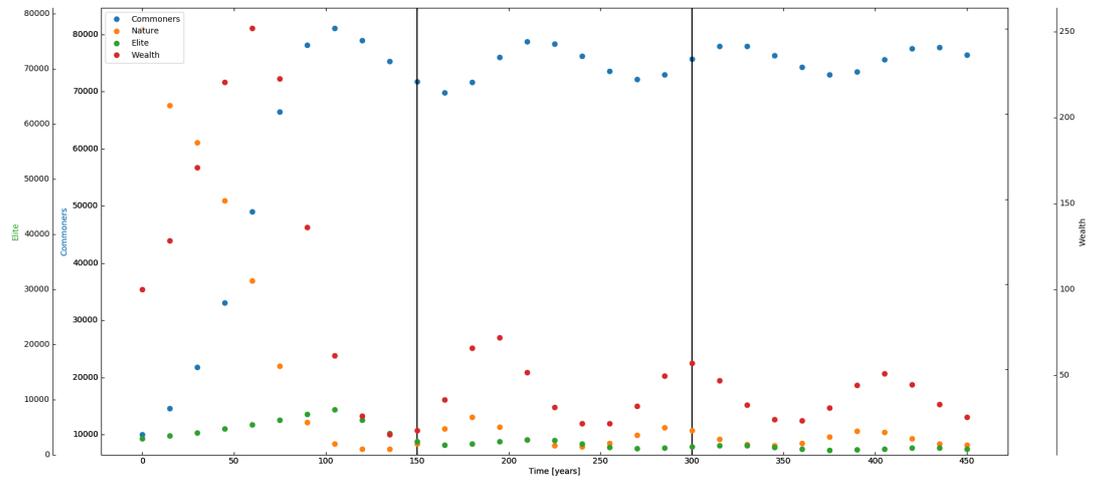

Fig. 6.4 Sampling of the selected model with the frequency $f_2 = 10$ - each of the presented intervals is sampled in the same way.

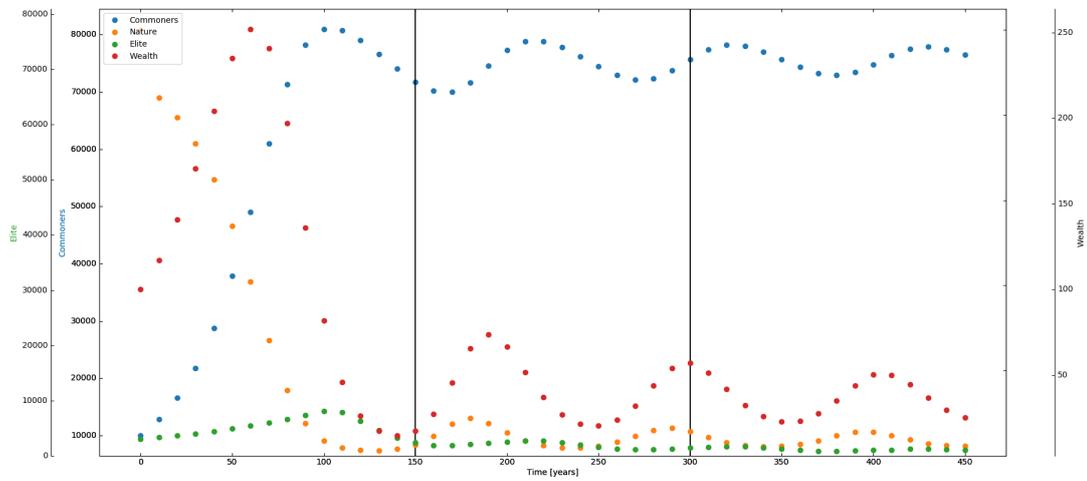

Fig. 6.5 Sampling of the selected model with the frequency $f_3 = 15$ - each of the presented intervals is sampled in the same way.

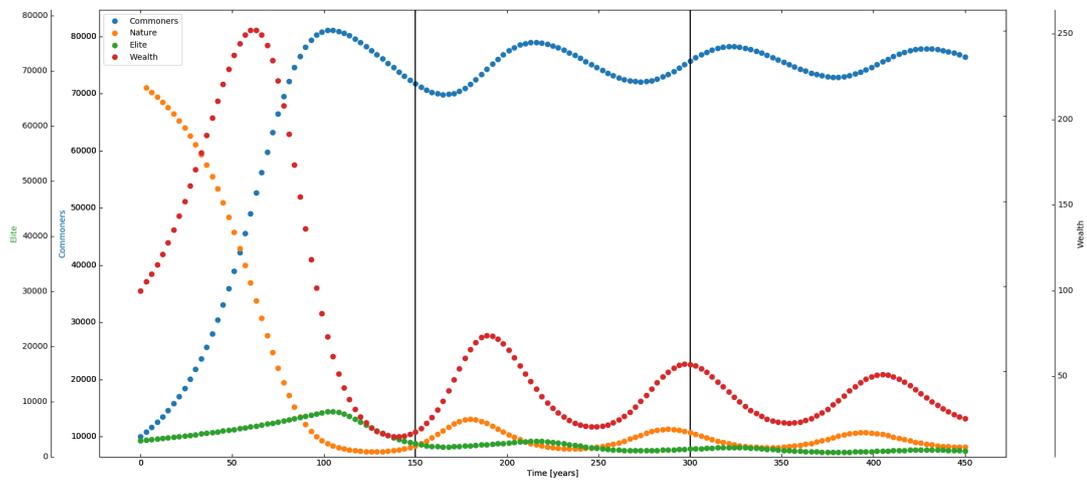

Fig. 6.6 Sampling of the selected model with the frequency $f_4 = 50$ - each of the presented intervals is sampled in the same way.

## 6.3 Approximate Bayesian Computation - reference model

Due to the fact that main topic of this thesis is oriented on comparison between new data assimilation approaches and the more classical, existence of the reference model was necessary. That is why, we decided to use the Approximate Bayesian Computation as the first methodology and then we compared the results to the Ground Truth models. The proposed tests set looked similarly for each of the frequencies learning root-mean-square errors (RMSE) [36, 10].

The research process consisted of several stepd and looked as follows:

1. Generation of the Ground Truth model (GT) from the Handy model.

2. Repeating the tests for the sampling frequencies $f_i$ from the set $\{15, 50\}$:

    (a) Sampling observations data from the GT with the frequency $f_i$ (from the learning part, $t_{start} = 150$, $t_{end} = 300$).

    (b) Creating intervals for each of the model parameters as $\pm 10\%$ of the exact value.

    (c) Repeating 10 times the learning procedure:

        i. Running the ABC method for fixed parameters' intervals and GT data.

        ii. Generating predictions (forward and backward) from the learned model, sampling these data and computing the forecasting errors.

The error measure, which we considered as the most appropriate for the ABC method was the well-known root-mean-square error (RMSE). The initial parameters for the ABC method were set to the default values. However, the value of the population size was equal to 100. Moreover, it has to be said that we have used the implementation of the ABC method from a Python's library - pyABC in v. 0.9 [33].

Due to the fact that the test process required a maximal time for the ABC learning, we had to modify the implementation and add the adequate option. In addition, we have considered the mean time of learning to the certain RMSE as the time that allows to achieved a similar model (with similar learning error).

Because the number of the tests have to be greater than 5 (to be statistical significant), we have decided to repeat 10 times each of the tests. What is more, for the frequencies $f_3 = 15$ and $f_4 = 50$, we considered only 4 various RMSEs during the learning process: 10.0, 5.0, 2.0 and 1.0. However, for the rest of the frequencies, we slightly modified the test process and considered only the value of the learning RMSE that was equal to 1.0.

Nevertheless, during the observations we have noticed that reaching the RMSE of 2.0 or 1.0 is strongly connected with very large standard deviations of the mean times (up to several times greater than a mean time). Obviously, it was caused by stucking in a local minima by statistically one over ten the ABC test. Due to the above fact, we cleared the results by removing the quantiles of 0.05 and 0.95, which gave us the reduced results sets. The reduced sets had 8 elements and the much lower relative standard deviation.

After the tests, we calculated the prediction errors and their standard deviation. Once again, we have used the RMSE on the forecasting parts. In addition, we calculated not only the mean forecasting forward error ($\overline{ff}$), but also the mean forecasting both ways error ($\overline{f2w}$), which means that the errors were being computed on the forward and backward parts at the same time. However, after the research process, we decided that it would be better to demonstrate the mean errors for the backward prediction. The mentioned measures have been calculated with respect to the following equation (Eq. (6.1)):

$$\overline{fb} = \sqrt{2\overline{f2w}^2 - \overline{ff}^2} \qquad (6.1)$$

Where $\overline{fb}$ is a mean forecasting backward error. Let us notice that we could not compute the standard deviation for this measure.

The results that are presented in this thesis demonstrate the values for the following measures:

- $\overline{t}[s]$ - a mean learning time, presented in seconds

- $\overline{t_{std}}[s]$ - a standard deviation for the $\overline{t}[s]$, presented in seconds

- $\frac{\overline{t_{std}}}{\overline{t}}$ - a ratio of the $\overline{t_{std}}$ and $\overline{t}$

- $\overline{ff}$ - a mean forward forecasting error

- $\overline{ff_{std}}$ - a standard deviation for the $\overline{ff}$

- $\frac{\overline{ff_{std}}}{\overline{ff}}$ - a ratio of the $\overline{ff_{std}}$ and $\overline{ff}$

- $\overline{f2w}$ - a mean both ways (backward and forward) forecasting error

- $\overline{ff_{std}}$ - a standard deviation for the $\overline{f2w}$

- $\overline{fb}$ - a mean backward forecasting error

In the following table (Tab. 6.2), we present the results for the ABC method.

Table 6.2 Results for the ABC method.

| Learning RMSE | $\overline{t}$[s] | $\overline{t_{std}}$[s] | $\overline{\frac{t_{std}}{t}}$ | $\overline{ff}$ | $\overline{ff_{std}}$ | $\overline{\frac{ff_{std}}{ff}}$ | $\overline{f2w}$ | $\overline{f2w_{std}}$ | $\overline{fb}$ |
|---|---|---|---|---|---|---|---|---|---|
| | | | | *sampling frequency* $f = 5$ | | | | | |
| **1.0** | **473.17** | **118.12** | **0.25** | **5.41** | **3.87** | **0.72** | **27.63** | **10.97** | **38.70** |
| | | | | *sampling frequency* $f = 10$ | | | | | |
| **1.0** | **673.67** | **131.64** | **0.20** | **2.93** | **2.07** | **0.71** | **24.44** | **16.56** | **34.44** |
| | | | | *sampling frequency* $f = 15$ | | | | | |
| **1.0** | **368.25** | **42.00** | **0.11** | **2.50** | **1.03** | **0.41** | **17.11** | **10.20** | **24.07** |
| 2.0 | 307.52 | 66.12 | 0.22 | 5.88 | 2.23 | 0.38 | 27.69 | 9.21 | 38.72 |
| 5.0 | 242.66 | 45.77 | 0.19 | 20.23 | 15.81 | 0.78 | 56.81 | 25.09 | 77.75 |
| 10.0 | 137.02 | 39.87 | 0.29 | 49.25 | 60.46 | 1.23 | 112.07 | 85.06 | 150.64 |
| | | | | *sampling frequency* $f = 50$ | | | | | |
| **1.0** | **426.72** | **33.10** | **0.08** | **3.64** | **1.62** | **0.46** | **18.66** | **11.33** | **26.14** |
| 2.0 | 337.48 | 82.09 | 0.24 | 5.89 | 2.02 | 0.34 | 21.39 | 7.83 | 29.67 |
| 5.0 | 305.67 | 121.10 | 0.40 | 22.77 | 12.12 | 0.42 | 78.90 | 34.93 | 109.23 |
| 10.0 | 191.20 | 62.50 | 0.33 | 40.46 | 23.39 | 0.58 | 88.95 | 46.49 | 119.11 |

Taking the results under consideration, we have stated the following findings:

1. There is a correlation between mean learning times and learning RMSEs - The lower RMSE the longer mean learning time $\overline{t}$

2. Both $\overline{fb}$ and $\overline{ff}$ are lower for the smaller learning error, *Learning RMSE*.

3. The ABC method is much better for the forward prediction than the backward prediction. Moreover, it means that the ABC is worse for the more chaotic systems.

We have also demonstrated the results for the mean learning times (with their standard deviation) and $\overline{f2w}$ (with their standard deviations) for the various sampling frequencies. The plots are shown in the Fig. 6.7 and 6.8.

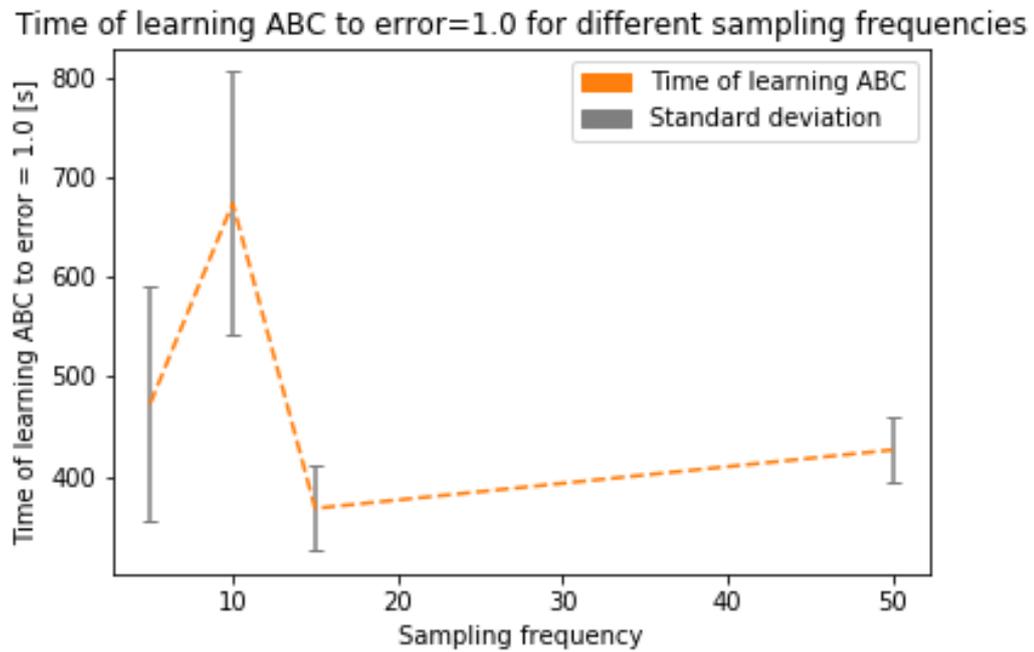

Fig. 6.7 Times of learning the ABC method for the various sampling frequencies.

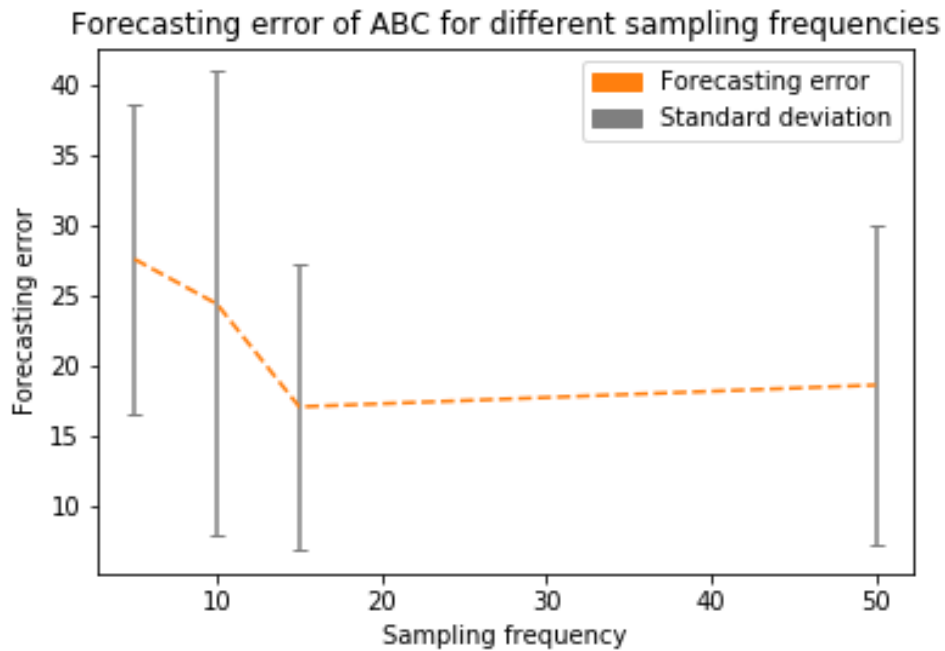

Fig. 6.8 Prediction errors (*f2w*) of the ABC method for the various sampling frequencies.

We have observed that the lower value for $\bar{t}$ has been reached when the value of sampling frequency $f$ was equal to 15. Moreover, the ABC method for $f = 50$ took slightly longer. Additionally, the mean time for the rest of the sampling frequencies ($f_1 = 5$ and $f = 10$) took respectively 28.5% and 83.0% longer than for the lowest time. Furthermore, it could be noticed that the mean errors $\overline{f2w}$ act in a similar way to the mean times. Therefore, we have achieved the nearest predictions to the Ground Truth with the sampling frequency $f_3 = 15$ and a little bit worse for the $f_4 = 50$. Moreover, we have observed the phenomenon of the *underfitting* for sparser sampling frequencies and the *overfitting* for the denser sampling. Because of that, we have decided to have two reference models:

- the ABC method with the learning RMSE of the value 1.0 for the sampling frequency $f_3 = 15$

- the ABC method with the learning RMSE of the value 1.0 for the sampling frequency $f_4 = 50$

The mean values of initial parameters for both the reference models are presented in the Tab. 6.3. Let us notice that the achieved parameters are near to the Ground Truth parameters and have negligible standard deviations. Morever, the reference model for $f = 15$ has initial parameters that are mostly closer to the Ground Truth than reference model for $f = 50$.

Table 6.3 Mean values for the parameters and the variables achieved by the ABC method. The results for two various sampling frequencies and the Ground Truth (GT).

| Parameter | Description | GT value | Frequency | | | |
|---|---|---|---|---|---|---|
| | | | 15 | | 50 | |
| | | | mean | std | mean | std |
| $\alpha_m$ | Normal (minimum) death rate | $1.00 \times 10^{-2}$ | $9.94 \times 10^{-3}$ | $3.07 \times 10^{-4}$ | $9.91 \times 10^{-3}$ | $7.08 \times 10^{-4}$ |
| $\alpha_M$ | Famine (maximum) death rate | $7.00 \times 10^{-2}$ | $7.00 \times 10^{-2}$ | $4.02 \times 10^{-4}$ | $6.99 \times 10^{-2}$ | $6.79 \times 10^{-4}$ |
| $\beta_C$ | Commoners birth rate | $6.50 \times 10^{-2}$ | $6.50 \times 10^{-2}$ | $3.94 \times 10^{-4}$ | $6.49 \times 10^{-2}$ | $6.79 \times 10^{-4}$ |
| $\beta_E$ | Elites birth rate | $2.00 \times 10^{-2}$ | $2.00 \times 10^{-2}$ | $3.13 \times 10^{-4}$ | $1.99 \times 10^{-2}$ | $6.89 \times 10^{-4}$ |
| $s$ | Subsistence salary per capita | $5.00 \times 10^{-4}$ | $4.99 \times 10^{-4}$ | $9.00 \times 10^{-6}$ | $5.05 \times 10^{-4}$ | $5.00 \times 10^{-6}$ |
| $\rho$ | Threshold wealth per capita | $5.00 \times 10^{-3}$ | $4.99 \times 10^{-3}$ | $9.30 \times 10^{-5}$ | $5.04 \times 10^{-3}$ | $4.80 \times 10^{-5}$ |
| $\gamma$ | Regeneration rate of nature | $1.00 \times 10^{-2}$ | $1.00 \times 10^{-2}$ | $2.19 \times 10^{-4}$ | $9.87 \times 10^{-3}$ | $1.55 \times 10^{-4}$ |
| $\lambda$ | Nature carrying capacity | $1.00 \times 10^{2}$ | $9.98 \times 10^{1}$ | $1.91 \times 10^{0}$ | $1.01 \times 10^{2}$ | $1.33 \times 10^{0}$ |
| $\kappa$ | Inequality factor | $1.00 \times 10^{1}$ | $1.00 \times 10^{1}$ | $1.62 \times 10^{-2}$ | $9.99 \times 10^{0}$ | $1.54 \times 10^{-2}$ |
| $\delta$ | Depletion (production) factor | $2.04 \times 10^{0}$ | $2.05 \times 10^{0}$ | $2.73 \times 10^{-2}$ | $2.05 \times 10^{0}$ | $4.49 \times 10^{-2}$ |
| $\mu$ | Elites to commoners equilibrium | $6.50 \times 10^{-1}$ | $6.46 \times 10^{-1}$ | $1.01 \times 10^{-2}$ | $6.47 \times 10^{-1}$ | $1.68 \times 10^{-2}$ |
| Variable | Description | GT value | mean | std | mean | std |
| $x_C$ | Commoners population | $1.00 \times 10^{4}$ | $9.97 \times 10^{3}$ | $7.38 \times 10^{1}$ | $9.94 \times 10^{3}$ | $1.57 \times 10^{2}$ |
| $x_E$ | Elites population | $3.00 \times 10^{3}$ | $3.0 \times 10^{3}$ | $3.74 \times 10^{0}$ | $3.00 \times 10^{3}$ | $6.73 \times 10^{0}$ |
| $y$ | Nature | $1.00 \times 10^{2}$ | $1.01 \times 10^{2}$ | $4.28 \times 10^{0}$ | $9.99 \times 10^{1}$ | $6.80 \times 10^{0}$ |
| $w$ | Accumulated wealth | $1.00 \times 10^{2}$ | $1.00 \times 10^{2}$ | $4.05 \times 10^{0}$ | $1.03 \times 10^{2}$ | $6.42 \times 10^{0}$ |

Besides, the results presentation in a form of a table, we have decided to demonstrate the attractor of the most changeable variable. Below (in the Fig. 6.9) we show the prediction of the *elites population* for the sampling frequency $f = 15$. The prediction is visualized in a normal form and also normalized form. Evenmore, we highlighted the standard deviation of the value in each point with grey colour. The results for the rest of variables and for the denser sampling are presented in the Appendix A (Fig. A.1 - A.8).

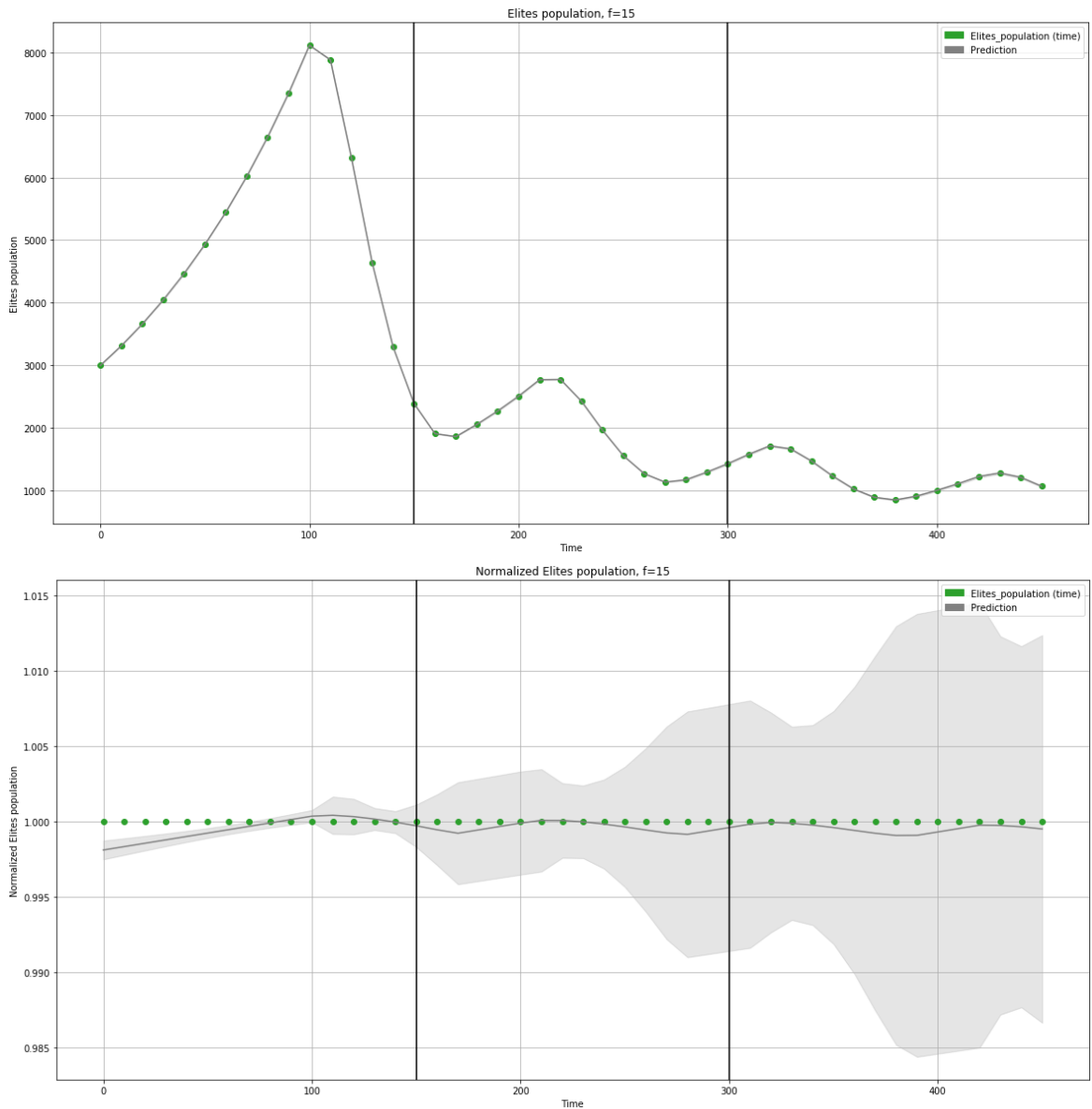

Fig. 6.9 Reference results - elites population (both real and normalized), *f = 15*

The presented reference attractor could provide to the following findings:

- The results are very close to the Ground Truth data. However, the ABC method has better predictions for the more chaotic behaviours, whereas the worse for the stable behaviours.

- The reference model converges after about the point that represents 100 years.

## 6.4 Sensitivity analysis of the model parameters

Basically, recognition of the model behaviour is one of the most demanding parts during the process of data assimilation. Because greater number of the model parameters results in increasing both time and complexity while minimization of cost functions. That is why one can consider reduction of initial parameters and chosing only the most significant - parameters which have the strongest influence into final state of the model.

After all, it could be true that minimization only a part of parameters can causes in better assimilation than all of them in the same fixed time.

In this thesis the Variance Based Sensitivity Analysis [62, 54, 44] method was used to get the knowledge about impact of each of the initial parameters and in effect:

- chose **only sensitive parameters** to the ABC method,

- chose **the most sensitive variable** to Supermodeling methodology.

### 6.4.1 Society quality measure

First question that come across after the idea of sensitivity measurement is finding an appropriate factor that could be change with respect to parameters modification. The natural choice seems to be analyse of the model variables. However, in this particular case, there is not one variable but four of them.

Another question is in fact consideration if variable's attractors should be investigate on that step of research. For the second question, answer is simple - at this step interesting is the final effect of model prediction (final value).

Due to the fact that the computational model should be a scalar value, finding one factor that could describe quality of life in population is considered as demanding. Understanding that in real life the value of natural resources could not be measure in analytical way with any reasonable accuracy, the factor should combine three other variables. Following methods can be considered:

- GDP per capita: $\frac{w}{x_C + x_E}$. Well-known method of measure in modern economical analysis, tell nothing about wealth distribution.

- Ratio between elites and commoners populations: $\frac{x_E}{x_C}$. The hidden measure of wealth distribution across population, but the accumulated wealth of a society is not taken under consideration.

- More complex trying to merge distribution and wealth of a society.

For the study GDP was used as a measure of quality of a society.

### 6.4.2   Sensitivity analysis tests

To achieve the aims that are mentioned before, the Python's library SALib (Sensitivity Analysis in Python) [26] with the Sobol method [54] have been chosen. For each of the models (Handy model and 4 simplified models) the sample size was set to $N = 5.0 \times 10^4$ and the number of both parameters and initial variables $k = 15$. Taking under consideration that not only $S_i$ and $S_{Ti}$ sensitivity indices were generated but also the second order measures $S_{ij}$, it means that each model was evaluated $N(2k+2) = 1.6 \times 10^6$ times.

In order to correctly perform the Sensitivity Analysis, each parameter has to be from a set interval. Lower and upper bounds of the intervals have been arbitrary fixed around the most common initial parameters for the complicated model behaviour (unequal society). That is to satisfy the selected Ground Truth model. Vast majority of initial parameters have bounds at most $\pm 50\%$ away from it. However, some parameters has to be in a wide interval to satisfy different model behaviours. The best example will be inequality factor $\kappa \in [5.0 \times 10^0, 1.0 \times 10^2]$ which is responsible for larger differences between commoners and elites and in fact is one of the parameters with the greatest impact on type of society.

### 6.4.3   Results

All the parameters intervals are presented below in the Tab. 6.4. Results of sensitivity analysis for the Handy model and $S_T$, $S_1$ indices for simplified models are presented in the Tab. 6.5, Tab. 6.6, Tab. 6.7 respectively.

Note that the values of $S_1$ often are lower than 0 which seems to be an unusual behavior because the $S_1$ indices should be greater or equal than 0. However, in these cases, the confidence bounds are greater than absolute value of parameters first order indice, which means that could be greater than 0.

Due to the above observation, the total sensitivity indices, $S_T$ appear to be much better measure of sensitivity for the considered models. In the following tables, confidence rate will be presented only for Handy model and results will be sorted by $S_T$ for Handy model or Handy4.

Table 6.4 Ranges of values for parameters and variables of Handy model that was used to Sensitivity Analysis.

| Parameter | Description | Lower bound | Upper bound |
|-----------|-------------|-------------|-------------|
| $\alpha_m$ | Normal (minimum) death rate | $5.0 \times 10^{-3}$ | $1.5 \times 10^{-2}$ |
| $\alpha_M$ | Famine (maximum) death rate | $5.0 \times 10^{-2}$ | $9.0 \times 10^{-2}$ |
| $\beta_C$ | Commoners birth rate | $1.0 \times 10^{-2}$ | $7.0 \times 10^{-2}$ |
| $\beta_E$ | Elites birth rate | $1.0 \times 10^{-2}$ | $7.0 \times 10^{-2}$ |
| $s$ | Subsistence salary per capita | $3.0 \times 10^{-4}$ | $7.0 \times 10^{-4}$ |
| $\rho$ | Threshold wealth per capita | $3.0 \times 10^{-3}$ | $7.0 \times 10^{-3}$ |
| $\gamma$ | Regeneration rate of nature | $7.0 \times 10^{-3}$ | $1.3 \times 10^{-2}$ |
| $\lambda$ | Nature carrying capacity | $7.0 \times 10^{1}$ | $1.3 \times 10^{2}$ |
| $\kappa$ | Inequality factor | $5.0 \times 10^{0}$ | $1.0 \times 10^{2}$ |
| $\delta$ | Depletion (production) factor | $1.0 \times 10^{0}$ | $1.5 \times 10^{1}$ |
| $\eta$ | Elites to commoners equilibrium | $5.0 \times 10^{-1}$ | $8.0 \times 10^{-1}$ |
| **Variable** | **Description** | **Lower bound** | **Upper bound** |
| $x_C(0)$ | Commoners population | $7.0 \times 10^{1}$ | $1.3 \times 10^{2}$ |
| $x_E(0)$ | Elites population | $1.0 \times 10^{-2}$ | $4.0 \times 10^{1}$ |
| $y(0)$ | Nature | $7.0 \times 10^{1}$ | $1.3 \times 10^{2}$ |
| $w(0)$ | Accumulated wealth | $0$ | $1.8 \times 10^{2}$ |

After the sensitivity analysis, the following results were obtained:

- **Handy model - parameters sensitivity**

As it was presented in the Tab. 6.5, all parameters could be divide into two separable groups - sensitive and insensitive. The parameters which total sensitive measure $S_T \geq 4.0 \times 10^{-2}$ are considered as sensitive ones. Such division gives ratio $8 : 15$ - sensitive parameters to all parameters, which effects in the possibility of reduction the number of initial parameters almost by halve.

Taking under consideration that initial parameters have strong impact on variables, it is possibly to select the most sensitive variable. Note that sensitive parameters have strong effects on mainly 2 variables: commoners population and elites population. Despite the fact that the most sensitive parameter is commoners birth rate, elites population $x_E$ is considered to be most significant. That is due to the fact that more sensitive parameters have association with $x_E$ by model equations ($\beta_E$, $\alpha_m$, $\kappa$, $x_E(0)$, $s$) than with $x_C$ ($\beta_C$, $\alpha_m$, $s$). Obviously, the second most sensitive variable is commoners population. Such information is crucial in the Supermodeling.

Moreover, results show that Handy model in the unequal society scenario behaves intuitively - elites and society inequalities have the greatest impact on model state. This behavior could be explain by model construction with respect to predator-prey type.

Table 6.5 Results of a Variance-based Sensitivity Analysis of Handy model in approximation to three significant figures. Ordered by $S_T$ measure.

| Parameter | Description | $S_1$ | $S_{1conf}$ | $S_T$ | $S_{Tconf}$ |
|---|---|---|---|---|---|
| $\beta_C$ | Commoners birth rate | $5.04 \times 10^{-2}$ | $2.79 \times 10^{-2}$ | $1.23 \times 10^0$ | $2.49 \times 10^{-1}$ |
| $\beta_E$ | Elites birth rate | $1.31 \times 10^{-2}$ | $7.63 \times 10^{-3}$ | $9.99 \times 10^{-1}$ | $1.99 \times 10^{-1}$ |
| $\alpha_m$ | Normal (minimum) death rate | $7.56 \times 10^{-4}$ | $1.09 \times 10^{-3}$ | $5.50 \times 10^{-1}$ | $1.01 \times 10^{-1}$ |
| $\delta$ | Depletion (production) factor | $-1.57 \times 10^{-4}$ | $6.27 \times 10^{-4}$ | $1.85 \times 10^{-1}$ | $7.00 \times 10^{-2}$ |
| $\kappa$ | Inequality factor | $1.74 \times 10^{-4}$ | $5.21 \times 10^{-4}$ | $1.54 \times 10^{-1}$ | $4.18 \times 10^{-2}$ |
| $x_E(0)$ | Elites population (initial value) | $1.68 \times 10^{-4}$ | $4.88 \times 10^{-4}$ | $1.25 \times 10^{-1}$ | $4.17 \times 10^{-2}$ |
| $\lambda$ | Nature carrying capacity | $2.22 \times 10^{-4}$ | $6.61 \times 10^{-4}$ | $1.24 \times 10^{-1}$ | $9.15 \times 10^{-2}$ |
| $s$ | Subsistence salary per capita | $5.51 \times 10^{-5}$ | $3.67 \times 10^{-4}$ | $4.92 \times 10^{-2}$ | $2.08 \times 10^{-2}$ |
| $x_C(0)$ | Commoners population (initial value) | $-4.20 \times 10^{-5}$ | $2.01 \times 10^{-4}$ | $2.85 \times 10^{-2}$ | $2.25 \times 10^{-2}$ |
| $\varepsilon$ | Elites to commoners equilibrium ratio | $1.79 \times 10^{-4}$ | $3.18 \times 10^{-4}$ | $2.56 \times 10^{-2}$ | $1.31 \times 10^{-2}$ |
| $y(0)$ | Nature (initial value) | $2.69 \times 10^{-4}$ | $2.91 \times 10^{-4}$ | $2.06 \times 10^{-2}$ | $2.12 \times 10^{-2}$ |
| $\gamma$ | Regeneration rate of nature | $-4.72 \times 10^{-5}$ | $1.79 \times 10^{-4}$ | $1.84 \times 10^{-2}$ | $5.79 \times 10^{-3}$ |
| $\alpha_M$ | Famine (maximum) death rate | $-4.00 \times 10^{-5}$ | $7.38 \times 10^{-5}$ | $4.24 \times 10^{-3}$ | $5.82 \times 10^{-3}$ |
| $w(0)$ | Accumulated wealth (initial value) | $1.64 \times 10^{-4}$ | $3.59 \times 10^{-4}$ | $1.24 \times 10^{-3}$ | $2.12 \times 10^{-3}$ |
| $\rho$ | Threshold wealth per capita | $3.39 \times 10^{-5}$ | $3.98 \times 10^{-5}$ | $4.39 \times 10^{-4}$ | $5.08 \times 10^{-4}$ |

- **Simplified models - parameters sensitivity**

Despite the fact that parameters results were sorted by $S_T$ measure for Handy4, it can be easily observe that for other simplified models the order is quite similar (see Tab. 6.6 and Tab. 6.7). It could be caused by removal the most crucial parts of model assumptions.

However, one of the most suprising results is that the sensitivity of both Handy3 and Handy4 are almost the same - in the meaning of not only $S_T$ measure but also the $S_1$. This lead to the conclusion that simplifies labeled **A** and **B** in the Sec. 2.6 have the strongest impact on model behavior as predator-prey one.

What is even more interesting, the strongest influence in the models state has elites birth rate, which means that the sizes of populations are significant. Moreover, the system inequalities like $\kappa$ do not have impact at all.

Same as in the Handy model sensitivity analysis, the most significant variable for Handy4

(most simplified) could be find. Because amongst significant parameters (set as $S_T \geq 1.0 \times 10^{-2}$) the greatest measure values achieve parameters connected with elites population ($\beta_E$, $\alpha_M$), the $x_E$ seems to e the most important. Commoners population $x_C$ one more time apper to be the second most important variable.

Note that the first order sensitivity indices $S_1$ are negative for the same parameters across simplified models.

Table 6.6 Results of $S_T$ measures of Variance-based Sensitivity Analysis of 4 simplified Handy models in approximation to three significant figure. Ordered by $S_T$ measure for *Handy4*.

| Parameter | Description | Handy1 | Handy2 | Handy3 | Handy4 |
|---|---|---|---|---|---|
| $\beta_E$ | Elites birth rate | $9.64 \times 10^{-1}$ | $9.58 \times 10^{-1}$ | $9.61 \times 10^{-1}$ | $9.61 \times 10^{-1}$ |
| $\alpha_M$ | Famine (maximum) death rate | $9.18 \times 10^{-1}$ | $9.14 \times 10^{-1}$ | $9.16 \times 10^{-1}$ | $9.16 \times 10^{-1}$ |
| $w(0)$ | Accumulated wealth (initial value) | $3.50 \times 10^{-1}$ | $3.14 \times 10^{-1}$ | $3.18 \times 10^{-1}$ | $3.18 \times 10^{-1}$ |
| $x_C(0)$ | Commoners population (initial value) | $3.81 \times 10^{-1}$ | $3.94 \times 10^{-2}$ | $3.91 \times 10^{-2}$ | $3.91 \times 10^{-2}$ |
| $\beta_C$ | Commoners birth rate | $9.11 \times 10^{-3}$ | $8.06 \times 10^{-3}$ | $1.12 \times 10^{-2}$ | $1.12 \times 10^{-2}$ |
| $\delta$ | Depletion (production) factor | $1.14 \times 10^{-2}$ | $1.88 \times 10^{-3}$ | $2.85 \times 10^{-3}$ | $2.85 \times 10^{-3}$ |
| $\alpha_m$ | Normal (minimum) death rate | $2.39 \times 10^{-3}$ | $1.63 \times 10^{-3}$ | $2.24 \times 10^{-3}$ | $2.24 \times 10^{-3}$ |
| $\rho$ | Threshold wealth per capita | $4.13 \times 10^{-3}$ | $1.49 \times 10^{-3}$ | $1.86 \times 10^{-3}$ | $1.86 \times 10^{-3}$ |
| $y(0)$ | Nature (initial value) | $9.51 \times 10^{-4}$ | $7.59 \times 10^{-4}$ | $9.95 \times 10^{-4}$ | $9.95 \times 10^{-4}$ |
| $\varepsilon$ | Elites to commoners equilibrium ratio | $6.05 \times 10^{-4}$ | $4.77 \times 10^{-4}$ | $7.15 \times 10^{-4}$ | $7.15 \times 10^{-4}$ |
| $x_E(0)$ | Elites population (initial value) | $5.97 \times 10^{-3}$ | $7.45 \times 10^{-5}$ | $5.69 \times 10^{-4}$ | $5.69 \times 10^{-4}$ |
| $s$ | Subsistence salary per capita | $8.89 \times 10^{-8}$ | $3.35 \times 10^{-8}$ | $3.35 \times 10^{-8}$ | $3.35 \times 10^{-8}$ |
| $\gamma$ | Regeneration rate of nature | $2.22 \times 10^{-20}$ | $7.41 \times 10^{-21}$ | $4.30 \times 10^{-21}$ | $9.13 \times 10^{-21}$ |
| $\kappa$ | Inequality factor | $0.00 \times 10^{0}$ | $0.00 \times 10^{0}$ | $0.00 \times 10^{0}$ | $0.00 \times 10^{0}$ |
| $\lambda$ | Nature carrying capacity | $0.00 \times 10^{0}$ | $0.00 \times 10^{0}$ | $0.00 \times 10^{0}$ | $0.00 \times 10^{0}$ |

Table 6.7 Results of $S_1$ measures of Variance-based Sensitivity Analysis of 4 simplified Handy models in approximation to three significant figure. Ordered by $S_T$ measure for *Handy4*.

| Parameter | Description | Handy1 | Handy2 | Handy3 | Handy4 |
|---|---|---|---|---|---|
| $\beta_E$ | Elites birth rate | $1.13 \times 10^{-1}$ | $1.21 \times 10^{-1}$ | $1.20 \times 10^{-1}$ | $1.20 \times 10^{-1}$ |
| $\alpha_M$ | Famine (maximum) death rate | $6.13 \times 10^{-2}$ | $6.38 \times 10^{-2}$ | $6.36 \times 10^{-2}$ | $6.36 \times 10^{-2}$ |
| $w(0)$ | Accumulated wealth (initial value) | $3.14 \times 10^{-3}$ | $2.82 \times 10^{-3}$ | $2.83 \times 10^{-3}$ | $2.83 \times 10^{-3}$ |
| $x_C(0)$ | Commoners population (initial value) | $1.05 \times 10^{-3}$ | $9.31 \times 10^{-4}$ | $9.45 \times 10^{-4}$ | $9.45 \times 10^{-4}$ |
| $\beta_C$ | Commoners birth rate | $-5.24 \times 10^{-4}$ | $-4.25 \times 10^{-4}$ | $-5.48 \times 10^{-4}$ | $-5.48 \times 10^{-4}$ |
| $\delta$ | Depletion (production) factor | $-5.03 \times 10^{-5}$ | $-2.91 \times 10^{-6}$ | $-7.02 \times 10^{-7}$ | $-7.02 \times 10^{-7}$ |
| $\alpha_m$ | Normal (minimum) death rate | $-1.06 \times 10^{-4}$ | $-8.74 \times 10^{-5}$ | $-9.60 \times 10^{-5}$ | $-9.60 \times 10^{-5}$ |
| $\rho$ | Threshold wealth per capita | $-1.36 \times 10^{-4}$ | $-9.53 \times 10^{-5}$ | $-1.04 \times 10^{-4}$ | $-1.04 \times 10^{-4}$ |
| $y(0)$ | Nature (initial value) | $-5.42 \times 10^{-5}$ | $-4.18 \times 10^{-5}$ | $-4.55 \times 10^{-5}$ | $-4.55 \times 10^{-5}$ |
| $\varepsilon$ | Elites to commoners equilibrium ratio | $-2.26 \times 10^{-5}$ | $-2.14 \times 10^{-5}$ | $-2.63 \times 10^{-5}$ | $-2.63 \times 10^{-5}$ |
| $x_E(0)$ | Elites population (initial value) | $3.45 \times 10^{-4}$ | $3.23 \times 10^{-5}$ | $5.42 \times 10^{-5}$ | $5.42 \times 10^{-5}$ |
| $s$ | Subsistence salary per capita | $-7.39 \times 10^{-7}$ | $-2.40 \times 10^{-7}$ | $-2.41 \times 10^{-7}$ | $-2.41 \times 10^{-7}$ |
| $\gamma$ | Regeneration rate of nature | $1.47 \times 10^{-14}$ | $1.10 \times 10^{-14}$ | $3.13 \times 10^{-14}$ | $1.26 \times 10^{-14}$ |
| $\kappa$ | Inequality factor | $0.00 \times 10^{0}$ | $0.00 \times 10^{0}$ | $0.00 \times 10^{0}$ | $0.00 \times 10^{0}$ |
| $\lambda$ | Nature carrying capacity | $0.00 \times 10^{0}$ | $0.00 \times 10^{0}$ | $0.00 \times 10^{0}$ | $0.00 \times 10^{0}$ |

## 6.5 Approximate Bayesian Computation on sensitive parameters

### 6.5.1 Idea

During the research process, we were considering various possibilities for the usage of the preknowledge about sensitivity of Handy parameters. The very first proposition that could possibly reduce the prediction errors is a longer learning time for the most sensitive parameters. The previous statement means that we could pretrain model on all parameters and then learn only the most influential. Consequently, we will have a problem with lower dimensionality than the initial one. In this thesis, we present a methodology to reduce the further dimensionality to be almost two times smaller (we have shown that the Handy initial parameters could be split into: sensitive - 8 parameters and insensitive - 7). Furthermore, for the pretrained models we assumed assumed that the achieved initial parameters are close to their real initial values. Because of that, we decided to minimize the intervals for the sensitive parameters to be $\pm 5\%$ of the pretrained value. The most significant advantage of using such method will be lowering the dimensionality of the problem. Moreover, the ABC method should get an inevitable speeding for the second phase of the learning process - learning sensitive parameters. However, the process of finding the appropriate time split between pretraining and learning sensitive parameters has been considered by us as the most important part of this methodology.

In fact, the Handy parameters were divided into two separable groups with respect to their sensitivity (as it was shown in the Tab. 6.5). Moreover, we decided to test this methodology for three various times $t_{pretrained}$ for the pretraining process. We obtained these times $t_{pretrained}$ from the Tab. 6.2. Each of them was the mean time $\bar{t}$ after that the ABC method learned the model to the certain RMSE: 10.0, 5.0, 2.0. Obviously, we were sure that usage of such times should effect with the models learned to corresponding errors. We prepared the test set in a similar way than for the ABC method. Due to above assumption, models were being learned with the ABC in the following way:

1. Generation of the Ground Truth model (GT) from the Handy model.

2. Repeating the tests for the sampling frequencies $f_i$ from the set $\{15, 50\}$:

    (a) Sampling observations data from the GT with the frequency $f_i$ (from the learning part, $t_{start} = 150$, $t_{end} = 300$).

    (b) Creating intervals for each of the model parameters as $\pm 10\%$ of exact the value.

    (c) Repeating 10 times the learning procedure:

        i. Learning model with the ABC method on **all** parameters for the time $t_{pretraining}$ (Learning model to certain RMSE: 10.0, 5.0 or 2.0).

        ii. Selecting the initial parameters from the pretrained model:
            • initial values for *insensitive parameters* - take from the pretrained model results
            • initial values for *sensitive parameters* - creating new intervals for these parameters ($\pm 5\%$ of their value from pretrained model)

        iii. Learning model with the ABC method with respect sensitive parameters (insensitive parameters have fixed values) for the rest of the time: $t_{sensitive} = t_{total} - t_{pretraining}$.

        iv. Generation the forward and backward predictions from the learned model. Computing the forecasting errors.

Due to the fact that we had created the reference models from the sets of only 8 repetitions, we decided to limit either the sets of the results for this methodology. Because learning time $t$ was almost equal for each of the repetitions, we proposed to remove the quantiles of 0.05 and 0.95 of the forward prediction errors ($ff$). In fact, such procedure helped to decrease the number of analysed samples also to 8.

### 6.5.2   Results

Results of the ABC method with respect to parameters' sensitivity were presented in the following Tab. 6.8. Note that the computed measures are the same as for the ABC method. However, the only difference is in the *Initial RMSE* column - we marked the learning error for which the models were pretrained. Firstly, we observed that the overall learning times were similar for all the repetitions. Secondly, the mean times ($\bar{t}$) are between 8-10 seconds longer than the mean times of reference models. Nevertheless, we have found the procedure of turning the ABC off after certain time as the main reason for these results. However, the mentioned time difference are insignificant for the process.

Table 6.8 Results for the ABC on sensitive parameters in comparison with reference models. The best models were marked bold.

| Initial RMSE | $\bar{t}$[s] | $\overline{t_{std}}$[s] | $\frac{\overline{t_{std}}}{\bar{t}}$ | $\overline{ff}$ | $\overline{ff_{std}}$ | $\frac{\overline{ff_{std}}}{\overline{ff}}$ | $\overline{f2w}$ | $\overline{f2w_{std}}$ | $\overline{fb}$ |
|---|---|---|---|---|---|---|---|---|---|
| | | | *frequency sampling $f = 15$* | | | | | | |
| 10.0 | 376.57 | 2.151 | 0.006 | 15.08 | 8.76 | 0.581 | 187.54 | 124.48 | 264.79 |
| **5.0** | **378.22** | **9.761** | **0.03** | **24.58** | **11.54** | **0.47** | **47.04** | **20.41** | **61.82** |
| 2.0 | 376.01 | 4.14 | 0.01 | 98.89 | 45.00 | 0.46 | 130.71 | 27.29 | 156.18 |
| *Reference model* | 368.25 | 42.00 | 0.11 | 2.50 | 1.03 | 0.41 | 17.11 | 10.20 | 24.07 |
| | | | *frequency sampling $f = 50$* | | | | | | |
| **10.0** | **435.38** | **6.74** | **0.02** | **12.63** | **5.70** | **0.45** | **63.17** | **27.54** | **88.44** |
| 5.0 | 434.34 | 5.84 | 0.02 | 46.21 | 15.54 | 0.34 | 123.12 | 104.47 | 167.87 |
| 2.0 | 434.85 | 6.08 | 0.01 | 58.56 | 36.22 | 0.62 | 105.56 | 92.92 | 137.32 |
| *Reference model* | 426.72 | 33.10 | 0.08 | 3.64 | 1.62 | 0.46 | 18.66 | 11.33 | 26.14 |

Afterward, we have found several propositions that could be the consequences of the results:

1. Application the preknowledge about sensitivity in the proposed way did not improve the ABC results and also the Sensitivity Analysis computations were needed.

2. We have observed the correlation between longer learning only sensitive parameters and the better results. Moreover, even if the models were less pretrained on all parameters, they had better results with a longer and an adequate time of learning sensitive parameters.

3. In the process of preparing initial parameters for the second part of learning, we set the intervals for sensitive parameters as $\pm 5\%$ of the pretrained value. Consequently, we found that the intervals were still too large and had the strongest impact on the magnitude of the prediction errors.

4. The ABC method on sensitive parameters has worse results than reference models.

We have recognised that the models pretrained to the RMSE of a value 5.0 (for the sampling frequency $f = 15$) and to RMSE of a value 10.0 (for the sampling frequency $f = 10.0$) were the best. This decision had been made after comparison of the $\overline{f2w}$ values. Moreover, for these models we presented the mean values for the initial parameters and variables in the Tab. 6.9. It has to be said that we also splited the initial parameters into two separable groups: *sensitive* and *insensitive* as in the Tab. 6.5.

Table 6.9 Mean values for the parameters and the variables achieved by the ABC on sensitive parameters method. The results for two various sampling frequencies and the Ground Truth (GT).

| Parameter | Description | GT value | Frequency | | | |
|---|---|---|---|---|---|---|
| | | | 15 | | 50 | |
| | | | mean | std | mean | std |
| *sensitive parameters* | | | | | | |
| $\beta_C$ | Commoners birth rate | $6.50 \times 10^{-2}$ | $6.46 \times 10^{-2}$ | $1.03 \times 10^{-3}$ | $6.54 \times 10^{-2}$ | $1.38 \times 10^{-3}$ |
| $\beta_E$ | Elites birth rate | $2.00 \times 10^{-2}$ | $1.98 \times 10^{-2}$ | $5.18 \times 10^{-4}$ | $1.97 \times 10^{-2}$ | $6.63 \times 10^{-4}$ |
| $\alpha_m$ | Normal (minimum) death rate | $1.00 \times 10^{-2}$ | $9.79 \times 10^{-3}$ | $5.60 \times 10^{-4}$ | $9.67 \times 10^{-3}$ | $7.23 \times 10^{-4}$ |
| $\delta$ | Depletion (production) factor | $2.04 \times 10^{0}$ | $2.03 \times 10^{0}$ | $3.71 \times 10^{-2}$ | $2.06 \times 10^{0}$ | $8.01 \times 10^{-2}$ |
| $\kappa$ | Inequality factor | $1.00 \times 10^{1}$ | $9.99 \times 10^{0}$ | $1.97 \times 10^{-1}$ | $1.00 \times 10^{1}$ | $7.24 \times 10^{-2}$ |
| $x_E(0)$ | Elites population (initial) | $3.00 \times 10^{3}$ | $2.99 \times 10^{3}$ | $4.82 \times 10^{1}$ | $2.99 \times 10^{3}$ | $8.19 \times 10^{1}$ |
| $\lambda$ | Nature carrying capacity | $1.00 \times 10^{2}$ | $9.99 \times 10^{1}$ | $1.97 \times 10^{-1}$ | $9.86 \times 10^{1}$ | $2.96 \times 10^{0}$ |
| $s$ | Subsistence salary per capita | $5.00 \times 10^{-4}$ | $5.00 \times 10^{-4}$ | $5.60 \times 10^{-4}$ | $4.92 \times 10^{-4}$ | $1.80 \times 10^{-5}$ |
| *insensitive parameters* | | | | | | |
| $x_C(0)$ | Commoners population (initial) | $1.00 \times 10^{4}$ | $9.96 \times 10^{3}$ | $9.98 \times 10^{1}$ | $9.89 \times 10^{3}$ | $3.18 \times 10^{2}$ |
| $\mu$ | Elites to commoners equilibrium | $6.50 \times 10^{-1}$ | $6.55 \times 10^{-1}$ | $1.70 \times 10^{-2}$ | $6.36 \times 10^{-1}$ | $2.94 \times 10^{-2}$ |
| $y(0)$ | Nature (initial) | $1.00 \times 10^{2}$ | $1.00 \times 10^{2}$ | $4.95 \times 10^{0}$ | $9.87 \times 10^{1}$ | $5.23 \times 10^{0}$ |
| $\gamma$ | Regeneration rate of nature | $1.00 \times 10^{-2}$ | $1.00 \times 10^{-2}$ | $1.77 \times 10^{-4}$ | $1.00 \times 10^{-3}$ | $2.90 \times 10^{-4}$ |
| $\alpha_M$ | Famine (maximum) death rate | $7.00 \times 10^{-2}$ | $6.96 \times 10^{-2}$ | $1.08 \times 10^{-3}$ | $7.05 \times 10^{-2}$ | $1.47 \times 10^{-3}$ |
| $w(0)$ | Accumulated wealth (initial) | $1.00 \times 10^{2}$ | $1.02 \times 10^{2}$ | $3.45 \times 10^{0}$ | $1.01 \times 10^{2}$ | $3.38 \times 10^{0}$ |
| $\rho$ | Threshold wealth per capita | $5.00 \times 10^{-3}$ | $5.00 \times 10^{-3}$ | $9.40 \times 10^{-5}$ | $4.94 \times 10^{-3}$ | $2.13 \times 10^{-4}$ |

Let us notice that the initial parameters are better estimated in the model with the sparser sampling (the sampling frequency $f = 15$). We supposed that it was caused by the fact that the best results were for the sampling frequency if a value equal to 15 (as shown in the Tab. 6.2 and in the Fig. 6.7 and 6.8). What is more, we have observed that the better estimation of the initial values were obtained for the insensitive parameters.

Below (in the Fig. 6.10) we have shown the prediction of the *elites population* for the sampling frequency $f = 15$. The prediction is visualized in a normal form and also normalized form. Evenmore, we highlighted the standard deviation of the value in each point with grey colour. The results for the rest of variables and for the denser sampling are presented in the Appendix A (Fig. A.9 - A.16).

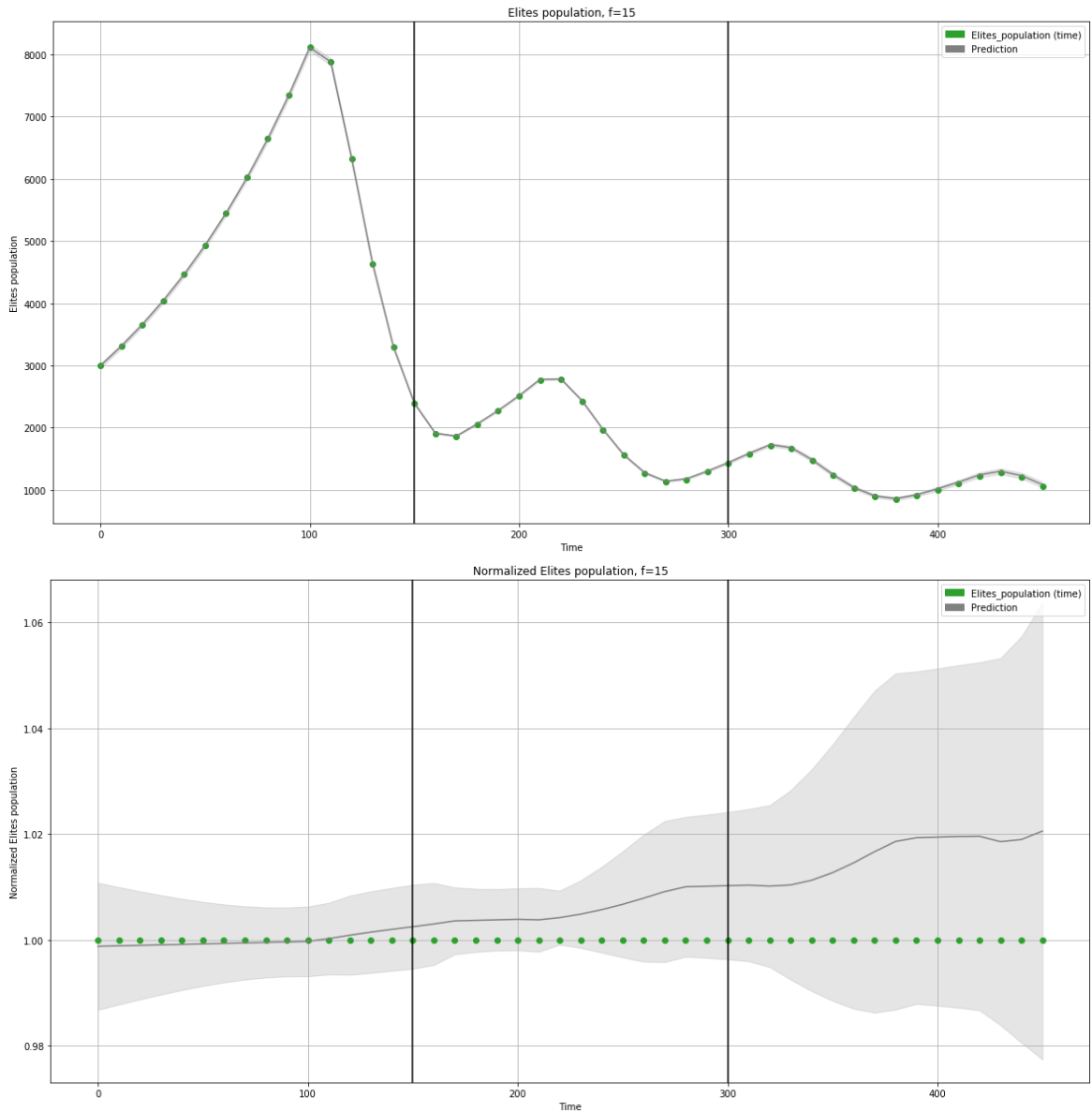

Fig. 6.10 ABC on sensitive parameters results - elites population (both real and normalized), *f = 15*

The analysis of the results has forced us to the following conclusions:

- We have demonstrated that the best model for sparser sampling case ($f = 15$) converged faster on the chaotic part and that the forward prediction errors were larger than on the reference models.

- The ABC on the sensitive parameters method gives quite good prediction errors. However, the errors are usually much higher than for the reference models (the simple ABC method) in the similar amount of time. What is more, this method requires the knowledge about parameters' sensitivities and in fact much longer computations.

## 6.6   Supermodel synchronized via the most sensitive variable

Another approach, which could help to enhance prediction results (that are obtained in the same time) seemed to be the methodology of the Supermodeling. Although, this method is being widely used in the community of chaotic systems, it is almost not known in the field of the data assimilation. One of the main fields of usage chaotic models is climatology. The weather forecasting in a large scale (in a system that contains the earth and it's numerous impacts like oceans or winds) could not be modelled mathematically or even very precisely. Due to that fact, considering the methods that connect some imperfect models is necessary. The well-known approach to deal with such applications is called the Supermodeling and was briefly described in the Ch. 4.

### 6.6.1   Idea

We have demonstrated that the ABC method (reference models) provides good prediction results for the phenomena that behave in a *smooth* way - e.g. not so chaotic as the case of the backward forecasting in the considered GT model. However, there are several difficulties for the usage of the ABC method:

1. Worse assimilations and predictions on the chaotic models (e.g. backward forecasting in the selected GT model).

2. We have presented that the mean time to learn model perfectly is much longer than to learn it to be imperfect model (see the results in the Tab. 6.2).

3. Moreover, we have found that the possibility of stucking in a local minimum is higher if the model is learn to very small RMSE (e.q. 1.0).

4. The decision about the most adequate sampling frequency could be complicated. For example, at some situations the researchers should select not all of the Ground Truth points, because the sparser sampling could be more effective. However, we have thought if the other methodologies could minimize the impact of the inappropriate decision.

In order to deal with the above difficulties, we have proposed to use the methodology of the Supermodeling. Firstly, the major pros of this method is well-known ability to cope with the chaotic systems. Secondly, we have been convinced that synchronization of the submodels could be a possibility to avoid stucking in a local minimum. Furthermore, the Supermodel is usually composed of several good but imperfect models, which imply that the submodels could be learned paralelly. Finally, we proposed to pretrain a few submodels paralelly with the ABC method to a certain error (see the mean times presented in the Tab. 6.2) and then learn only the Supermodel's coupling coefficients for the rest of the time.

However, even without *nudging*, the Supermodeling methodology could led to be a high-dimensional problem. We computed that the fully connected Supermodel required the number of the coupling coefficients between submodels that was equal to $v\binom{n}{2}$, where $v$ was a number of the variables and $n$ - the number of submodels. What is more, we proposed to minimize the time of learning the Supermodel by a minimization of the following complexity measure:

$$\mu = \frac{S_{cc}}{H_p} \tag{6.2}$$

where $S_{cc}$ - the number of Supermodel's coupling coefficients and $H_p$ - the number of HANDY's initial parameters (the size of the inital problem).

We have decided to use only 3 submodels, which has meant that the fully connected Supermodel would be synchronized via $4\binom{3}{2} = 12$ parameters. However, we have shown that in such case the number of the Supermodel's coefficients was comparable to the number of the HANDY's parameters ($\mu = 0.8$) and in fact, did not have a great influence on minimization of the measure $\mu$. Due to the above observation, we found the preknowledge from the Sensitivity Analysis to be beneficial. As we have demonstrated in the Sec. 6.4, the most significant variable is *elites population $x_E$*. Moreover, the synchronization of the submodels via only one variable provides the smallest possible value of the complexity measure $\mu$, which is equal to 0.2. Please observe that the submodels connected via insensitive variables could not be synchronized. Because of the fact that we have decided to use only 3 coupling coefficients, the Supermodeling methodology could be faster than the ABC method (on all 15 parameters). The synchronization of the Supermodel via only a part of variables for similar application was presented in the Fig. 6.11.

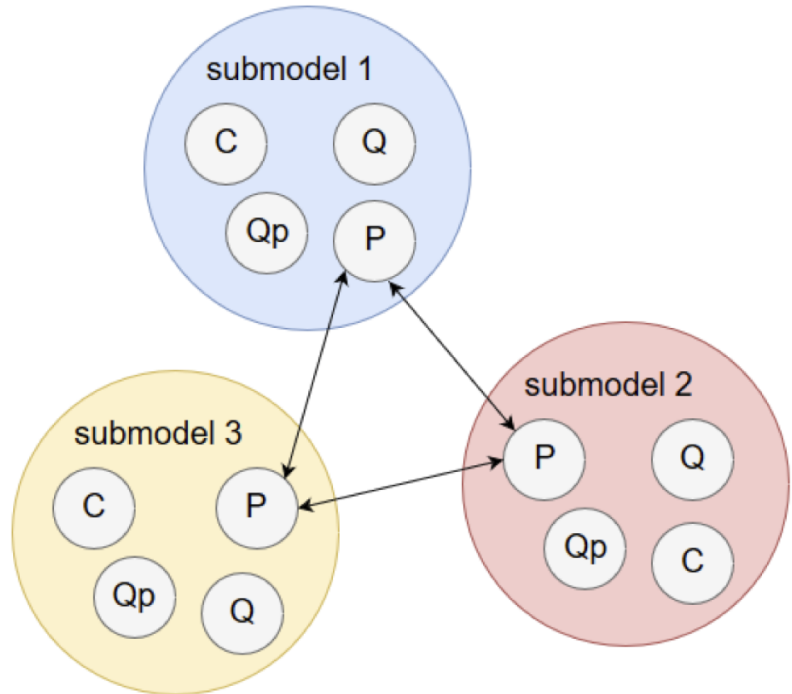

Fig. 6.11 Synchronization the Supermodel via only a part of variables. Illustration have been taken from the presentation of Dzwinel et al. to be given at SIAM Conference on Applications of Dynamical Systems, May 19-23, 2019. First appeared in the master's thesis of Pedrycz, *Approximate Bayessian Computing in data adaptaion of tumor dynamics model*, AGH, 2018 [47]

According to the previously mentioned assumptions, we have proposed the Supermodeling methodology in the version without *nudging* method in order to minimize the problem dimensionality. We have created the Supermodel by synchronization of 3 submodels that were connected by the most sensitive variable - *elites population $x_E$*. The rest of the methodology parameters had the following parameters for each of the considered versions:

- The coupling coefficients intervals were equaled to $[0, 0.5]$ (according to the tested values in Dzwinel et al. [19]).

- We have used the error measure given by the Eq. 4.6.

- We splited the learning part into 5 same-length intervals, which means that the value of the parameter $K$ was equal to 5.

- The value of the initial factor $\gamma$ was equal to 0.5.

What is more, we have decided to use the ABC method to learn both submodels and the Supermodel. The last assumption has allowed us to keep the methodology consistent. Each of the submodels was learned up to the one of the following RMSEs: 10.0, 5.0 or 2.0 (accrding to the results shown in the Tab. 6.2). Although, the process of learning the submodels was iterative, we recommend to perform it parallely. That is why, we have decided to simulate the situation of the parallel learning the submodels by taking the mean value of learning the set of 3 submodels, $\overline{t_{submodels}}$.

We have learned the Supermodel for the rest of the available time. Consequently, we have calculated the time for learning the Supermodel in a following way: $t_{Supermodel} = t_{max} - \overline{t_{submodels}}$, where $t_{max}$ is once again a mean time to achieved the reference model. Moreover, we have presented the test case for the proposed methodology. In fact, it looks in tha same way for each of the reqested submodels RMSEs:

1. Generation of the Ground Truth model (GT) from the Handy model.

2. Repeating the tests for sampling frequencies $f_i$ from the set $\{15, 50\}$:

   (a) Sampling observations data from the GT with the frequency $f_i$ (from the learning part, $t_{start} = 150$, $t_{end} = 300$).

   (b) Creating intervals for each of the model parameters as $\pm 10\%$ of the exact value.

   (c) Repeating 10 times the learning procedure:

      i. Learning 3 submodels with the ABC method, each of them in a time $t_{submodel}$ (Pretraining submodels to a certain RMSE: 10.0, 5.0 or 2.0).

      ii. Creating the Supermodel by synchronization the most sensitive variables *elites population $x_E$* of each of the submodels.

      iii. Learning the Supermodel for the $t_{Supermodel}$, which is computed as $t_{Supermodel} = t_{max} - \overline{t_{submodels}}$.

      iv. Generation the forward and backward predictions from the learned Supermodel. Computing the forecasting errors.

Let us remind that the reference models were compound of the sets of only 8 repetitions. Due to the above, we had to either select the sets of 8 models from the Supermodeling results. Furthermore, we have decided to perform the procedure of removing the best and the worst model in reference to the forward prediction errors $ff$. In addition, this procedure has been performed in the same way as in the Sec. 6.5. In fact, we reduced the number of the samples in each of the test sets to 8 samples.

## 6.6.2   Results

We have presented the results of the Supermodeling methodology in the Tab. 6.10. Analogous to the ABC method on sensitive parameters, we marked bold the best model for each of the sampling frequencies. Please note that the selection of the best models was obvious. Moreover, in the column *RMSE (for pretraining)* we presented the approximated initial error of the submodels. On the other hand, we have decided to abandon the presentation of $\overline{t_{std}}$ and $\overline{\frac{t_{std}}{t}}$. Instead, we proposed to demonstrate the time for learning Supermodel $t_{sumo}$ as well as the normalized time measure $t_{norm} = \overline{t_{submodels}} + t_{sumo}$ (based on possibility of the parallel learning submodels). We have been aware of the disadvantages of the computing the $\overline{t_{submodels}}$, however we have not got the information about the times for learning the particular submodel. What is more, we have demonstrated the mean values of the proposed measures ($\overline{t_{norm}}$, $\overline{t_{sumo}}$). Obviously, we have identified that the normalized time $\overline{t_{norm}} = \bar{t}$ for the reference models.

Table 6.10 Forecasting results for the Supermodeling with submodels learned in the same time.

| RMSE (for pretraining) | $\bar{t}$[s] | $\overline{t_{norm}}$[s] | $\overline{t_{sumo}}$[s] | $\overline{ff}$ | $\overline{ff_{std}}$ | $\overline{\frac{ff_{std}}{ff}}$ | $\overline{f2w}$ | $\overline{f2w_{std}}$ | $\overline{fb}$ |
|---|---|---|---|---|---|---|---|---|---|
| | | | *frequency sampling $f = 15$* | | | | | | |
| 10.0 | 690.95 | 383.86 | 230.32 | 55.99 | 21.01 | 0.38 | 119.09 | 39.69 | 158.84 |
| 5.0 | 891.88 | 382.47 | 127.77 | 20.93 | 11.11 | 0.53 | 38.92 | 22.80 | 50.91 |
| **2.0** | **1025.51** | **386.50** | **66.99** | **5.95** | **3.15** | **0.53** | **29.18** | **10.34** | **40.84** |
| *Reference model* | 368.25 | 368.25 | – | 2.50 | 1.03 | 0.41 | 17.11 | 10.20 | 24.07 |
| | | | *frequency sampling $f = 50$* | | | | | | |
| 10.0 | 704.68 | 387.90 | 229.51 | 78.98 | 48.15 | 0.61 | 117.28 | 41.36 | 145.85 |
| 5.0 | 906.87 | 388.42 | 129.20 | 15.62 | 7.35 | 0.47 | 40.85 | 23.64 | 55.62 |
| **2.0** | **1030.11** | **383.51** | **60.21** | **11.08** | **4.41** | **0.40** | **34.88** | **12.94** | **48.07** |
| *Reference model* | 426.72 | 426.72 | – | 3.64 | 1.62 | 0.46 | 18.66 | 11.33 | 26.14 |

The best predictions were achieved by the synchronization of the most pretrained submodels. However, we also observed that the Supermodeling forecasting errors were worse than the reference models. Moreover, we have achieved better predictions for each of the submodels separately than for the Supermodels comounded of these submodels (please, compare Tab. 6.10 and 6.2). In other words, we have demonstrated that the Supermodeling with the similar submodels is worse than the ABC method. What is more, we have found several propositions that could be the consequences of these results:

1. The Supermodeling is much better application of the sensitivity preknowledge than presented ABC on sensitive parameters. It could be easily found by a comparison of the results.

2. We have observed that only the $\overline{f2w}$ predictions were better for the Supermodels than for their submodels.

3. We proposed the possible explanation that the reason for such results is the fact that the submodels were very similar to each other. Even more, we have decided to make an analysis of the submodels' similarity.

Furthermore, we have demonstrated the prediction of the *elites population* for the sampling frequency $f = 15$ (in the Fig. 6.12). The prediction is visualized in a normal form and also normalized form. Evenmore, we highlighted the standard deviation of the value in each point with grey colour.

The results for the rest of variables and for the denser sampling are presented in the Appendix A (Fig. A.17 - A.24).

The analysis of the results has forced us to the following conclusions:

- We have observed that the forecasting with the Supermodeling were really close to the GT's attractor.

- Moreover, the Supermodel converged faster than the reference model.

- In addition, the prediction of the most sensitive variable (the variable, which was responsible for synchronization) has very small errors.

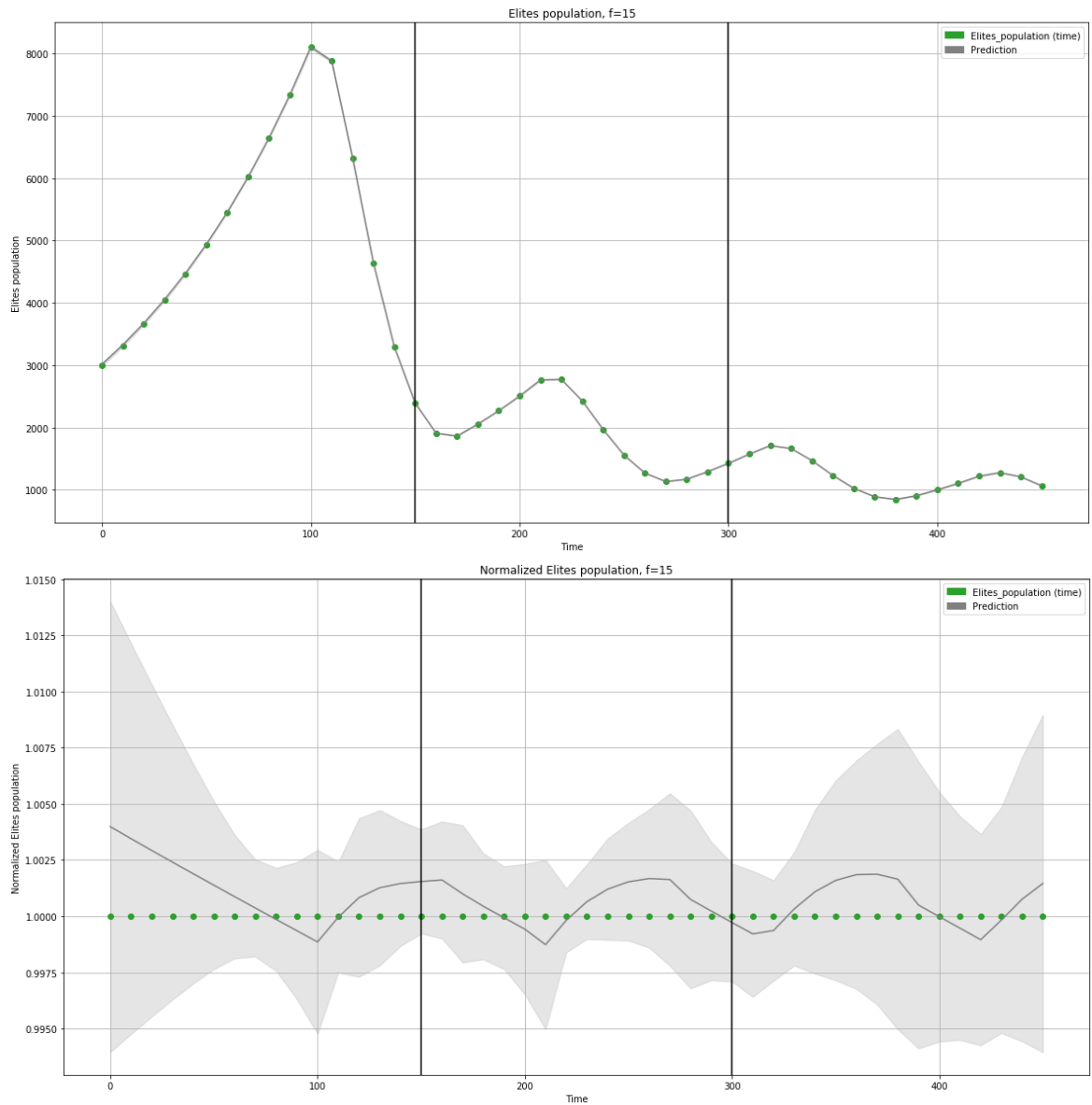

Fig. 6.12 Supermodeling with similar submodels results - elites population (both real and normalized), *f = 15*

## 6.7    Analysis of the submodels similarity

Because we have demonstrated that the results of the Supermodeling methodology with similar submodels were worse than the reference models, we decided to check if the submodels were not too similar to each other. In fact, we have shown even that the Supermodeling predictions were worse than the ABC method. Furthermore, we considered the parameters' space $S_p$ and it was obvious that the submodels could not be too close one another. Otherwise, the submodels were not able to converge to the Ground Truth, because mostly the GT was not between them. Due to that fact, we decided that the performance of a study of the submodels' relative positions was necessary. We have proposed to use the well-known method of the *Principal Components Analysis* (PCA) [65] to the mentioned analysis. Moreover, the PCA method helped to reduce the $S_p$ space's dimensionality and allowed for visualization the GT as well as submodels in 2-dimensional space.

Afterward, we have picked 8 sets of pretrained submodels from the results of the Supermodeling methodology with the value of sampling frequency $f$ equaled to 15. The mentioned sets of the submodels could be splitted into two groups: 4 sets were compound of the submodels pretrained to the error of an approximate value equaled to 2.0 (for 307.52s, see Tab. 6.2) and the rest of the sets were learned to the error of 5.0 (for 242.66s, see Tab. 6.2). However, each of these groups consisted of the following sets of submodels (with respect to the overall $f2w$ Supermodeling errors):

- **the best set of submodels** - the submodels that synchronized to the Supermodel, which achieved the lowest RMSE for forward prediction; the Supermodel that was not included in the reduced set of results

- **the worst set of submodels** - the submodels that synchronized to the Supermodel, which achieved the highest RMSE for forward prediction; the Supermodel that was not included in the reduced set of results

- **two other sets of submodels**

We have shown results for the PCA method in the following graphs - Fig. 6.14 and Fig. 6.15. The figures contain the results respectively for better pretrained submodels (to an error of a value 2.0) and for less pretrained (to an error of a value 5.0). Please, notice that each set of the submodels was colored with the same tone of the main colour group.

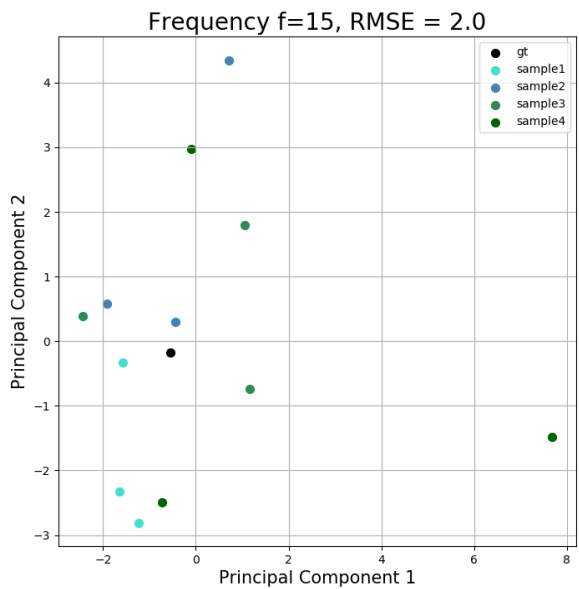

Fig. 6.13 PCA (2 components) results for 4 sets of the submodels pretrained in the same time (to an error $\leq 2.0$) for sampling frequency $f = 15$. Ground Truth is black, each submodel from a specific set has the same colour.

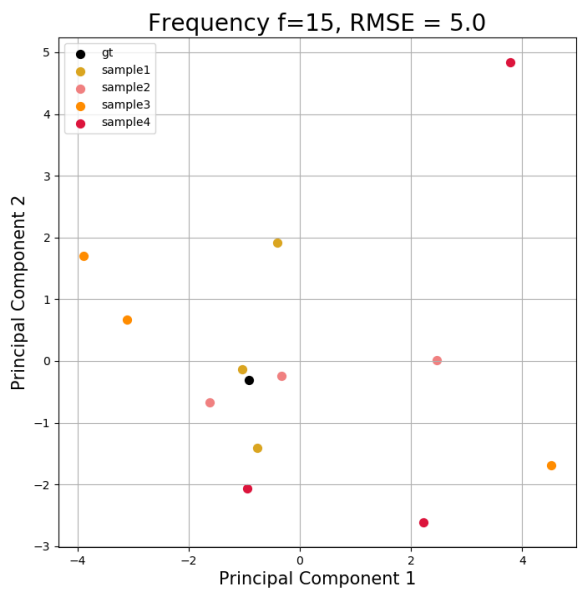

Fig. 6.14 PCA (2 components) results for 4 sets of the submodels pretrained in the same time (to an error $\leq 5.0$) for sampling frequency $f = 15$. Ground Truth is black, each submodel from a specific set has the same colour.

We have observed that in a large number of cases the sets of the longer learned submodels were localized in the groups, however the GT was not between them. Furthermore, submodels could converged to the same or similar local minima, which were different from the GT. In addition, it was obvious that the Supermodel consisted of such submodels could not converged to the GT.

On the other hand, the PCA results for the less pretrained submodels were more optimistic than the previous. We have found that distances between submodels and the GT were greater than for the better submodels. However, the sets of submodels also formed groups, but *around* the GT value. Consequently, we supposed that such behaviour could have meant that any of the submodels did not stuck in a local minimum and the Supermodel would converge to the GT in a longer learning time.

Finally, in order to have better understanding of the submodels' relative positions, we decided to present all of the examined submodels in the common two-dimensional space with GT value (in the Fig. 6.15). The dimensionality reduction was achieved with the PCA method. However, we changed the colour scale - all of the similar submodels (with respect to a learning time) were coloured in the same way: blue for the longer and red for the shorter learning time. We have found that many times the submodels formed groups (pairs) around the same local minimum. Moreover, groups were mostly composed of the submodels learned for the same time. For this reason, we have planned to learn the submodels for different times and possibly got the submodels that converged to the various local minima.

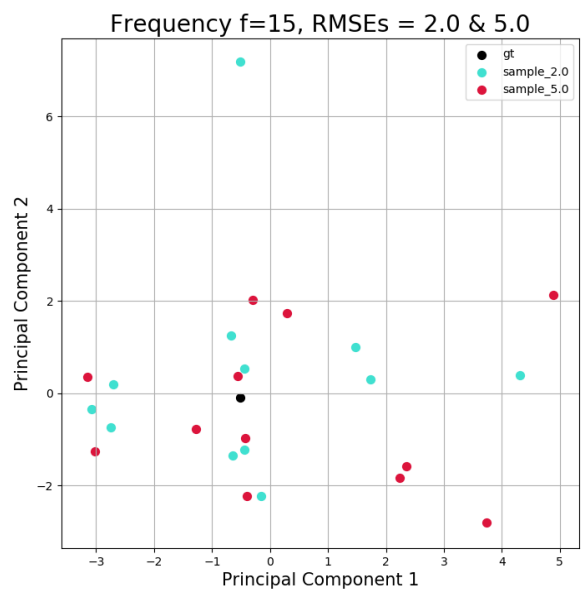

Fig. 6.15 PCA (2 components) of a few sets of submodels learned with different times - half to the RMSE lower than 2.0 (blues) and the rest to the RMSE lower than 5.0 (reds) for sampling frequency $f = 15$. Ground Truth value is black.

The above observations have implied the proposition to form the Supermodel with the submodels that were learned for different times. Due to the above concept, we have predicted that the less learned submodel could help to avoid converging to any local minimum. The main reasons for this were the longer distances between submodels in a parameters's space $S_p$ (and values's space in a specific timestamp) that could have *pulled out* the Supermodel from a local minimum. What is more, we have found also that composing the Supermodel from only fully learned submodels would inevitably provide to converge into a local minimum. Finally, after this analysis, we understood that the major advatage (and the initial idea) of the Supermodeling have been lost. Moreover, the Supermodel would converge to a local minimum that is worse than it's submodels.

Obviously, similar results could be achieved also when the value of the sampling frequency $f$ is equal to 50.

## 6.8 Supermodeling with different submodels

### 6.8.1 Idea

We have demonstrated that the Supermodeling methodology could achieved better predictions for the more chaotic systems, but the submodels should not be pretrained to the same error, because of the possibility of stucking in a local minimum. According to the previously performed analysis of the submodels' similarity, we have decided to use various submodels (less and more pretrained) to improve the Supermodels' synchronizations.

What is more, the expected behaviour of the submodels has been successfully presented in the Fig. 6.16. In the mentioned figure, two submodels were presented, each of them was learned and was converging into different attractor. We were strongly convinced that it was possible to get another (closer to the GT) attractor by the process of a synchronization.

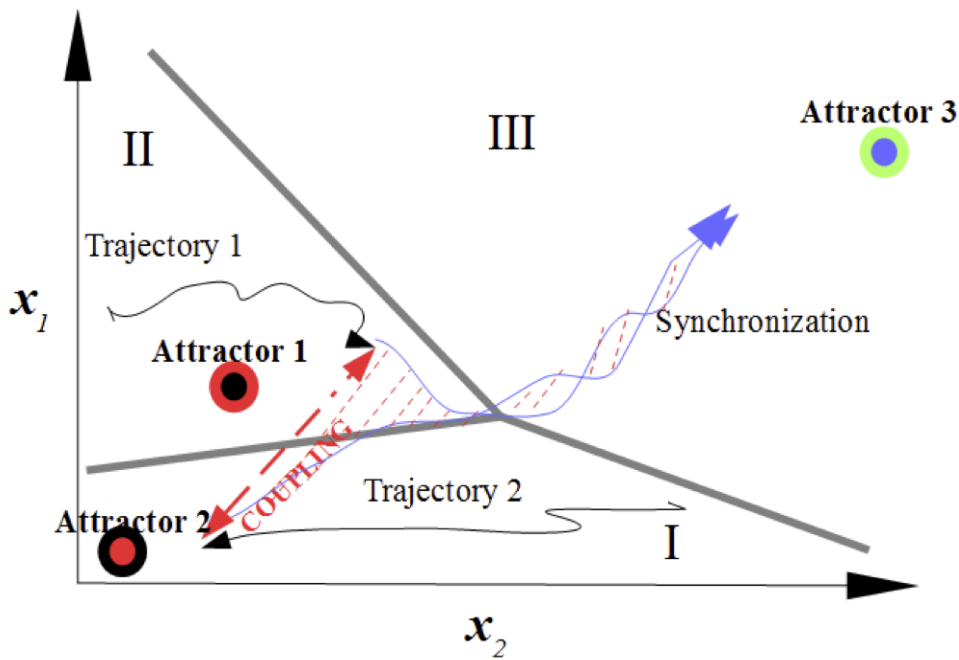

Fig. 6.16 Converging of the Supermodel based on various submodels. Illustration taken from presentation of Dzwinel et al. to be given at SIAM Conference on Applications of Dynamical Systems, May 19-23, 2019.

Taking under consideration that the submodels learned to an error of a value close to 10.0 gave worse predictions than the longer pretrained submodels, we decided to compose the Supermodels only via synchronization of the submodels pretrained to the errors of values:

2.0 and 5.0. However, there were only two symmetrical possibilities of composing the Supermodel with these assumptions - two similar submodels and one lerned to the different error. In fact, the Supermodels have been compunded of:

- 2 submodels pretrained to the learning error of a value 2.0 & 1 submodel pretrained to th error of a value 5.0

- 2 submodels pretrained to the learning error of a value 5.0 & 1 submodel pretrained to th error of a value 2.0

What is more, we decided to perform the Supermodeling with almost the same parameters. On the contrary, we changed the error measure for learning the Supermodel to the RMSE. Furthermore, the test procedure have been slightly changed and looked as follows:

1. Generation of the Ground Truth model (GT) from the Handy model.

2. Repeating the tests for sampling frequencies $f_i$ from the set $\{15, 50\}$:

   (a) Sampling observations data from the GT with the frequency $f_i$ (from the learning part, $t_{start} = 150$, $t_{end} = 300$).

   (b) Creating intervals for each of the model parameters as $\pm 10\%$ of the exact value.

   (c) Repeating 10 times the learning procedure:

      i. Learning **2 submodels** with the ABC method , each of them in a time $t_{submodel1}$ (Pretraining submodels to a certain RMSE: 2.0 or 5.0).

      ii. Learning **1 submodel** with the ABC method in a time $t_{submodel2}$ (Pretraining a submodel to a different RMSE: 5.0 or 2.0).

      iii. Creating the Supermodel by synchronization of the most sensitive variable *elites population $x_E$* of each of the submodels.

      iv. Learning Supermodel for the rest of the time $t_{Supermodel}$, which is computed as $t_{Supermodel} = t_{max} - max(t_{submodel1}, t_{submodel2})$.

      v. Generation the forward and backward predictions from the learned Supermodel. Computing the forecasting errors.

## 6.8.2   Results

We have presented the results of the Supermodeling methodology with different submodels in the Tab. 6.11. In the first column *RMSE for pretraining 2 submodels*, we presented the information about the approximated initial RMSE for 2 of the submodels. Whereas in the second column, we presented the approximated RMSE for the last submodel. One more time, the best Supermodels were marked bold. Moreover, we decided to recognize the synchronization of the 2 longer pretrained submodels and 1 shorter pretrained submodel as better for both of the sampling frequencies.

Table 6.11 Results for the Supermodeling with different submodels.

| RMSE for pretraining 2 submodels | RMSE for pretraining 1 submodel | $\bar{t}$[s] | $\overline{t_{norm}}$[s] | $\overline{t_{sumo}}$[s] | $\overline{ff}$ | $\overline{ff_{std}}$ | $\frac{\overline{ff_{std}}}{\overline{ff}}$ | $\overline{f2w}$ | $\overline{f2w_{std}}$ | $\overline{fb}$ |
|---|---|---|---|---|---|---|---|---|---|---|
| | | | | *frequency sampling f = 15* | | | | | | |
| 5.0 | 2.0 | 885.52 | 332.51 | 56.00 | 17.34 | 4.15 | 0.24 | 26.72 | 14.17 | 33.57 |
| **2.0** | **5.0** | **976.53** | **360.65** | **52.71** | **11.75** | **6.17** | **0.53** | **18.16** | **3.29** | **22.84** |
| | *Reference model* | 368.25 | 368.25 | – | 2.50 | 1.03 | 0.41 | 17.11 | 10.20 | 24.07 |
| | | | | *frequency sampling f = 50* | | | | | | |
| 5.0 | 2.0 | 1072.84 | 414.15 | 84.81 | 16.83 | 7.76 | 0.46 | 38.80 | 22.49 | 52.23 |
| **2.0** | **5.0** | **1101.55** | **425.71** | **87.79** | **17.21** | **7.20** | **0.42** | **24.03** | **11.93** | **29.30** |
| | *Reference model* | 426.72 | 426.72 | – | 3.64 | 1.62 | 0.46 | 18.66 | 11.33 | 26.14 |

However, the results were worse than the reference models, it has to be said that using submodels pretrained to different errors provided several advantages for the Supermodeling methodology. We presented below the most important propositions:

1. We have observed that the usage of different submodels caused better prediction errors $\overline{f2w}$ than the previous Supermodeling.

2. What is more, these Supermodels had better or similar predictions on the more chaotic part of the analised system than the reference models.

3. However, the forward forecasting errors ($\overline{ff}$) were negligible worse than for the previous Supermodeling method.

4. Finally, we have confirmed that usage of the different submodels allowed for achieving impact from the synchronization methodology.

Furthermore, we have demonstrated the prediction of the *elites population* for the sampling frequency $f = 15$ (in the Fig. 6.17). The prediction is visualized in a normal form and also normalized form. Evenmore, we highlighted the standard deviation of the value in each point with grey colour.

The results for the rest of variables and for the denser sampling are presented in the Appendix A (Fig. A.25 - A.32).

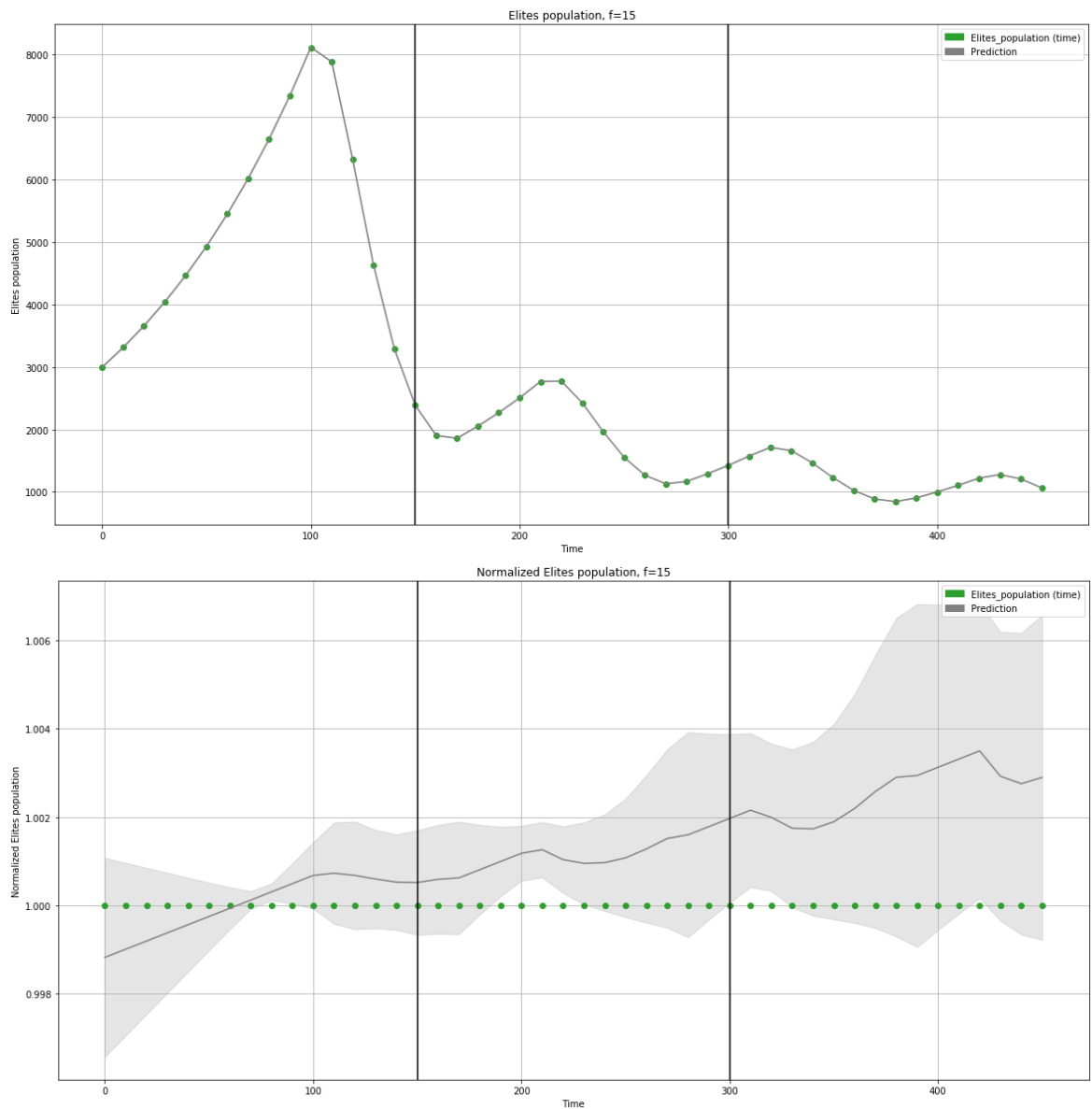

Fig. 6.17 Supermodeling with different submodels results - elites population (both real and normalized), *f = 15*

The analysis of the forecasting plots has forced us to the following conclusions:

- We have observed that the best model for sparser sampling ($f = 15$) achieved similar forecasting to the reference model. Moreover, we were sure that much faster converging on chaotic part to the attractor caused such results.

- On the other hand, we obtained worse predictions on the stable part of the system (forward). However, the diffrences were negligible.

# 6.9 Supermodeling with a longer learning time - converging analysis

## 6.9.1 Idea

The results that we have received, led us to a good question if it was possible to defeat the reference model on the forward forecasting also (the periodic and less chaotic part of the system). To answer this question, we have decided to find if the Supermodeling methodology would converge to the better attractor after the longer time of learning the Supermodel. We set the time for learning only the Supermodel to 600 seconds, which was approximately 10 times longer than in the previous approaches. Moreover, we tested both the Supermodeling with similar submodels (only for submodels pretrained to the approximate errors of values: 2.0 and 5.0) and also the Supermodeling with different submodels.

## 6.9.2 Results

We have presented the results of the Supermodeling methodology with similar submodels and the Sueprmodeling with different submodels in the Tab. 6.12 and Tab. 6.13 respectively. The best models were marked bold as previously.

Table 6.12 Results for the Supermodeling with similar submodels but longer time for learning the Supermodel exclusively (600s).

| RMSE (for pretraining) | $\bar{t}$[s] | $\bar{t}_{norm}$[s] | $\bar{t}_{sumo}$[s] | $\bar{ff}$ | $\bar{ff}_{std}$ | $\frac{\bar{ff}_{std}}{\bar{ff}}$ | $\bar{f2w}$ | $\bar{f2w}_{std}$ | $\bar{fb}$ |
|---|---|---|---|---|---|---|---|---|---|
| *frequency sampling $f = 15$* | | | | | | | | | |
| **5.0** | **1369.57** | **870.60** | **621.11** | **6.78** | **5.26** | **0.78** | **18.96** | **9.60** | **25.94** |
| 2.0 | 1591.41 | 954.33 | 635.79 | 3.22 | 1.60 | 0.50 | 23.58 | 12.18 | 33.19 |
| *Reference model* | 368.25 | 368.25 | – | 2.50 | 1.03 | 0.41 | 17.11 | 10.20 | 24.07 |
| *frequency sampling $f = 50$* | | | | | | | | | |
| **5.0** | **1404.93** | **909.32** | **661.51** | **3.07** | **2.37** | **0.77** | **14.23** | **6.85** | **19.89** |
| 2.0 | 1570.39 | 936.88 | 620.13 | 4.50 | 2.38 | 0.53 | 17.91 | 11.11 | 24.93 |
| *Reference model* | 426.72 | 426.72 | – | 3.64 | 1.62 | 0.46 | 18.66 | 11.33 | 26.14 |

Table 6.13 Results for the Supermodeling with different submodels but longer time for learning the Supermodel exclusively (600s).

| RMSE for pretraining 2 submodels | RMSE for pretraining 1 submodel | $\bar{t}$[s] | $\overline{t_{norm}}$[s] | $\overline{t_{sumo}}$[s] | $\overline{ff}$ | $\overline{ff_{std}}$ | $\overline{\frac{ff_{std}}{ff}}$ | $\overline{f2w}$ | $\overline{f2w_{std}}$ | $\overline{fb}$ |
|---|---|---|---|---|---|---|---|---|---|---|
| | | | | *frequency sampling* $f = 15$ | | | | | | |
| 5.0 | 2.0 | 1495.80 | 937.02 | 657.63 | 17.42 | 6.73 | 0.38 | 28.64 | 11.75 | 36.57 |
| **2.0** | **5.0** | **1515.73** | **923.77** | **627.78** | **13.31** | **5.60** | **0.42** | **15.54** | **4.48** | **17.49** |
| *Reference model* | | 368.25 | 368.25 | – | 2.50 | 1.03 | 0.41 | 17.11 | 10.20 | 24.07 |
| | | | | *frequency sampling* $f = 50$ | | | | | | |
| 5.0 | 2.0 | 1639.36 | 982.70 | 654.37 | 15.79 | 7.22 | 0.46 | 26.30 | 15.59 | 33.68 |
| **2.0** | **5.0** | **1631.79** | **965.75** | **632.72** | **10.82** | **9.02** | **0.83** | **17.76** | **8.11** | **22.67** |
| *Reference model* | | 426.72 | 426.72 | – | 3.64 | 1.62 | 0.46 | 18.66 | 11.33 | 26.14 |

Presented results have implied the succeeding propositions:

1. We have demonstrated that it was possible for the Supermodeling to converge into better attractor than the reference model, but the learning process took much longer.

2. Moreover, for the Supermodeling with similar submodels, we achieved better results for the less pretrained submodels. However, we observed that it was caused by the fact that the submodels were further from the local minima.

3. What is more, we could use the different submodels to enhance the prediction results or equivalently extend the time of learning Supermodel compound of the similar submodels.

4. The presented results have shown that it was possible to achieve better or similar $f2w$ results than the reference models.

5. The Supermodeling is better methodology for the more chaotic systems. We have observed that the overall forecasting ($\overline{f2w}$) was similar to reference models and the backward forecasting ($\overline{fb}$) results were even better.

Furthermore, we have demonstrated the prediction of the *elites population* for the sampling frequency $f = 15$ (in the Fig. 6.18). The prediction is visualized in a normal form and also normalized form. Evenmore, we highlighted the standard deviation of the value in each point with grey colour.

The results for the rest of variables and for the denser sampling are presented in the Appendix A (Fig. A.33 - A.40).

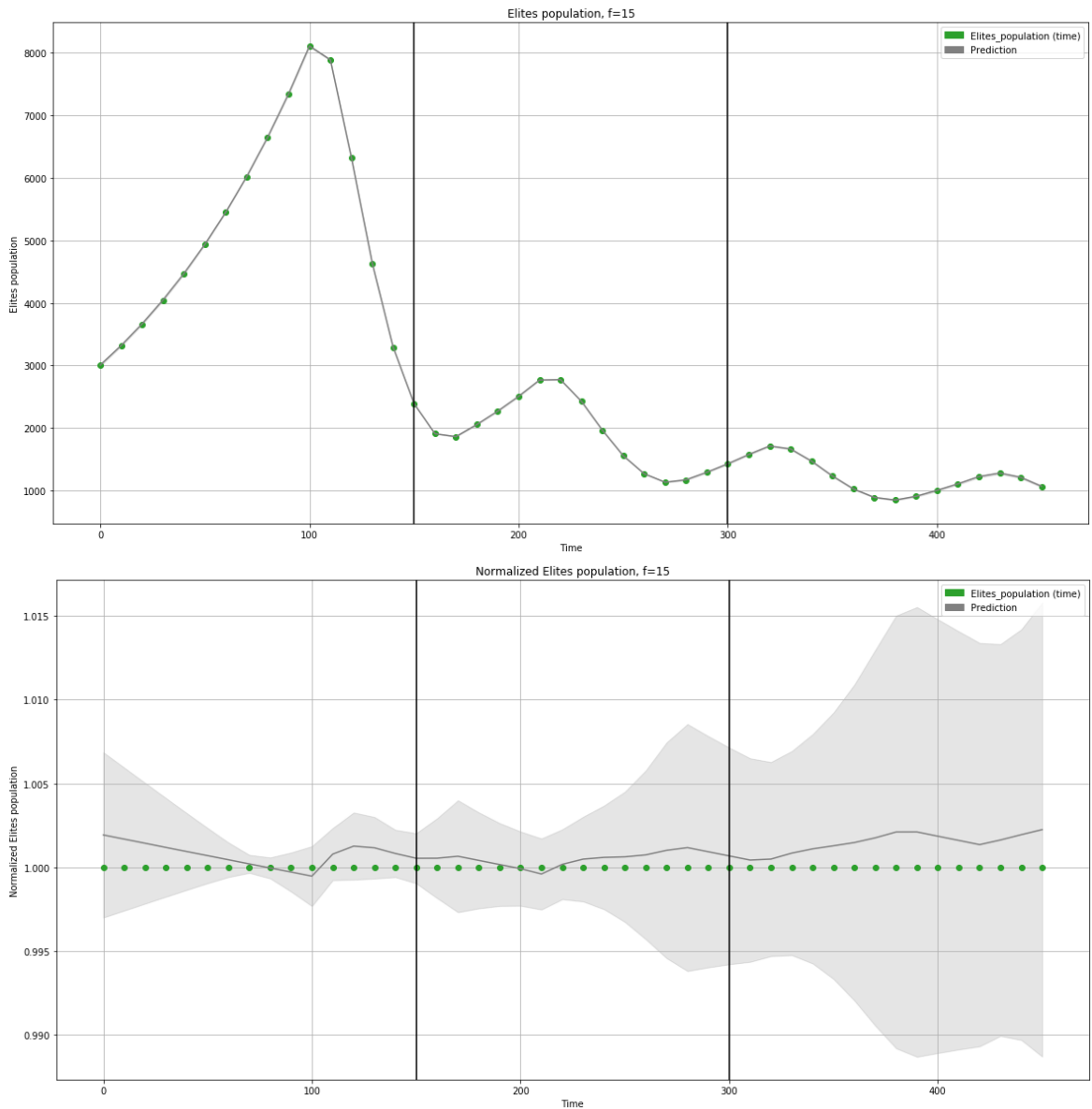

Fig. 6.18 Supermodeling with Supermodel learned longer (600s) results - elites population (both real and normalized), *f = 15*

The analysis of the forecasting plots has forced us to the following conclusions:

- All in all, the best models had better attractors than not only the other Supermodeling approaches but also the reference models. It could be caused by the better synchronization on the chaotis part and more stable forward forecasting.

- However, the Supermodels learned for the longer time have converged slower than the ABC method that was learned for a similar amount of time.

Although, we have proved that the Supermodeling converged to the better attractors than for the other approaches. The obtained results did not answer one of the main questions if using the Supermodeling was more beneficial than the ABC method also for the forward predictions (less chaotic systems).

## 6.10   Supermodeling with the better pretrained submodels

### 6.10.1   Idea

Previously, we have presented that the Supermodeling converged into the better attractors than the reference model, but the question for more stable systems still did not have a possitive answer. However, the prior tests gave us an idea of the another - similar methodology. We have been strongly convinced that it could finally improve the reference models. Due to the fact that the best results were obtained for the Supermodeling with different submodels methodology, we decided to use various submodels. However, we have proposed to improve the methodology by learning two of the submodels to even lower learning error. For this reason, we have performed the tests of the approximation mean time for the ABC method to achieve the learning RMSE of a value which was equaled to 1.5.

### 6.10.2   Results

We presented results for the ABC method in the Tab. 6.14.

Table 6.14 Results for the ABC method for learning RMSE lower than 1.5 in comparison with the reference models.

| RMSE | $\bar{t}$[s] | $\overline{t_{std}}$[s] | $\overline{\frac{t_{std}}{\bar{t}}}$ | $\overline{ff}$ | $\overline{ff_{std}}$ | $\overline{\frac{ff_{std}}{ff}}$ | $\overline{f2w}$ | $\overline{f2w_{std}}$ | $\overline{fb}$ |
|---|---|---|---|---|---|---|---|---|---|
| | | | *frequency sampling $f = 15$* | | | | | | |
| 1.5 | 317.48 | 28.58 | 0.09 | 5.32 | 1.82 | 0.34 | 27.88 | 9.00 | 39.07 |
| *Reference model* | 368.25 | 42.00 | 0.11 | 2.50 | 1.03 | 0.41 | 17.11 | 10.20 | 24.07 |
| | | | *frequency sampling $f = 50$* | | | | | | |
| 1.5 | 336.36 | 59.95 | 0.18 | 3.62 | 1.40 | 0.39 | 25.89 | 9.81 | 36.43 |
| *Reference model* | 426.72 | 33.10 | 0.08 | 3.64 | 1.62 | 0.46 | 18.66 | 11.33 | 26.14 |

We have compared the resulst with the previously performed tests for different learning errors (Tab. 6.2) and have spotted that the mean time for the RMSE of a value 1.5 is very similar to the mean times for RMSE of a value 2.0. Due to that fact, the difference between them could be disputed. However, we have decided perform tests only for the cases of two submodels pretrained to RMSE, which value was equal to 1.5 and the last submodel that was pretrained to the errors: 2.0 or 5.0. The results were presented in the following table (Tab. 6.15).

Table 6.15 Results for the Supermodeling with different submodels pretrained to errors: 1.5, 2.0 or 5.0.

| RMSE for pretraining 2 submodels | RMSE for pretraining 1 submodel | $\bar{t}$[s] | $\overline{t_{norm}}$[s] | $\overline{t_{sumo}}$[s] | $\overline{ff}$ | $\overline{ff_{std}}$ | $\frac{\overline{ff_{std}}}{\overline{ff}}$ | $\overline{f2w}$ | $\overline{f2w_{std}}$ | $\overline{fb}$ |
|---|---|---|---|---|---|---|---|---|---|---|
| | | | | *frequency sampling f = 15* | | | | | | |
| 1.5 | 5.0 | 961.98 | 350.93 | 45.40 | 12.61 | 7.08 | 0.56 | 33.07 | 14.20 | 45.04 |
| **1.5** | **2.0** | **1022.92** | **372.88** | **47.87** | **6.88** | **4.19** | **0.61** | **27.87** | **14.06** | **38.81** |
| *Reference model* | | 368.25 | 368.25 | – | 2.50 | 1.03 | 0.41 | 17.11 | 10.20 | 24.07 |
| | | | | *frequency sampling f = 50* | | | | | | |
| 1.5 | 5.0 | 1133.13 | 457.95 | 120.36 | 7.10 | 3.38 | 0.48 | 29.14 | 22.09 | 40.59 |
| **1.5** | **2.0** | **1128.02** | **426.08** | **75.11** | **3.16** | **2.05** | **0.65** | **14.13** | **6.69** | **19.73** |
| *Reference model* | | 426.72 | 426.72 | – | 3.64 | 1.62 | 0.46 | 18.66 | 11.33 | 26.14 |

The results implied the succeeding propositions:

1. We have observed that the proposed method did not achieve as well results as the reference model for the overall prediction ($\overline{f2w}$) in the situation of spraser sampling ($f = 15$). Even more, the previous approaches to the Supermodeling had better ($\overline{f2w}$) results, which could be caused by the similarity of the submodels.

2. However, we have presented that thod approach to the Supermodeling has better forward forecasting results than each of the previously proposed methods.

3. Furthermore, we have demonstrated that the Supermodeling could have better results for the predictions on more stable systems than the reference model (the sampling frequency $f$ of a value equal to 50).

Furthermore, we have demonstrated the prediction of the *elites population* for the sampling frequency $f = 15$ (in the Fig. 6.19). The prediction is visualized in a normal form and also normalized form. Evenmore, we highlighted the standard deviation of the value in each point with grey colour.

The results for the rest of variables and for the denser sampling are presented in the Appendix A (Fig. A.41 - A.48).

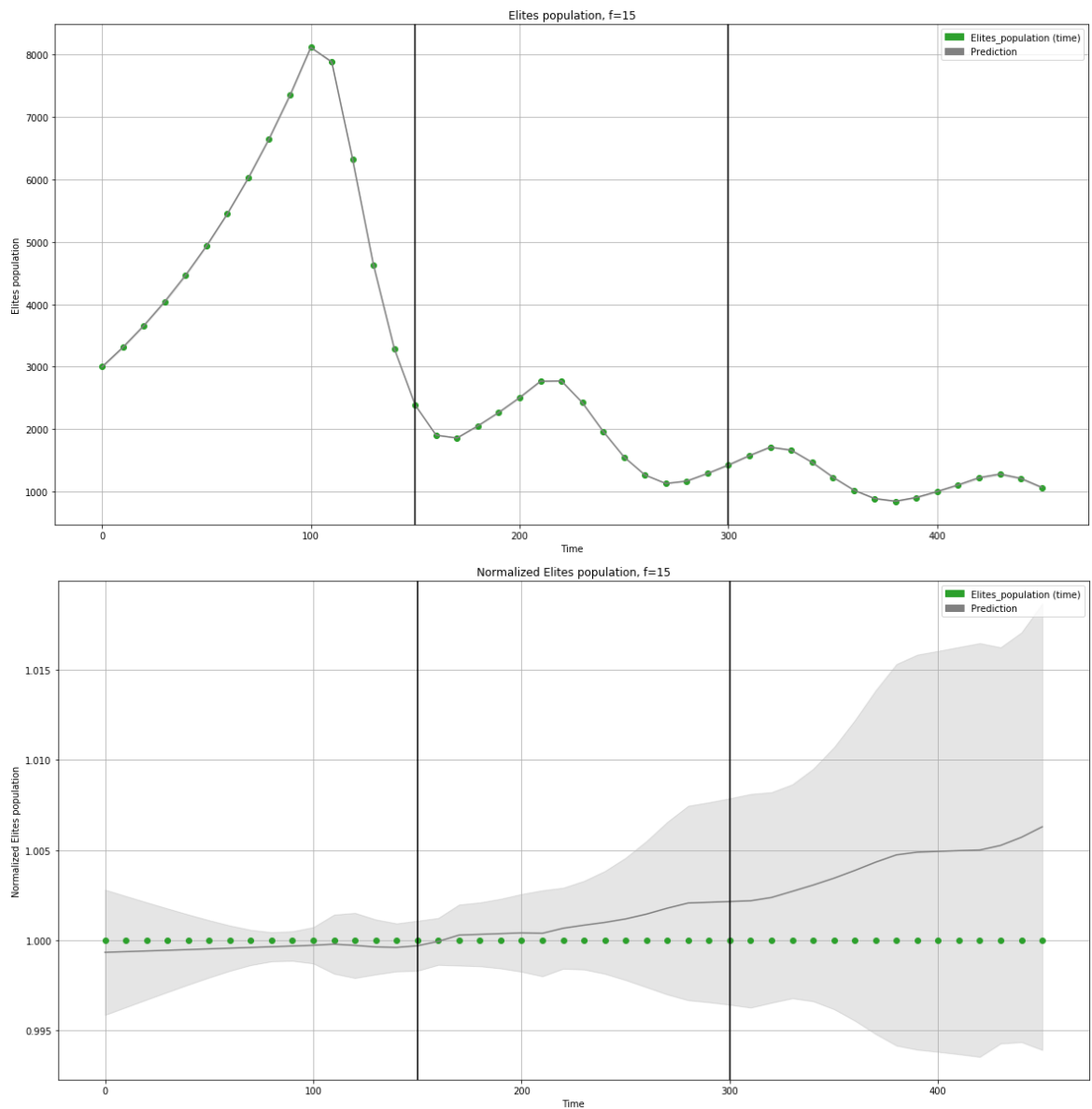

Fig. 6.19 Supermodeling with with lower RMSE submodels results - elites population (both real and normalized), *f = 15*

The analysis of the forecasting plots has forced us to the following conclusions:

- The attractors of the best models had advantages of the other Supermodeling examples - faster converging on the chaotic part and being closer to the GT values (in some plots we presented that the best results are almost the same as GT).

- For the sparser sampling ($f = 15$) the elites population had increasing error for forward prediction, but for denser sampling ($f = 50$), the results' attractors behaved in the same way as the GT.

Finally, all of the above propositions have confirmed that it was beneficial to use the Supermodeling methodology also in the situation of the less chaotic attractors. However, the Supermodeling had better results for such systems only for the denser sampling. We have found that it could be caused by the fact that this methodology were often used for the more complex models.

## 6.11  Findings

During the process of improvement the proposed methodology, we have understood model behaviour and both the advantages and disadvantages of each method better than before. Moreover, it has to be said that all of the methods have been learned with a high precision. Due to this fact, the prediction errors seemed to be marginal. However, even if the forecasting errors were small, they allowed us to prepare the classification of the methods. Furthermore, we have demonstrated the several findings.

First of all, we have found that preknowledge about parameters' sensitivity could improve time of converging the ABC method. However it did not allow to get the attractors better than the reference models. The reason for the last statement was the fact that the process of selection and fixing (after prelearning) several parameters radically decreased the number of the possible attractors that could be achieved. What is more, we were not sure if the fact that this method converged on the learning part, implied similarly good behaviour on the chaotic or more stable part. Although, the initial parameters were quite well estimated, the results were worse than for the ABC method. Moreover, it required the additional time for the Sensitivity Analysis.

Secondly, we have shown that the another methodology that could successfully use sensitivity analysis is the Supermodeling. Obviously, we have expected that usage of the Supermodeling methodology would probably enhance the converging and attractors behaviours on the chaotic part of the system (backward prediction). However, the more demanding seemed to be a way of finding an appriopriate methodology for the more stable part. Taking under consideration the longer computing on each step of learning, it was clear for us that synchronization by the most sensitive variable (based on sensitivity preknowledge) would improve the time of the learning. We have been sure that the trials of the synchronization the insensitive variables would not create the proper (synchronized) Supermodel. The process of the improvement of the Supermodeling methodology had different steps that consisted of finding answers for the various questions.

At the very beginning, we have found that the Supermodels created by the synchronization of similar submodels could resulted with good attractors. However, they have been worse than the reference models. After recognizing the similarity of the submodels (with the PCA method), we decided to use various submodels pretrained for the different times. Because of that, we obtained the sets of three submodels that were not close to the same local minimum. This has led to the better predictions on the chaotic part of the considered system, which were achieved in the same times (after normalization) as the reference models.

Due to the fact that it was still unsure if it was achievable to get better prediction on the stable system behaviour, we have performed the tests with the longer time of learning the Supermodel. The positive answer allowed and encouraged us for the further tests. We decided to use two better pretrained submodels (up to the error of a value 1.5, which was lower then in all of the previous tests) and one less learned submodel. The achieved results have smaller errors for the forward prediction (on the more stable system) for the denser sampling ($f = 50$), and similar for the sparser sampling. Let us notice that the denser sampling was the more demanding case for the ABC method.

In conclusion, the performed tests and the presented results have demonstrated that the preknowledge of parameters' sensitivity could improve the Supermodeling methodology. What is more, the improve Supermodeling methodology could result with the better predictions than the classical data assimilation method - ABC. The synchronization of the Supermodel via the most influential variable have successfully reduced the time requested for learning the Supermodel. It was not suprising that the Supermodeling have better attractors for chaotic systems (backward forecasting), but it was proved that the Supermodeling could be successfully used even for the more stable systems (if the problem is complicated enough). Moreover, the differences for the forward prediction between the ABC and the Supermodeling were negligible. For the chaotic and stable models we proposed to use the Supermodeling synchronized via the most sensitive parameter, whereas for the stable behaviour, the submodels should be learned better (longer).

What is more, one of the possible reason for not as strong impact of the Supermodeling methodology (and the necessity of its improvement) could be usage of the same data assimilation method for learning submodels and the Supermodel. It was possible that ABC have converged to the similar local minimum, despite the fact if it was the process of learning submodels or the Supermodel. One of the necessary further improvement of this methodology is a trial to answer the question if usage of different data assimilation methods for learning some of the submodels or the Supermodel will result with better predictions.

During the process of improvement the methodology, we performed more tests - e.g. with larger initial errors for the submodels or with forward prediction with usage of last known GT data (parameters were the same, however the model started forecasting and synchronization not from point $t_0 = 0$ but $t_1 = 300$).

What is more, we performed even the tests of the Supermodeling with synchronization of the simplified Handy models (see Sec. 2.6 and 6.4). However, such Supermodel did not converge into attractor in the same scale as the GT. It was the main reason to leave the usage of the simplified models. Moreover, it suggested us that the models were probably oversimplified and could not successfully imitate the Handy model behaviour. However, we did not include the mentioned tests in this thesis, because they did not improve the predictions.

# Chapter 7

# Computations overview

Computer simulations are widely used in a field of the mathematical modeling and undeniably need massive computational resources to perform the tasks. In fact, complexity of the computational models (in dimensionality and operations) is the main reason for using *data assimilation methods* and massive computations are the inevitable consequences.

We have come across a similarly essential need during the process of the development of this thesis. Especially important was the fact that the thesis have been based on a research process, which consisted of several steps up to finding the best methodology and answering the questions that have been raised. Although, HANDY model's complexity is not too demanding for modern personal computers, the number of tests that we had to perform, required the much more computational resources. Due to the previously mentioned need, we have decided to use the PLGrid Infrastructure [32] with a support from the National Science Centre to perform the calculations.

## 7.1 Implementation

Handy model as well as each of the tests for the proposed methodologies have been implemented in the Python v. 3.6 [70] with wide usage of Python's scientific libraries - NumPy v. 1.15 [45], SciPy v. 1.1 [31], Scikit-learn v. 0.17 [46, 65, 25] and Pandas v. 0.23 [42]. Whereas, all of the prepared plots were created with usage of the Matplotlib v. 2.2 library [28].

In order to test the proposed methodology and to find significant characteristic of the Handy model (e.g. Sensitivity Analysis) the existing Python's libraries have been used each time when the implementations were publicly available. In such cases, we have used the following dedicated libraries:

- SALib v. 1.1 [26] for the Sensitivity Analysis (SA)

- pyABC v. 0.9 [33] for the Approximate Bayesian Computation (ABC)

Because of not so large popularity of the Supermodeling and it's connection to a specific problem, the entire method had to be implemented.

## 7.2    Computations on a personal computer

As it was mentioned in the previous section, we performed the Sensitivity Analysis tests with usage of the SALib library's implementation of the Sobol method [62, 52, 54]. Due to the fact that the number of Handy's initial parameters $k$ was equal to 15 and the decision of setting the sample size $N$ to be equal $5.0 \times 10^4$, each of the models (Handy and simplified models) had to be evaluated $N(2k+2) = 1.6 \times 10^6$ times (see Sec. 6.4.2).

However, even if the number of evaluations of the complex model seemed to be large (about $6.4 \times 10^6$), we decided to perform tests on a personal computer with the Intel 5257U processor.

Besides the Sensitivity Analysis, also all not so demanding computations, during the analysis of the model and the methodology, have been performed on the personal computer. To mention only the most important: checking the sampling frequencies, the PCA method [65, 25] and the results analysis.

All of the computations on the personal computer have been performed with usage of the PyCharm and the Jupyter Notebooks [34, 48].

## 7.3    Computations on a cluster

The rest of the computations were much more demanding. There are several reasons for that fact - to mention only the most important:

- Learning the ABC up to error equaled to 1.0, took almost an hour on a personal computer.

- Each of the test sets consisted of 10 evaluations of the proposed method for two different sampling frequencies. Consequently, each method had to be performed 20 times.

- During the process of the research and development, each of the methods had to be run with various initial parameters.

• Due to the fact that the main thesis quastions were about mean times of the computations and the prediction errors (achieved after mentioned times), each of the repetitions had to be performed on the same hardware.

The above highlighted requirements were the factors which enforced the necessity of usage the massive computational resources. Thanks to support from the National Science Centre, Poland (grant no. 2016/21/B/ST6/01539) the usage of the PLGrid Infrastructure [32] for large-scale calculations and simulations was possible.

The computations were performed on the Prometheus supercomputer (localized in the ACK Cyfronet AGH), which allowed for using computational cluster with following parameters (Tab. 7.1):

Table 7.1 Prometheus cluster configurations.

| Number of nodes | Processors in node | Cores in processor | Cores in node | Processor model | GHz |
|---|---|---|---|---|---|
| 4 | 2 | 12 | 24 | Intel Xeon E5-2680 v3 | 2.5 |

During the tests, we have used the walltime on the Prometheus, which value was equal to 167042 hours. Let us notice that the used walltime have proved the necessity of the massive computational resources. What is more, we are sure that these computations could not be performed on any personal computer.

# Chapter 8

# Conclusions

In the following chapter, we will demonstrate research objectives, further recommendations and finally the contributions to knowledge. First of all, it has to be mentioned that this thesis is mostly focused on comparison between the classical and novelty data assimilation methods in the field of mathematical modelling. Furthermore, during the research process we were especially interested in finding a detailed comparison of the Approximate Bayesian Computation [64] and the Supermodeling methodology [75] for the prediction of model behaviour. The thesis had two main aims which should be recalled. Firstly, we have endeavoured to the simplification of the selected computational model (Human and Nature Dynamics - HANDY [43]). Secondly, we have studied the capabilities of the mentioned data assimilation methods used for such nontirvial model.

## 8.1   Research objectives

During the thesis development, the literature review was an essential step for the deeper understanding about the research field. We found that the Supermodeling methodology has not been neither compared to classical data assimilation methods nor used for more stable models. Moreover, the studies of the Supermodeling are mainly focused on less complex systems (with fewer number of variables - e.g. [20, 19, 47]) or very chaotic models (e.g. [75, 59]). However, in the second case, the comparison with classical methods could not be made. Anyway, this methodology has proven convergence to the appriopriate systems' attractors for any of the previously mentioned applications.

In the presented study, we have demonstrated the possibilities for simplification the Handy model with the preknowledge about parameters' sensitivity. However, it has to be pointed out that the proposed simplified models cannot be successfully used for the prediction. This finding has led us to the conclusion that transformations of the core part of the model could result with an inability to approach system attractor.

Nevertheless, the main aim of the thesis - a complete study on capabilities of both ABC and Supermodeling has been presented. First of all, let us point out that the analysed methods have been learned with a high precision. Because of that, the prediction errors seem to be marginal. However, even if the forecasting errors are small, they allow to prepare the methods classification.

First of all, we have demonstrated that the Supermodeling methodology could achieve better predictions than classical ABC for the chaotic system's behaviour. On the other hand, the differences between these methods for stable attractors are insignificant. During the reaserch process, we have shown that the Supermodel could be successfully synchronized via the most sensitive variable. The proposed methodology allow to reduce the complexity of the Supermodel and consequently enhance the speed of the method convergence. Anyway, the analysis of the submodels similarity is crucial for predictions accuracy - we have demonstrated the Supermodeling improvement after synchronization the submodels with various initial pretraining times.

What is more, it was presented that the Sensitivity Analysis [62, 54] allow for decreasing the time needed for learning the model. Despite the fact that ABC on sensitive parameters fitted initial parameters sufficiently good, this method had not better predictions than reference model. In addition, it requires the preknowledge about sensitive parameters.

## 8.2   Recommendations

The Supermodeling has been demonstarted to defeat the ABC method for the prediction the more chaotic part of the model attractor. However, it should be noticed that we have chosen the model behaviour which is not completely disordered. Due to the above, we recommend the further research of relation between both Supermodeling and ABC predictions for chaotic systems. Another appriopriate way of development the proposed methodology should be the comprehensive study of the Ground Truth interval - forecasting range correlation.

According to the fact that the Supermodeling is still a novelty methodology in the field of data assimilation, we propose several recommendations for the further and research. First of all, we suggest developing the proposed methodology by synchronization of the slightly pretrained submodels and consequently set the longer time for learning the connected Supermodel. Secondly, it would be beneficial to answer the question if the more complex Supermodel gives better predictions. We suggest to either synchronize the Supermodel via two of the most sensitive variables or use the *nudging* method. Moreover, we recommend to try the synchronize the Supermodel via different variables. For example, two pairs of the submodels could be synchronized via the most sensitive variable, whereas the last pair vie the second most sensitive variable. We have been strongly convinced that such method will result with the interesting findings.

Finally, we have observed and presented that if submodels were converged into the similar local minima, the Supermodel would not have better forecasting results. For this reason, we recommend to use various data assimilation methods to learn submodels or the Supermodel. The last proposed recommendation would give the another significant contribution to knowledge.

## 8.3    Contributions to knowledge

Supermodeling is the methodology that have been developed mostly in the field of climatology. That is the main and most obvious reason why it have not been comprehensively compare with another data assimilation methods. Furthermore, analyses of the very chaotic systems and the less interest in the converging time have contributed to the lack of such researches. Anyway, the studies of the Supermodeling are mainly focused on less complex systems (with fewer number of variables) or very chaotic models.

Nevertheless, we have considered the detailed study of the comparison between the Supermodeling and classical methods as the necessity. What is more, we hope that our findings will not only be influencial in the climatologists community but also will help bring the Supermodeling to the core of data assimilation methods.

# Appendix A

# Forecasting results

The appendix contains forecasting results for the models that achieved the lowest prediction errors for each step of the research process (Ch. 6). Results are presented for both sampling frequencies ($f = 15$ and $f = 50$) and show the model behaviour on the time interval of interests ($[0, 450]$). The plots are ordered in the following way:

- Commoners population results - both with and without normalization

- Elites population results (presented in the same way)

- Nature results (presented in the same way)

- Wealth results (presented in the same way)

Let us notice that each of the plots presents both the Ground Truth models in the mentioned sampling frequency (blue, green, orange and red points respectively for the variables) and the prediction results in grey colour. What is more, the mean results for each of the time stamps are presented as a grey line, whereas the rest of the grey area is connected to a standard deviation of each of the predicted value. That is why the predictions with the standard deviations are much better visible on the normalized plots. The appendix contains results for:

1. Reference models (ABC)

2. ABC on sensitive parameters

3. Supermodeling with similar submodels

4. Supermodeling with different submodels

5. Supermodeling with Supermodel learned longer (600s)

6. Supermodeling with lower RMSE submodels

# A.1   Reference ABC

## A.1.1   Sampling frequency *f = 15*

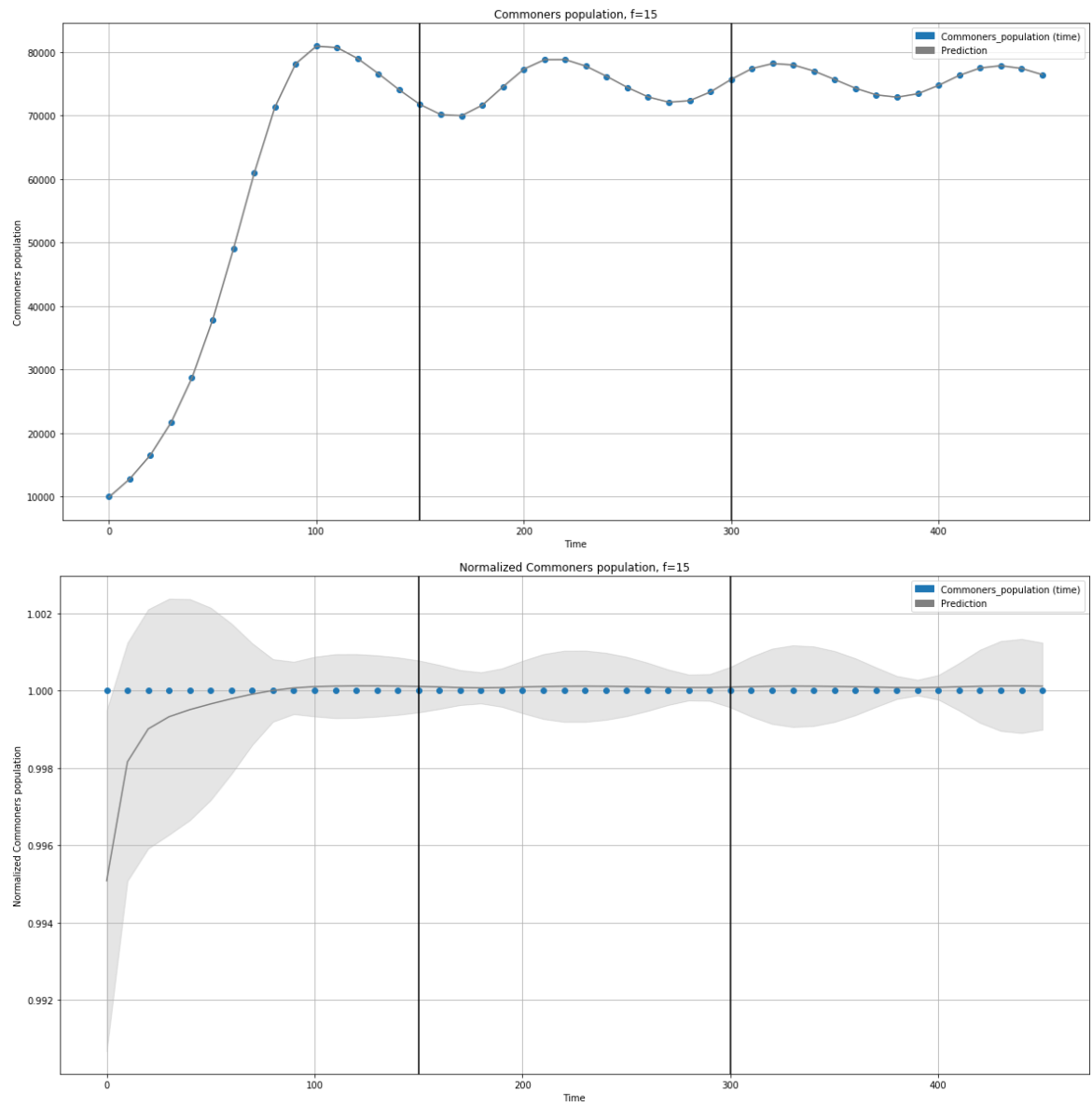

Fig. A.1 Reference ABC results - commoners population (both real and normalized), *f = 15*

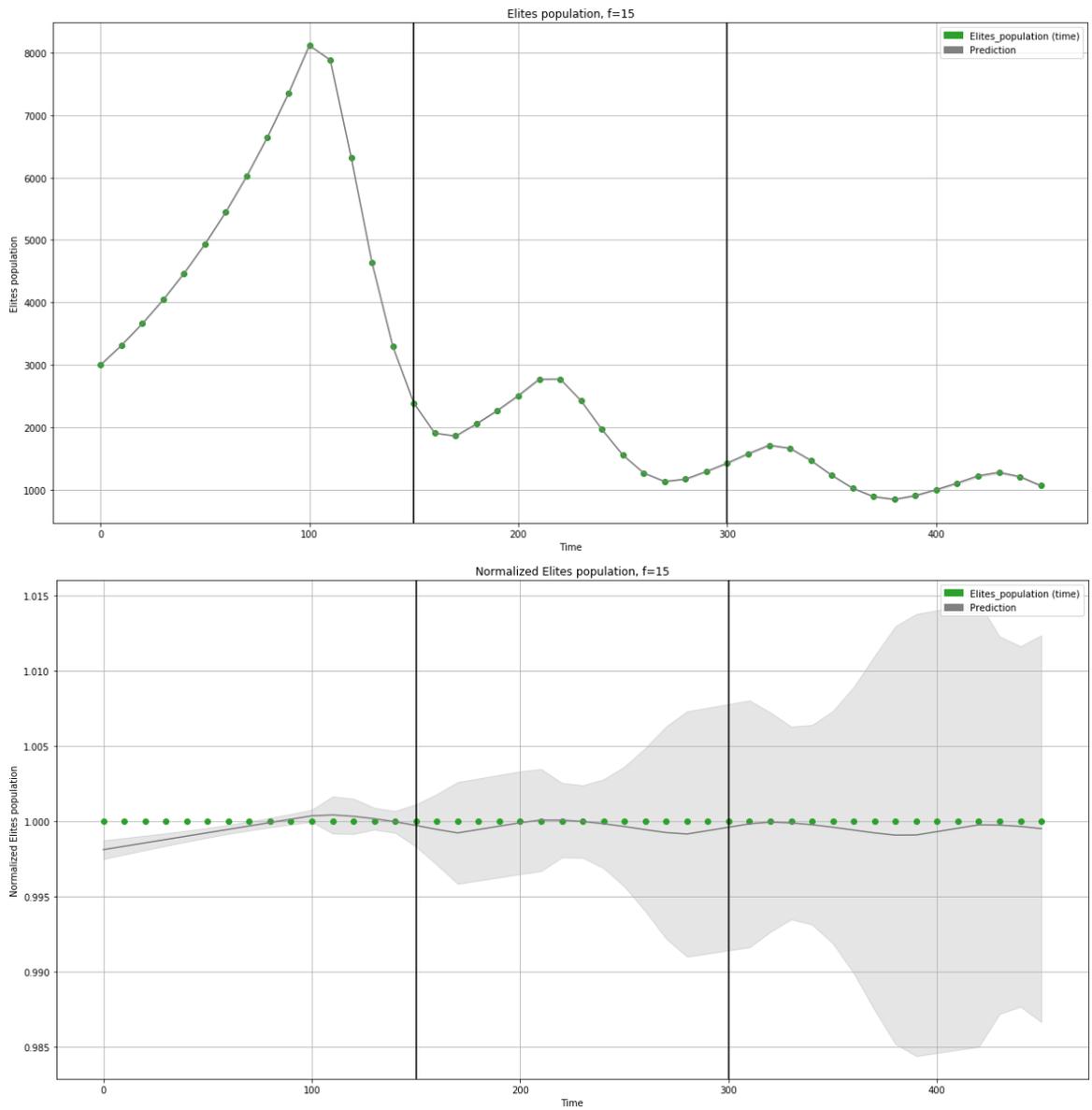

Fig. A.2 Reference ABC results - elites population (both real and normalized), *f = 15*

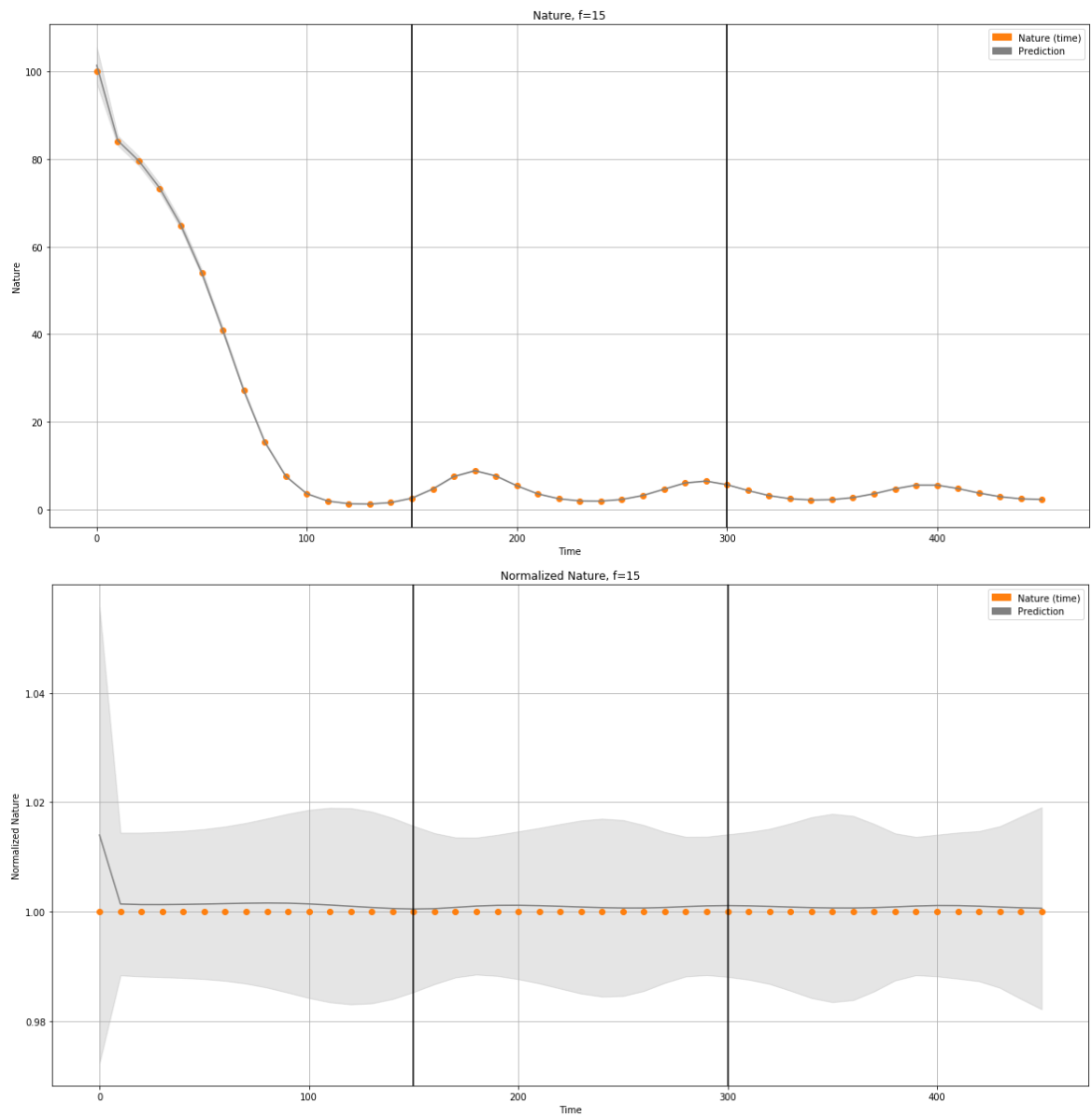

Fig. A.3 Reference ABC results - nature (both real and normalized), $f = 15$

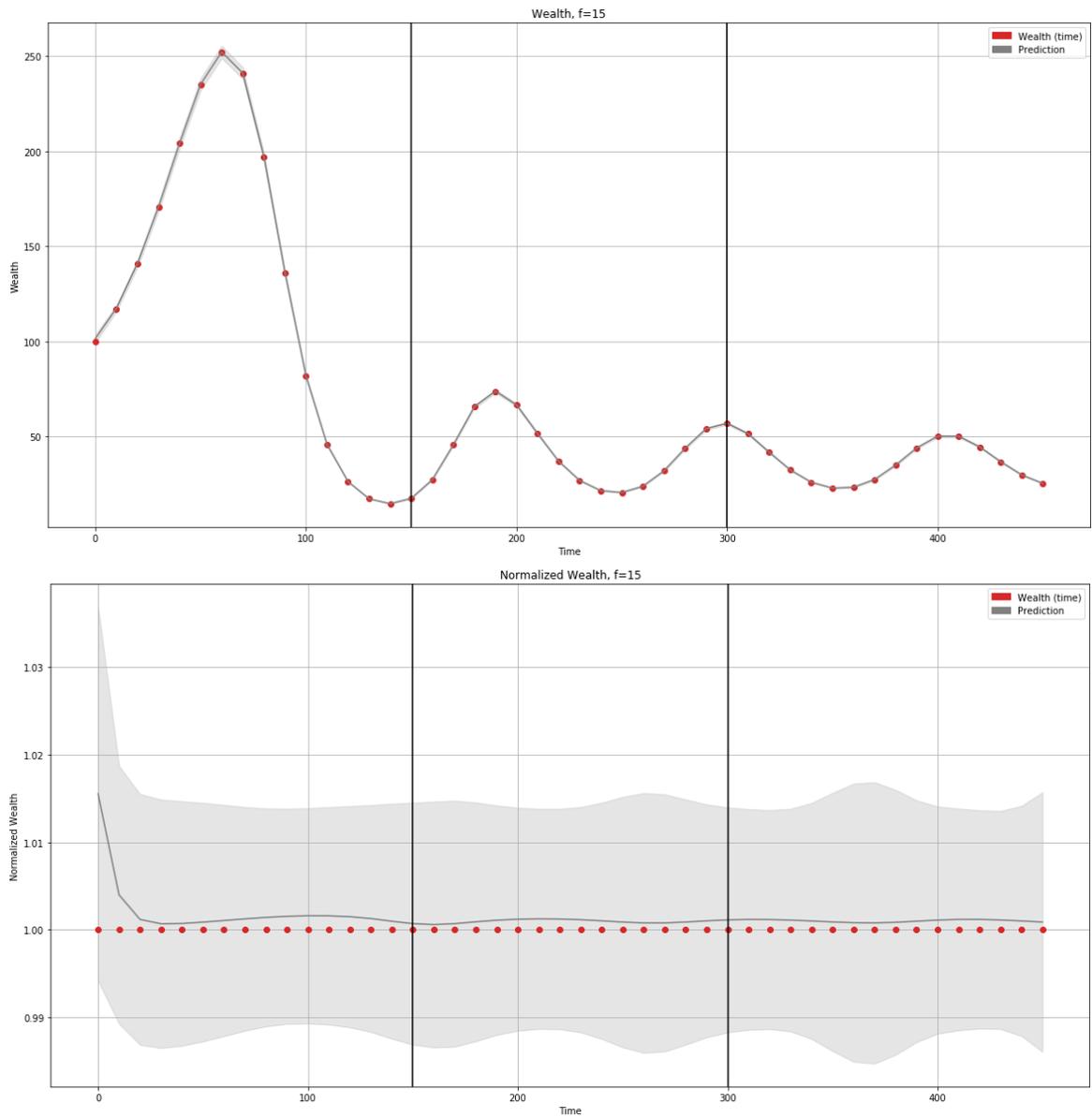

Fig. A.4 Reference ABC results - wealth (both real and normalized), *f = 15*

## A.1.2 Sampling frequency *f = 50*

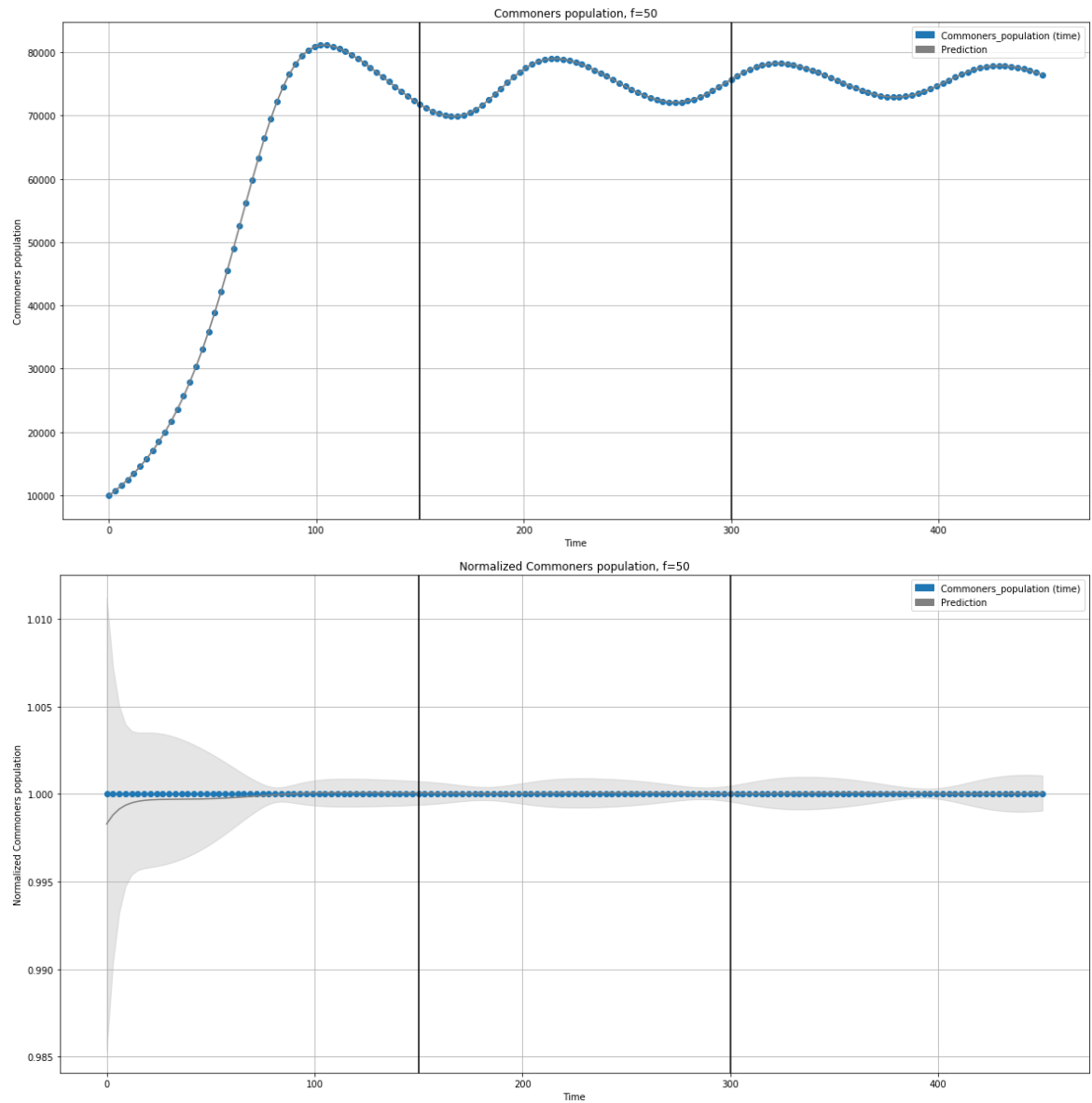

Fig. A.5 Reference ABC results - commoners population (both real and normalized), *f = 50*

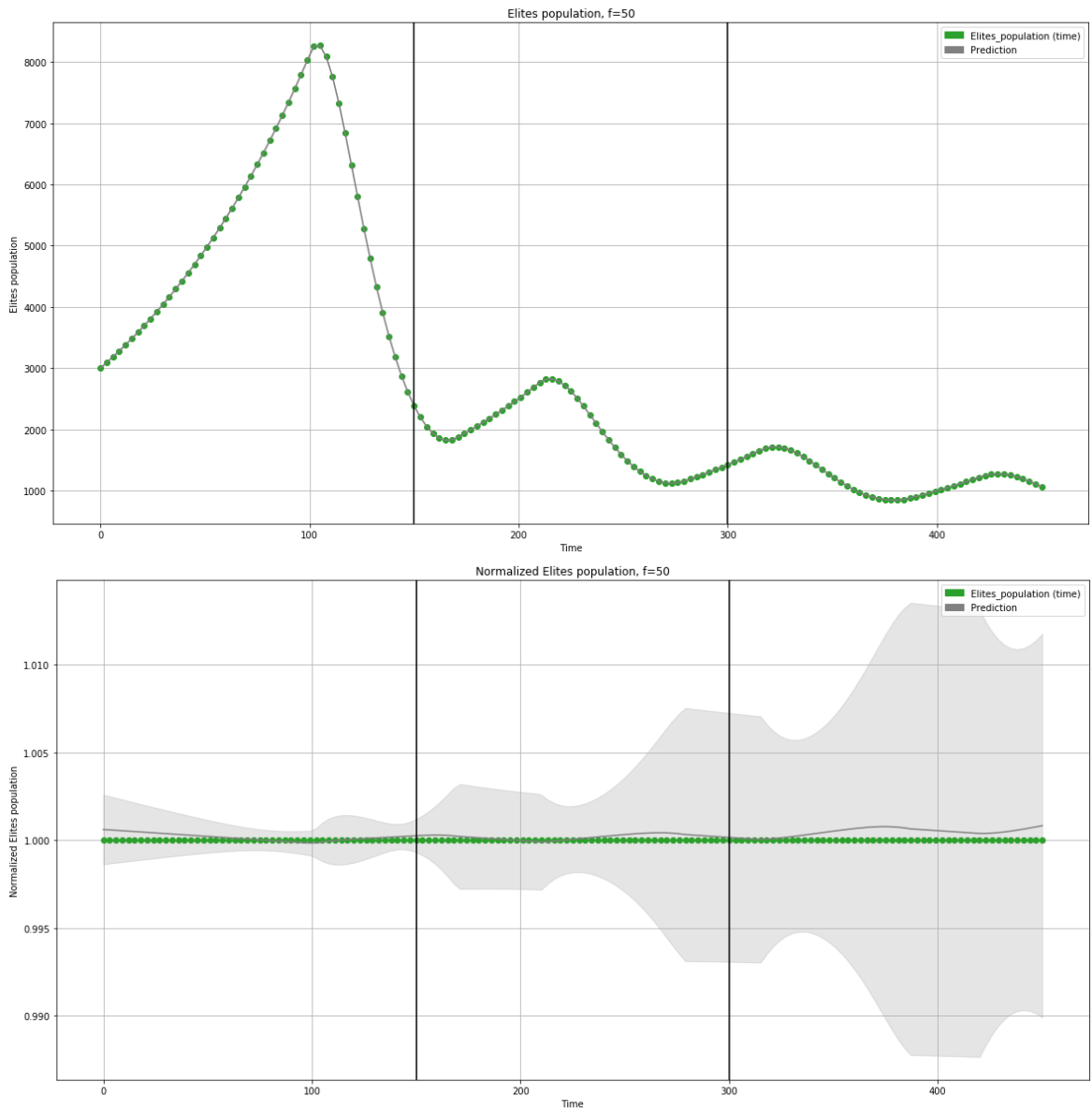

Fig. A.6 Reference ABC results - elites population (both real and normalized), *f = 50*

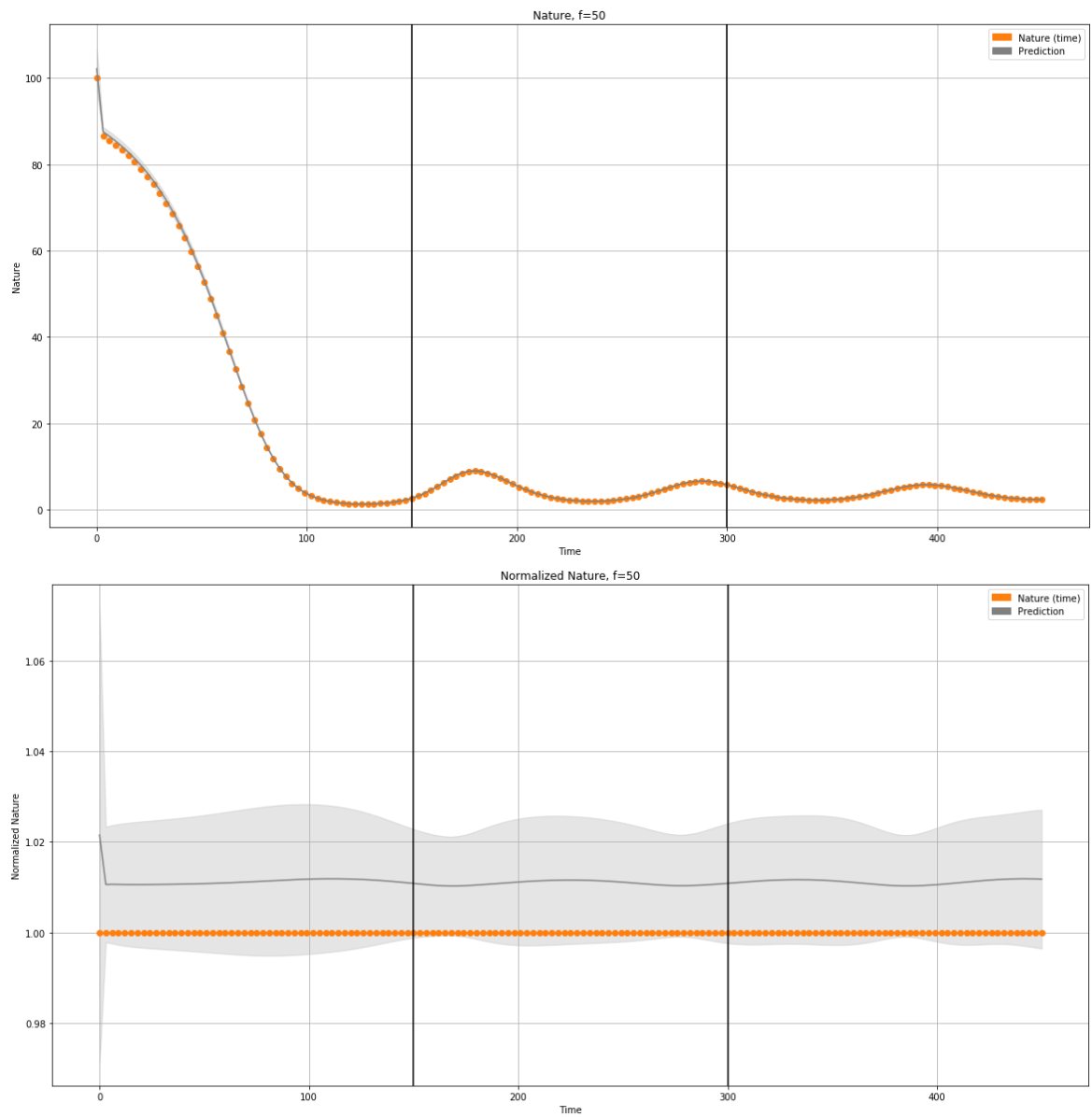

Fig. A.7 Reference ABC results - nature (both real and normalized), *f = 50*

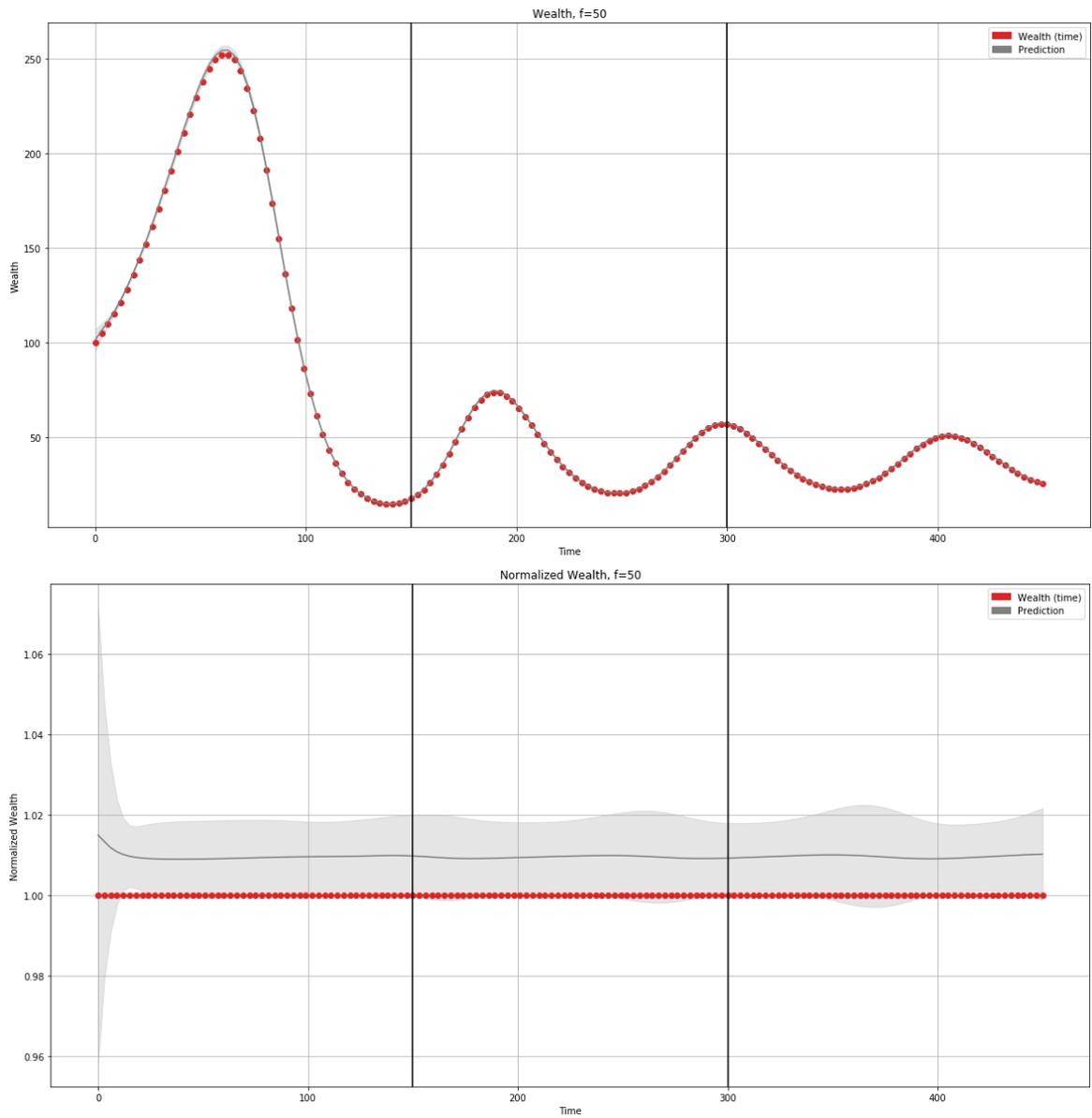

Fig. A.8 Reference ABC results - wealth (both real and normalized), *f = 50*

## A.2   ABC on sensitive parameters

### A.2.1   Sampling frequency *f = 15*

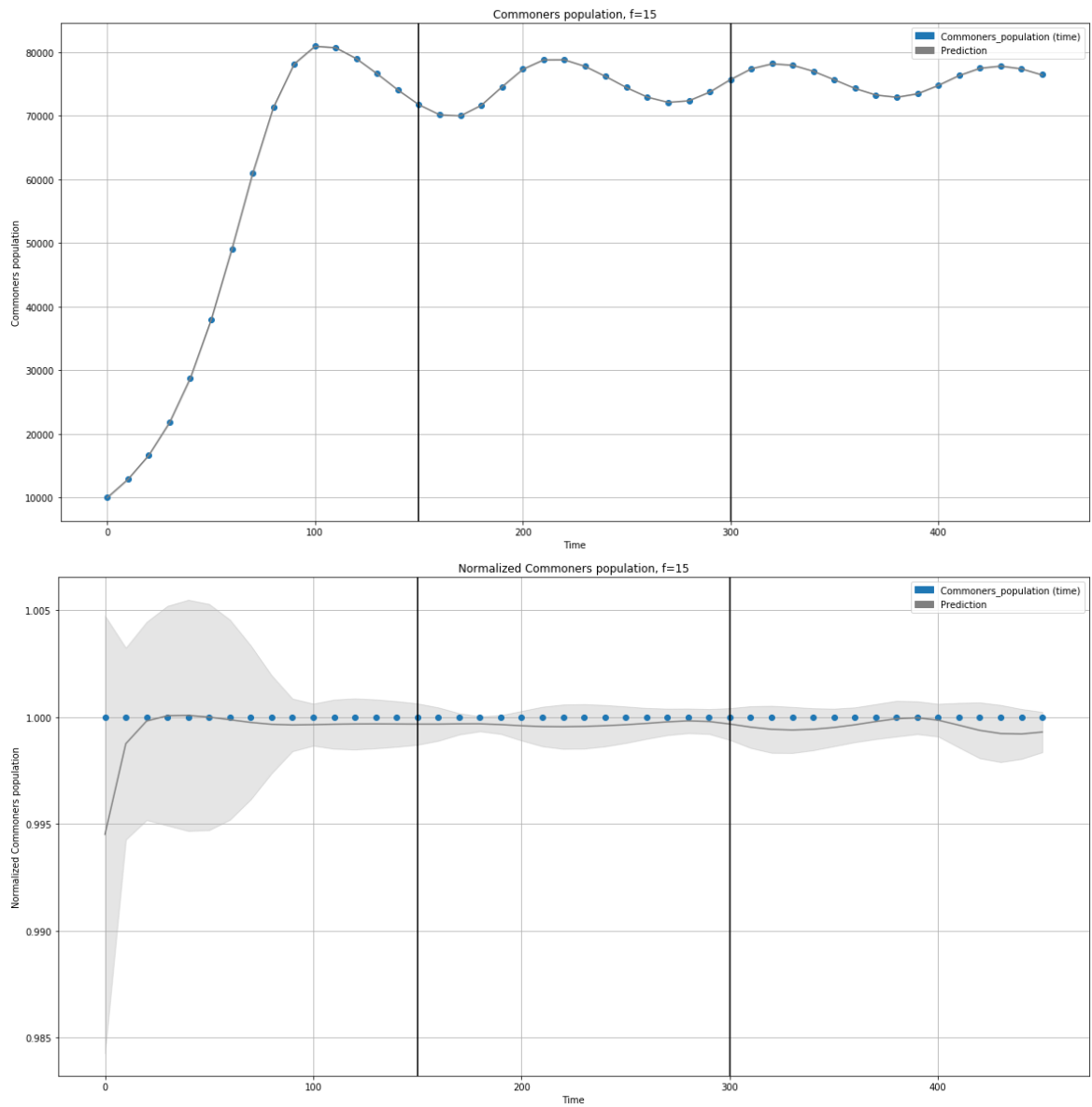

Fig. A.9 ABC on sensitive parameters results - commoners population (both real and normalized), *f = 15*

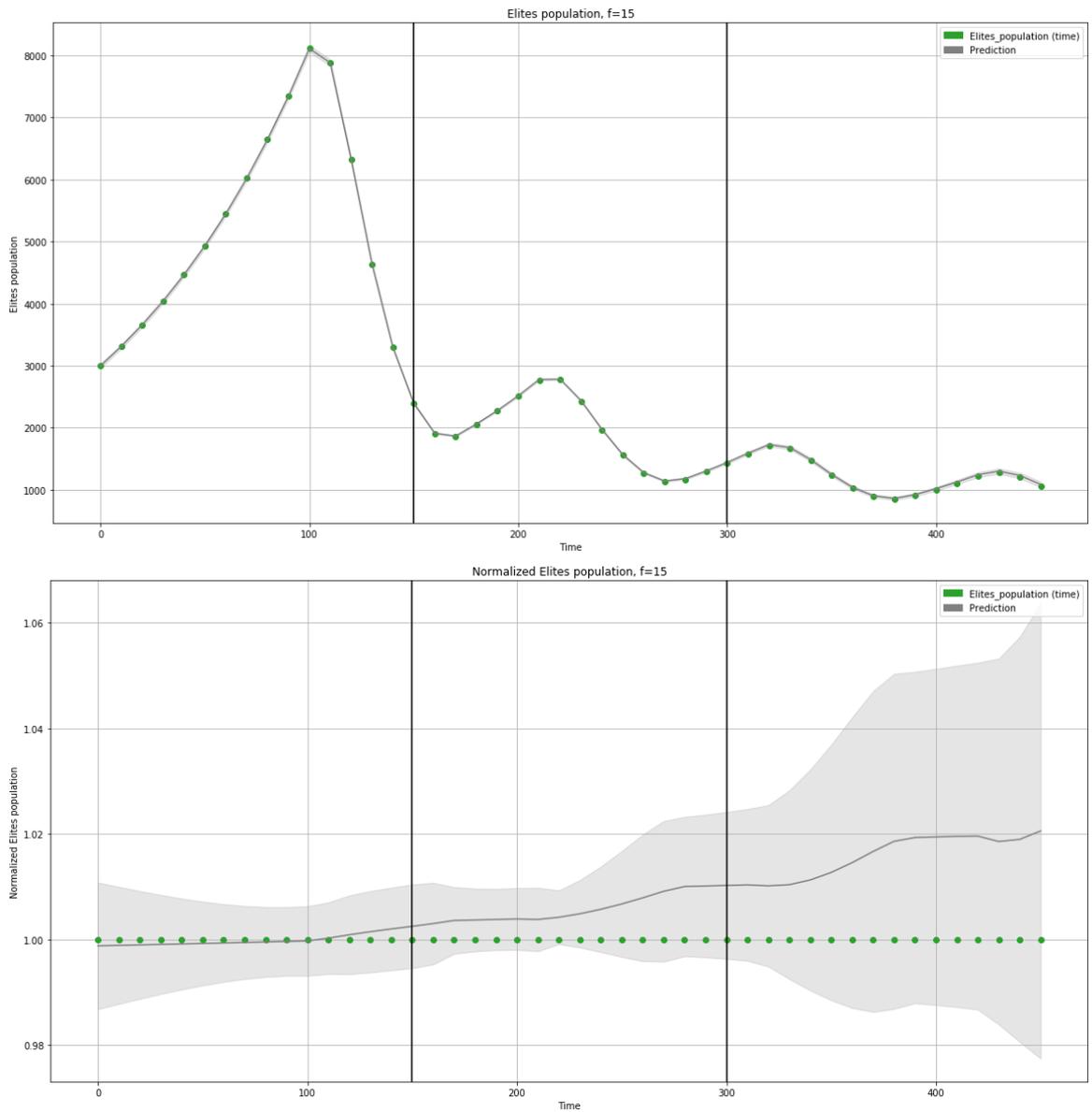

Fig. A.10 ABC on sensitive parameters results - elites population (both real and normalized), *f = 15*

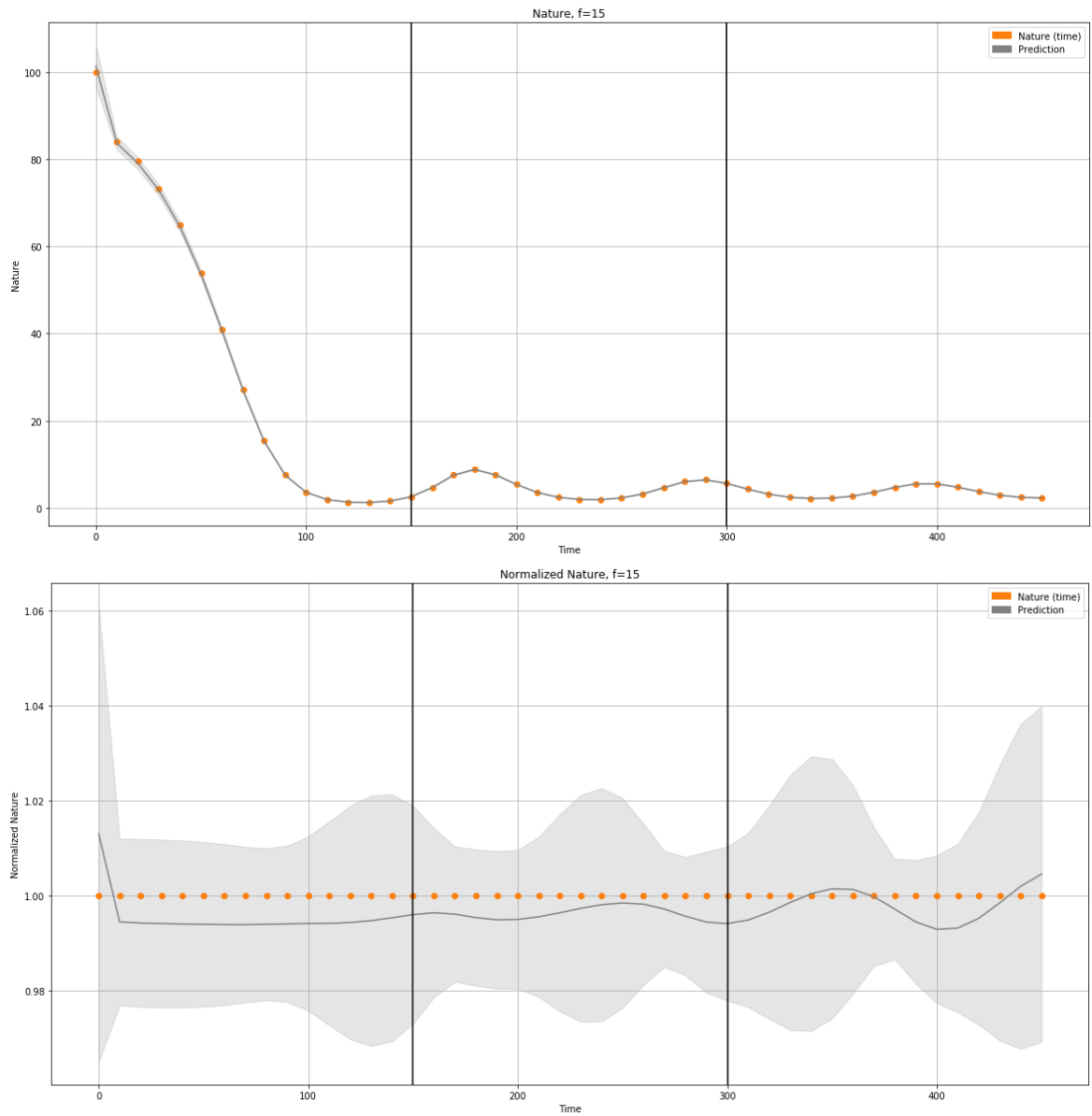

Fig. A.11 ABC on sensitive parameters results - nature (both real and normalized), $f = 15$

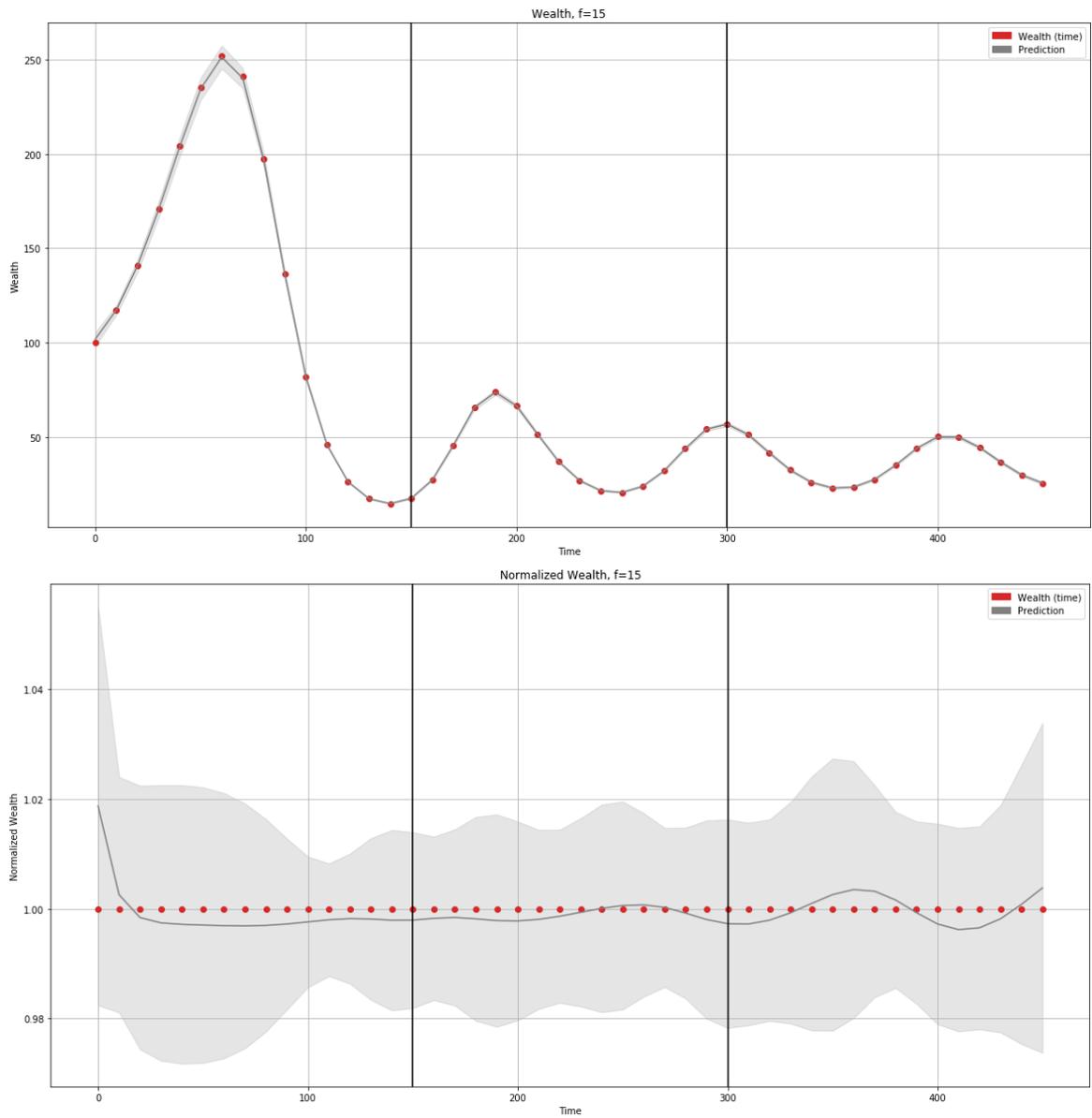

Fig. A.12 ABC on sensitive parameters results - wealth (both real and normalized), *f = 15*

## A.2.2   Sampling frequency *f = 50*

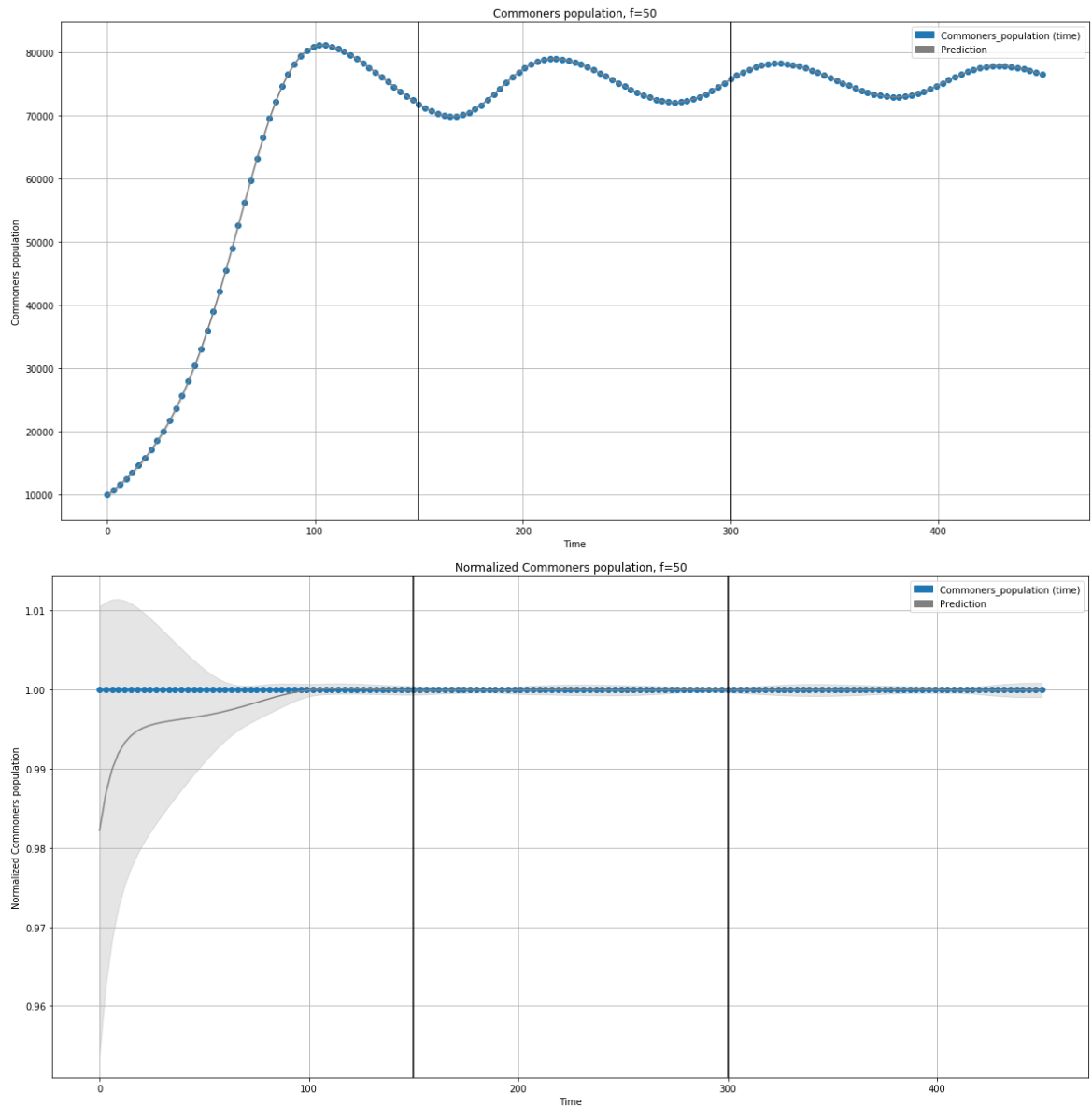

Fig. A.13 ABC on sensitive parameters results - commoners population (both real and normalized), *f = 50*

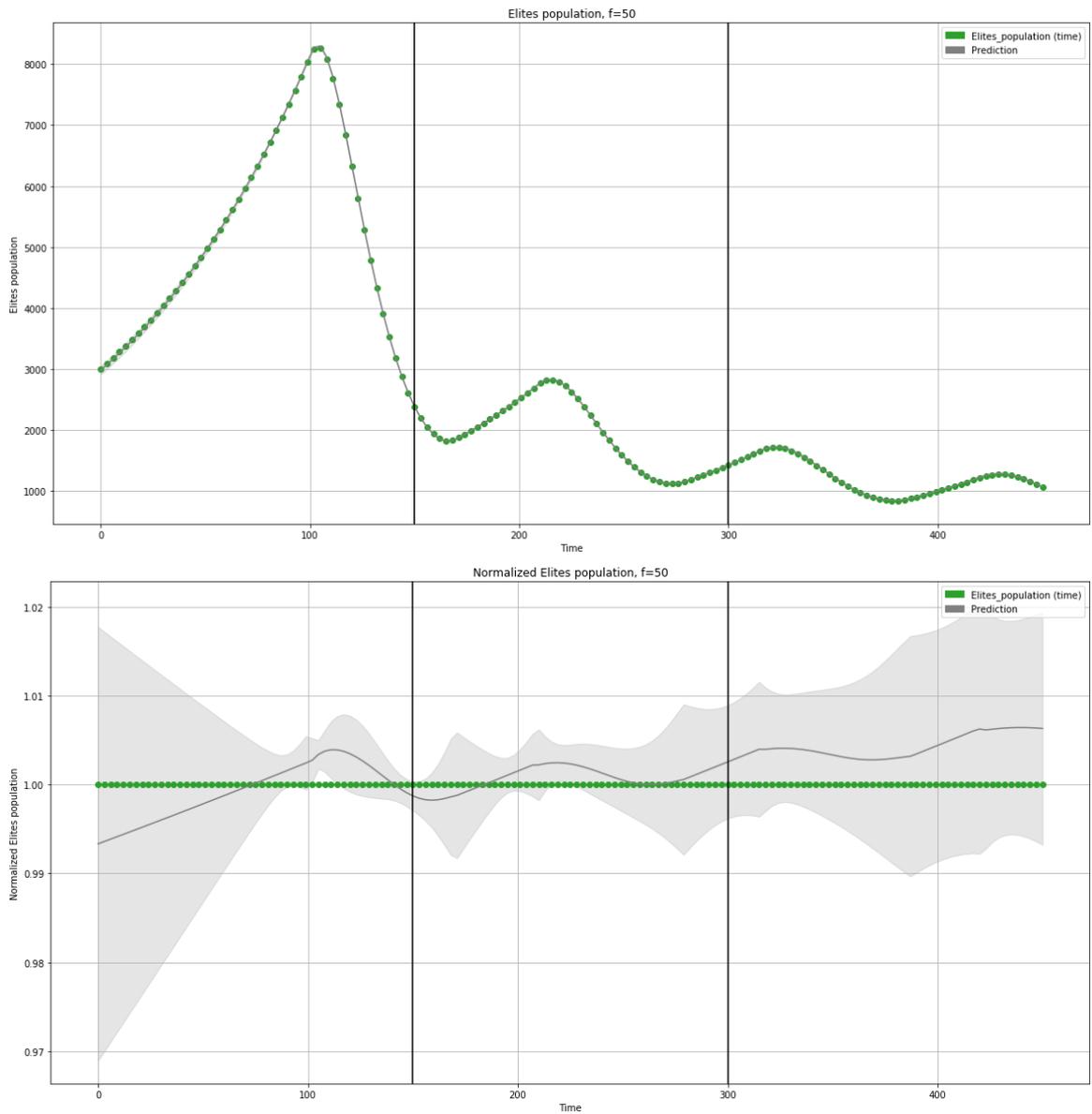

Fig. A.14 ABC on sensitive parameters results - elites population (both real and normalized), *f = 50*

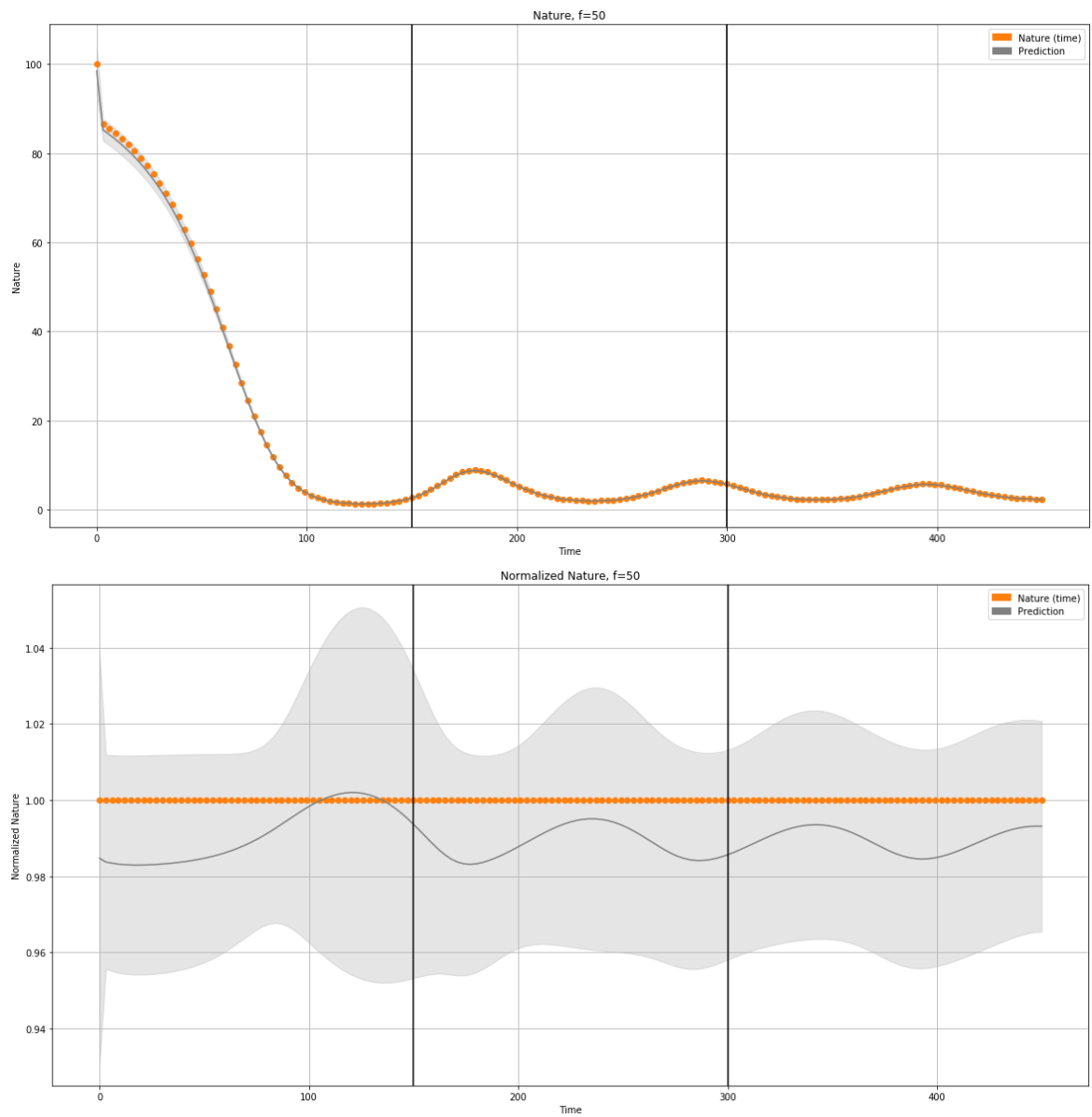

Fig. A.15 ABC on sensitive parameters results - nature (both real and normalized), *f = 50*

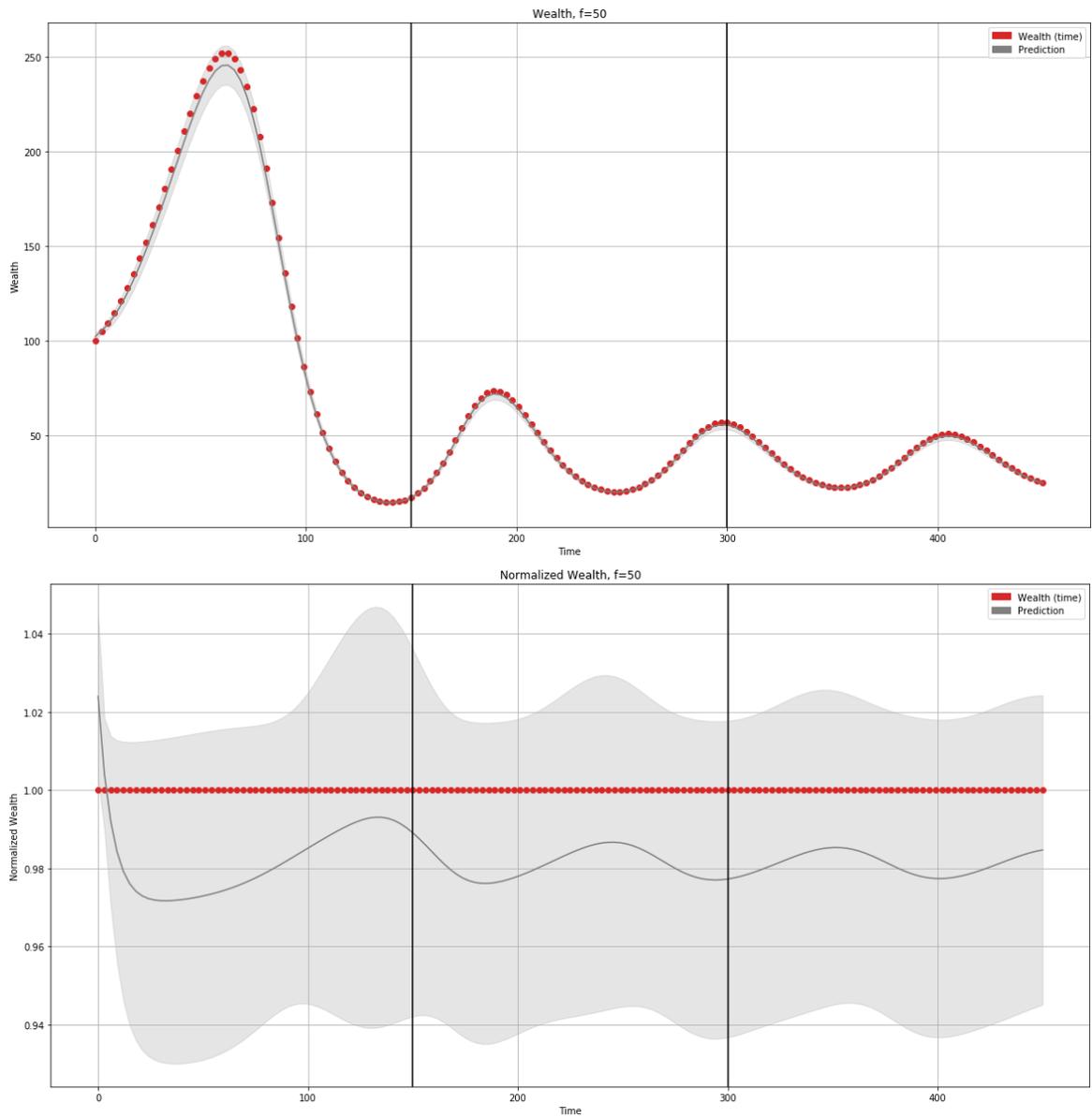

Fig. A.16 ABC on sensitive parameters results - wealth (both real and normalized), *f = 50*

## A.3   Supermodeling with similar submodels

### A.3.1   Sampling frequency *f = 15*

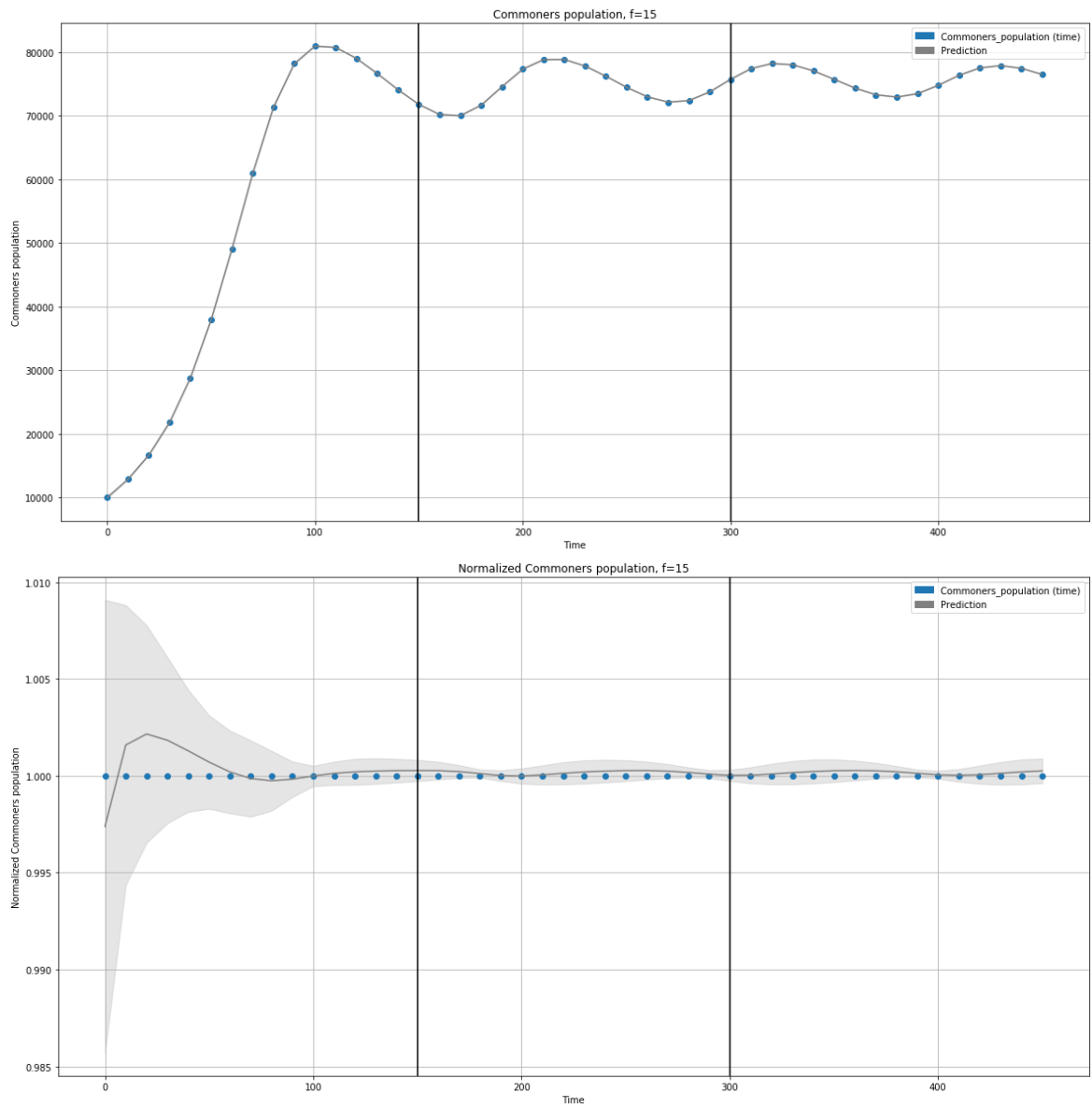

Fig. A.17 Supermodeling with similar submodels results - commoners population (both real and normalized), *f = 15*

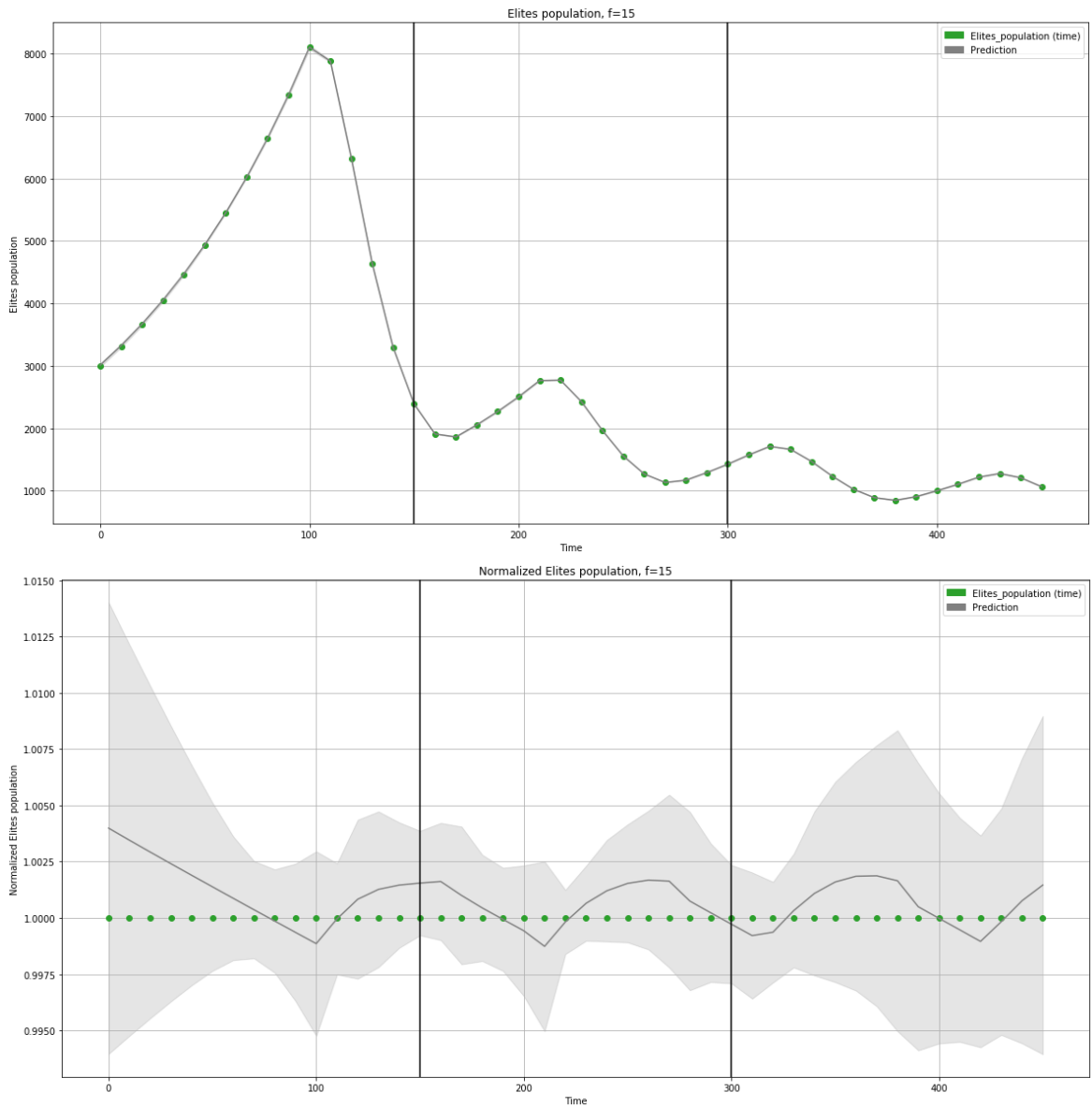

Fig. A.18 Supermodeling with similar submodels results - elites population (both real and normalized), *f = 15*

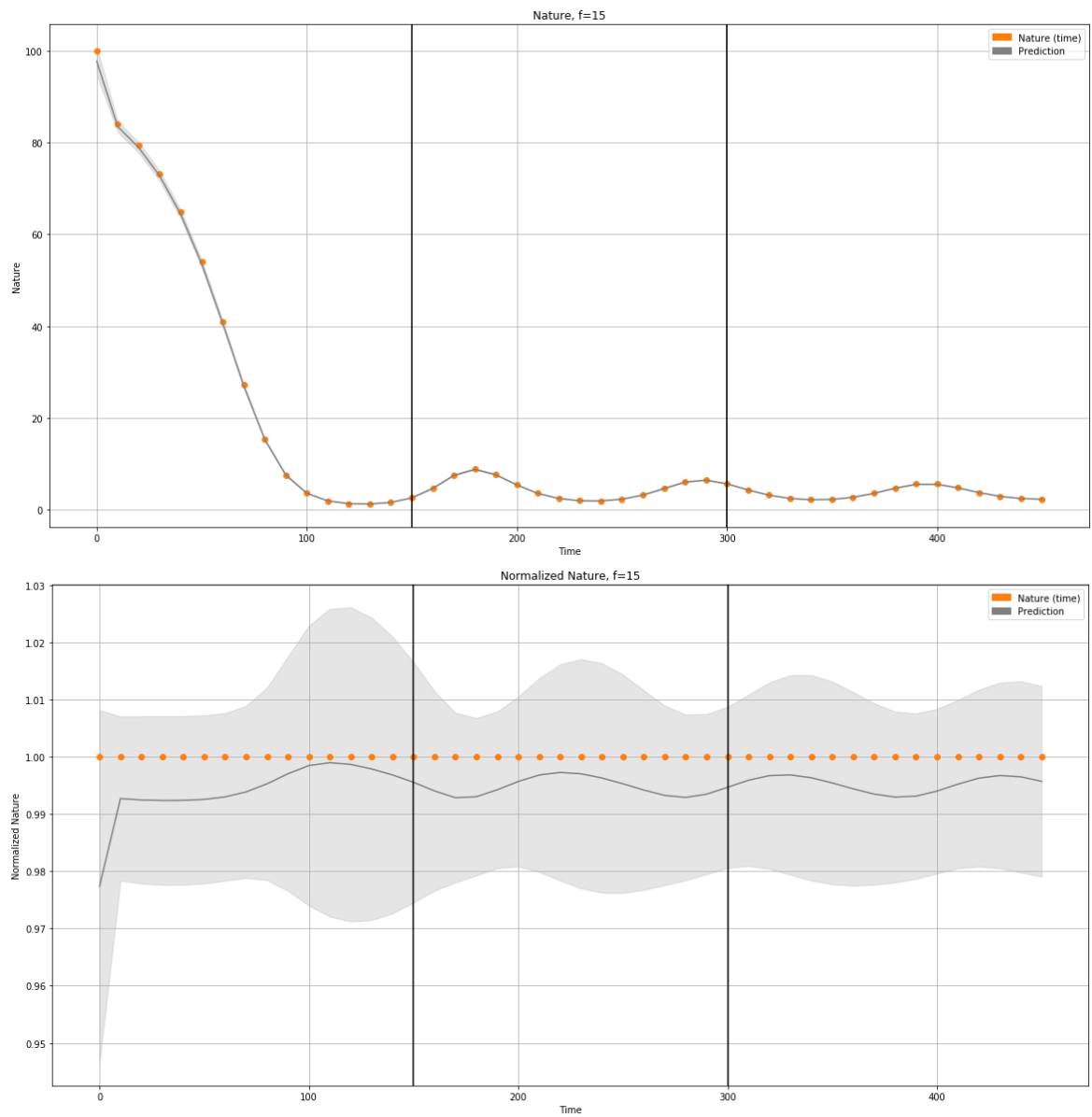

Fig. A.19 Supermodeling with similar submodels results - nature (both real and normalized), *f = 15*

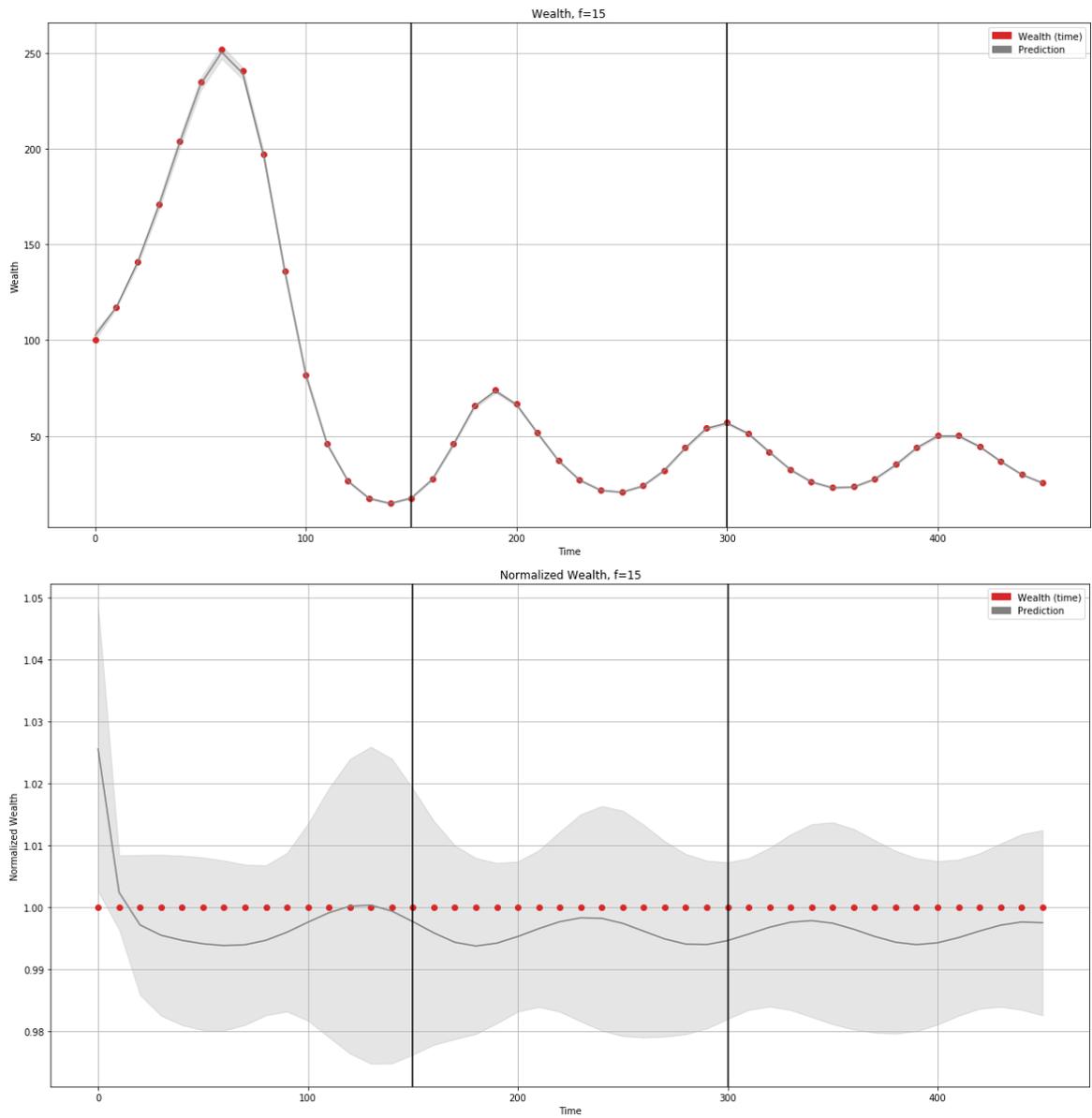

Fig. A.20 Supermodeling with similar submodels results - wealth (both real and normalized), *f = 15*

## A.3.2    Sampling frequency *f = 50*

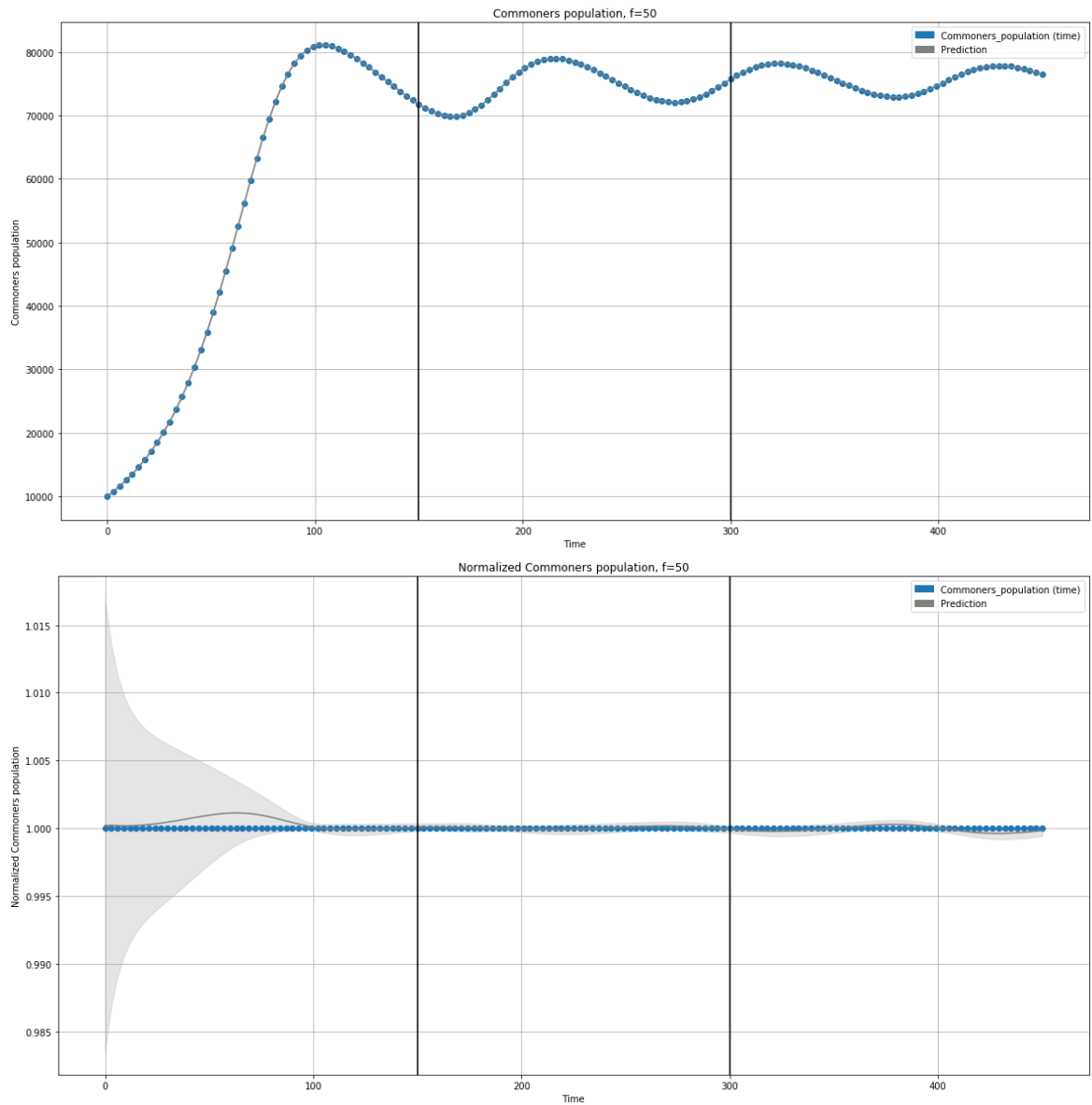

Fig. A.21 Supermodeling with similar submodels results - commoners population (both real and normalized), *f = 50*

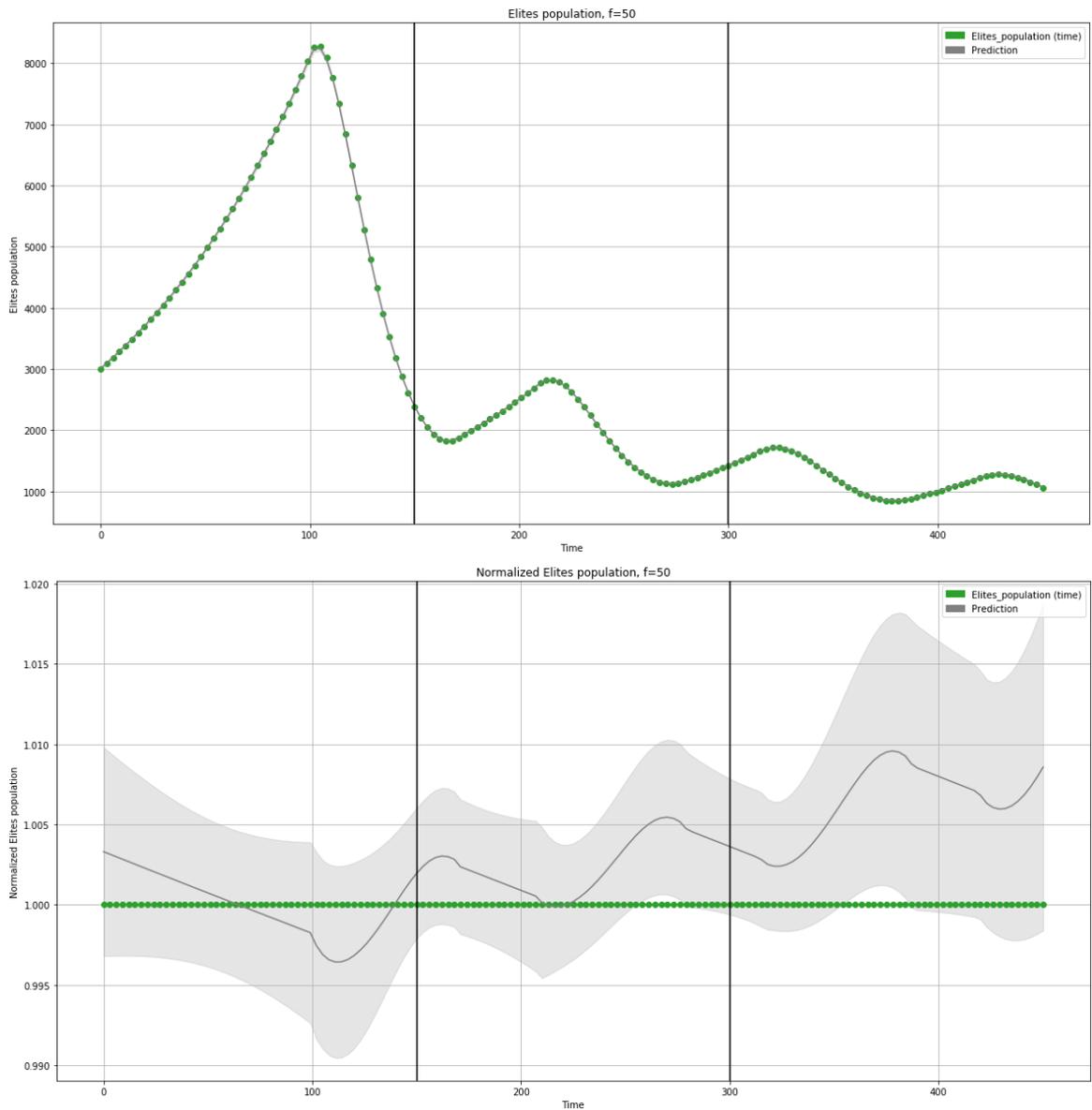

Fig. A.22 Supermodeling with similar submodels results - elites population (both real and normalized), $f = 50$

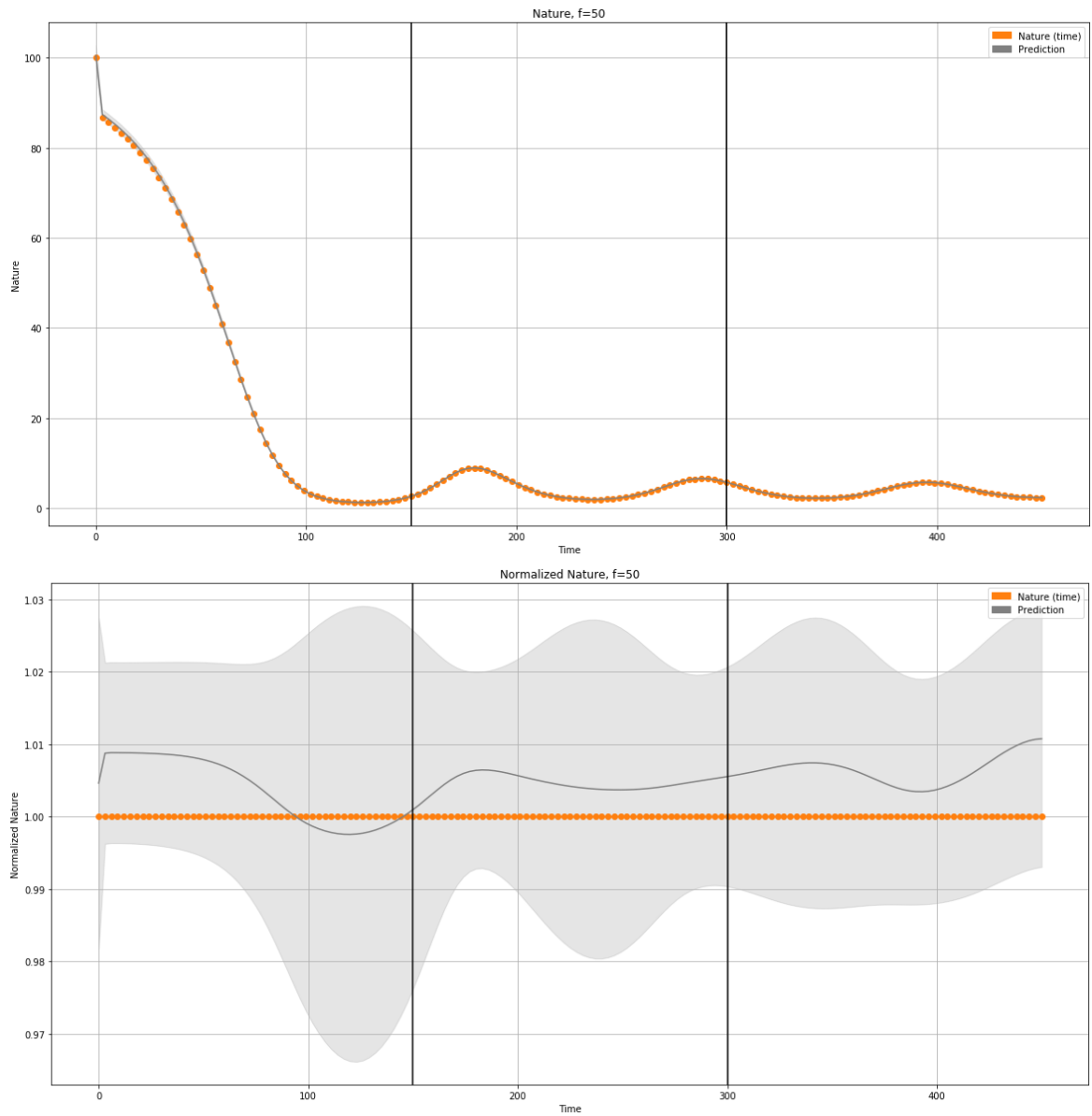

Fig. A.23 Supermodeling with similar submodels results - nature (both real and normalized), *f = 50*

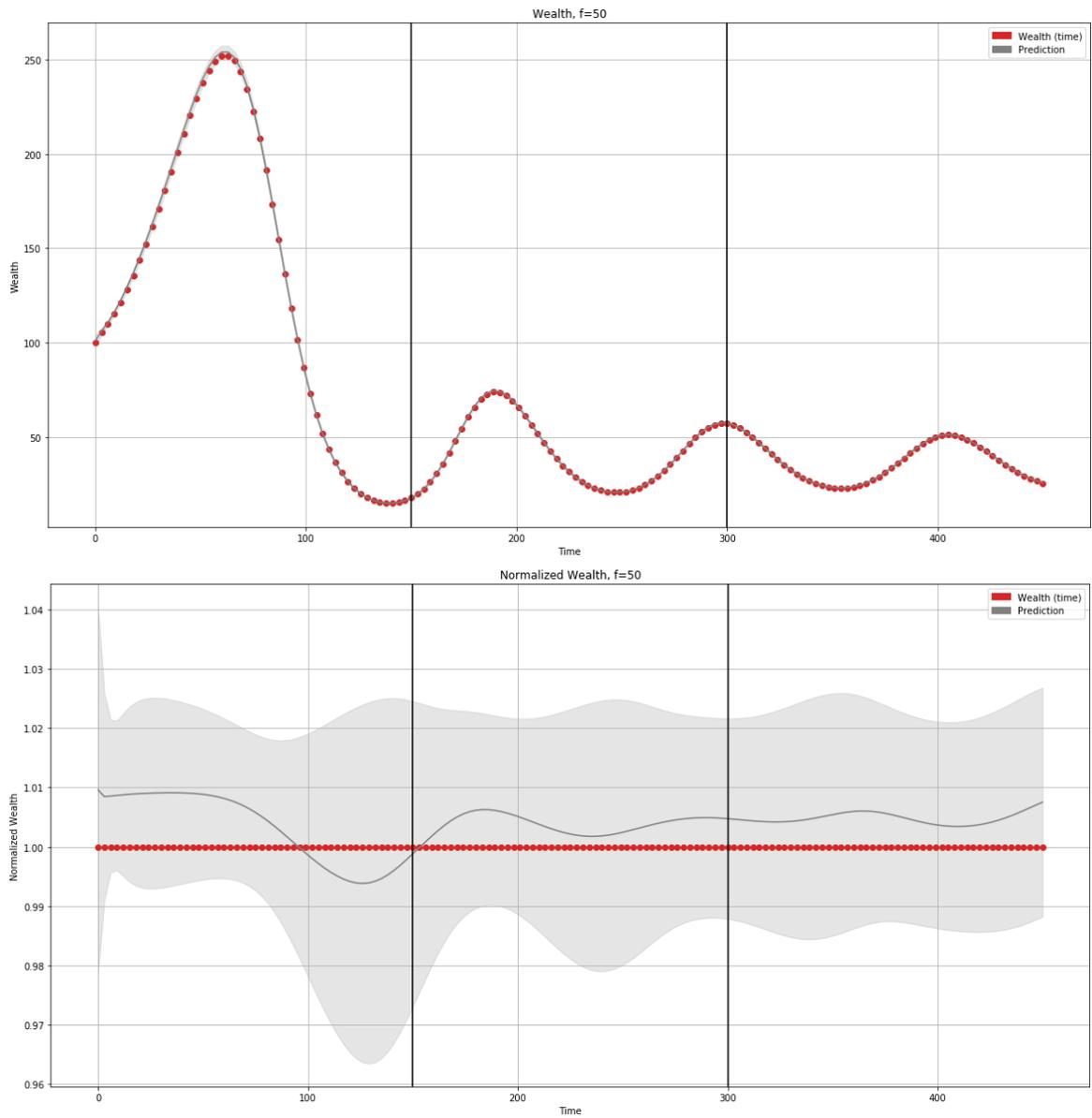

Fig. A.24 Supermodeling with similar submodels results - wealth (both real and normalized), *f = 50*

## A.4    Supermodeling with different submodels

### A.4.1    Sampling frequency *f = 15*

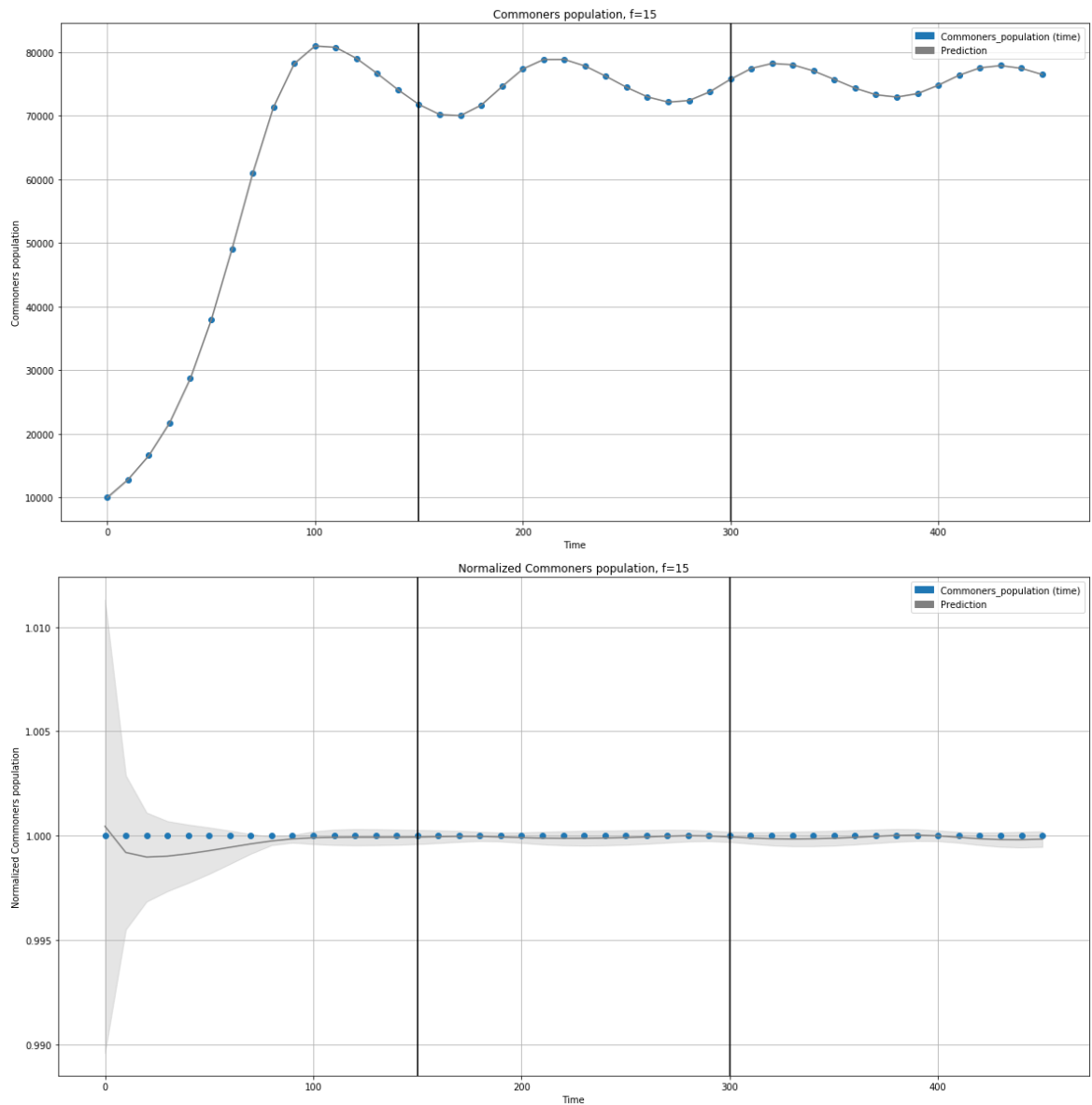

Fig. A.25 Supermodeling with different submodels results - commoners population (both real and normalized), *f = 15*

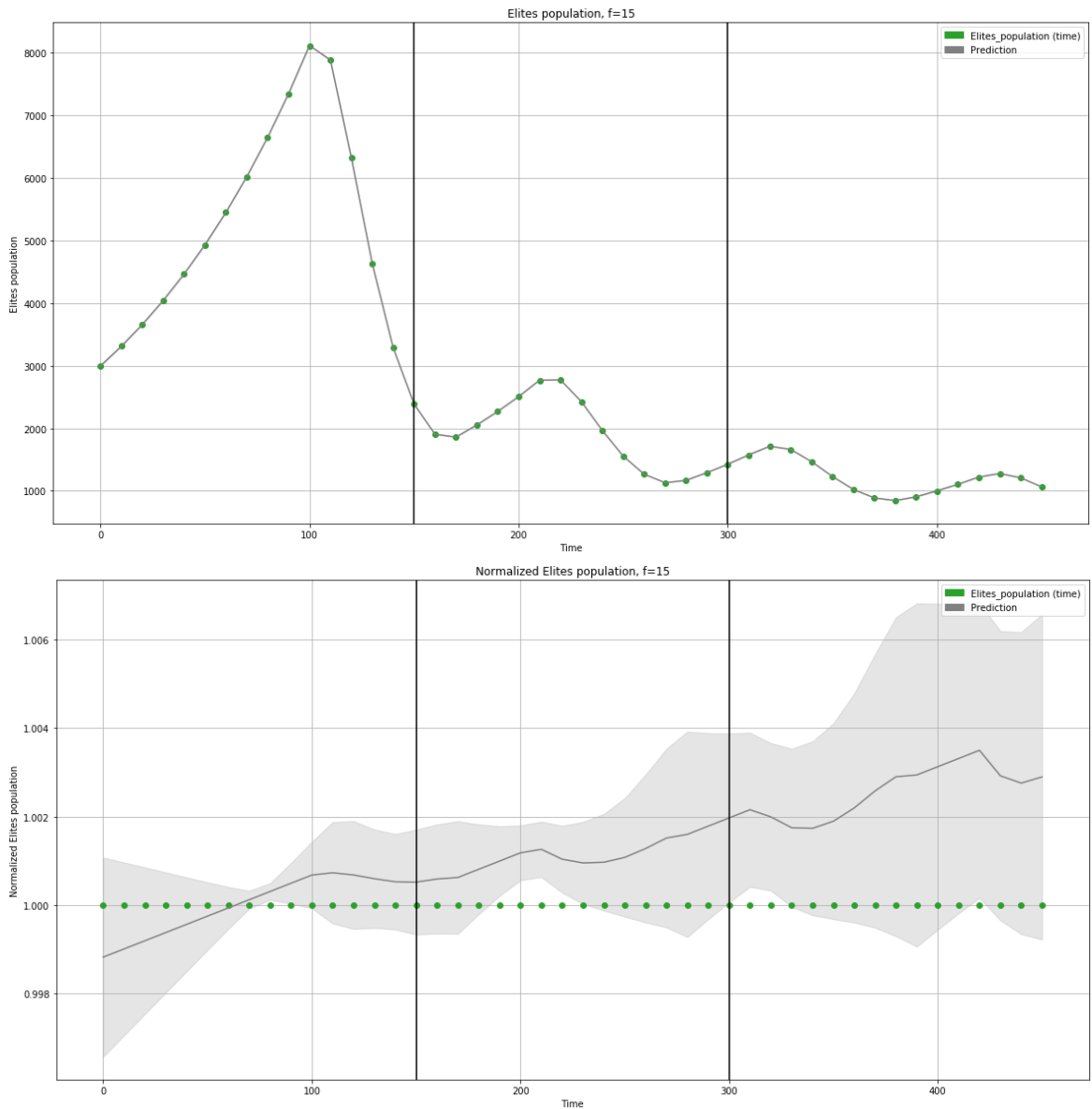

Fig. A.26 Supermodeling with different submodels results - elites population (both real and normalized), $f = 15$

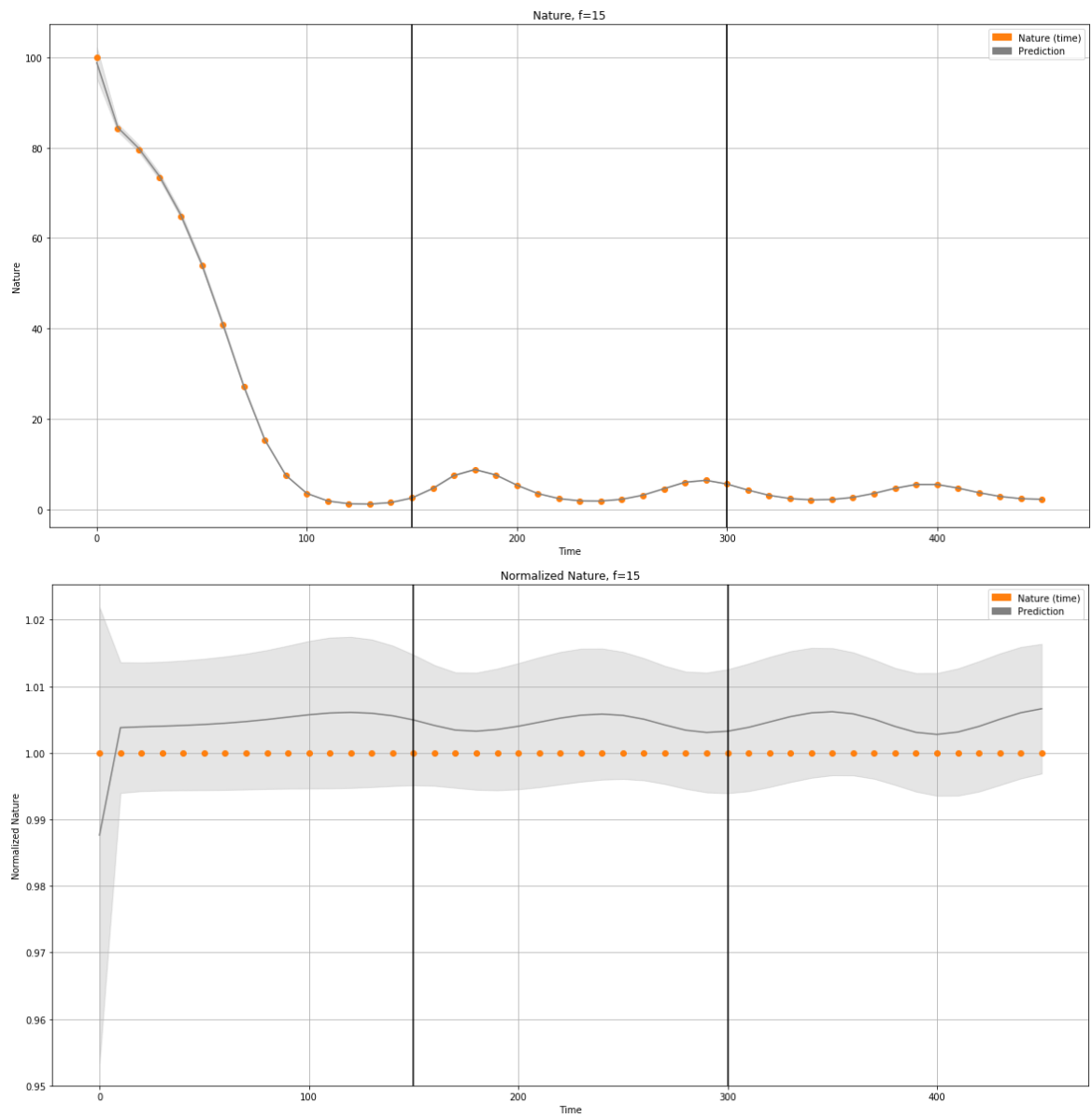

Fig. A.27 Supermodeling with different submodels results - nature (both real and normalized), *f = 15*

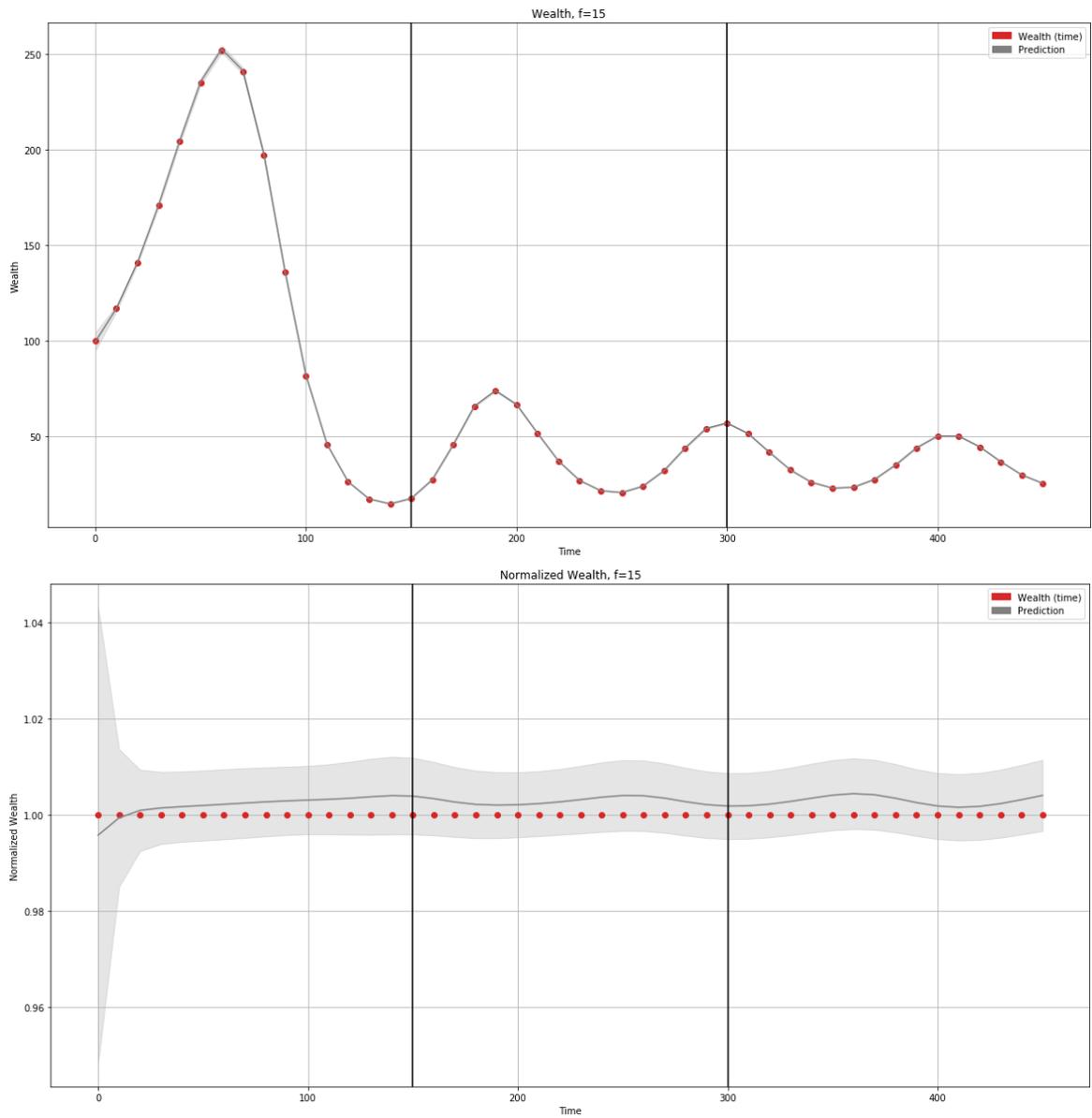

Fig. A.28 Supermodeling with different submodels results - wealth (both real and normalized), *f = 15*

## A.4.2  Sampling frequency *f = 50*

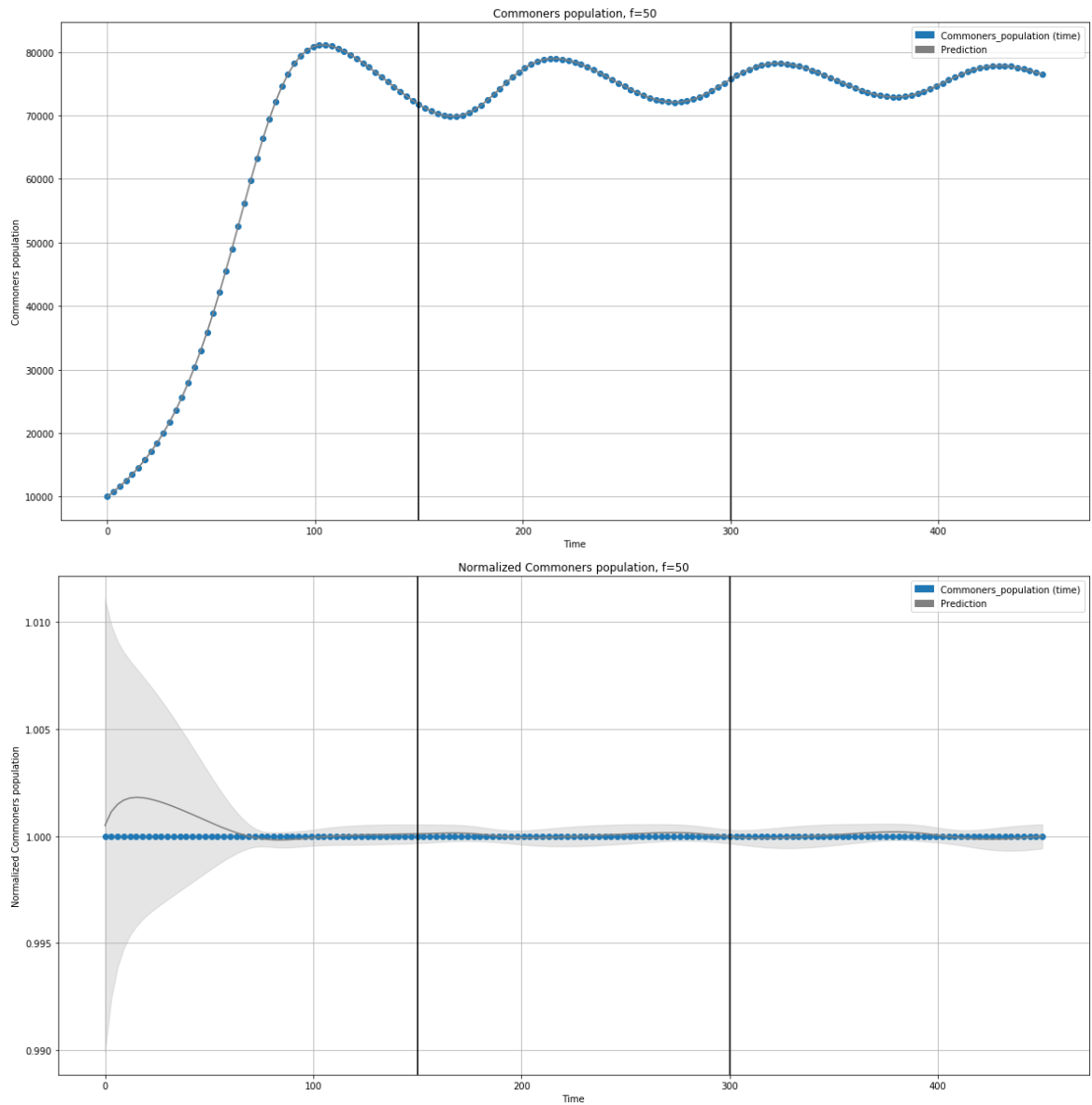

Fig. A.29 Supermodeling with different submodels results - commoners population (both real and normalized), *f = 50*

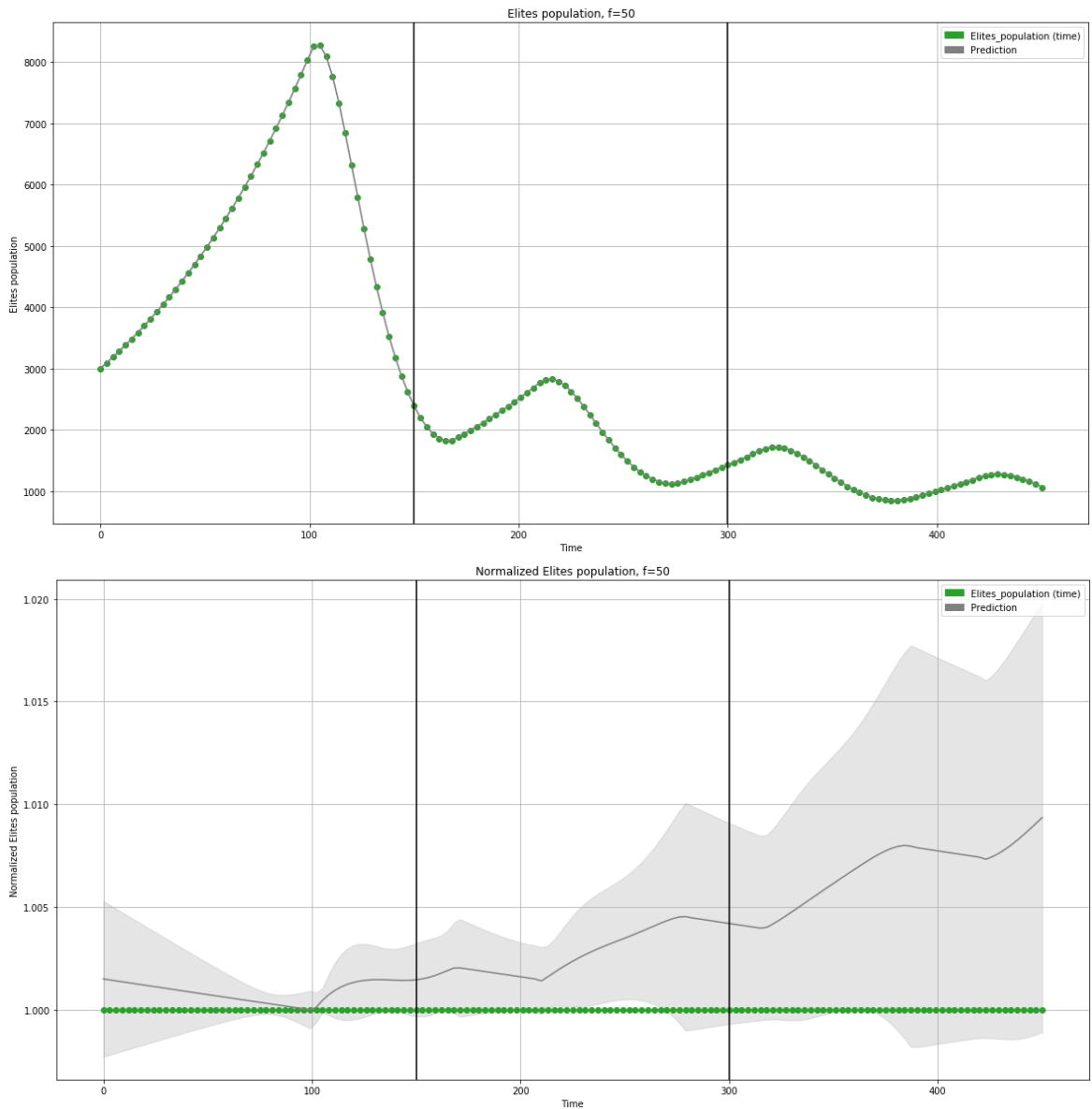

Fig. A.30 Supermodeling with different submodels results - elites population (both real and normalized), *f = 50*

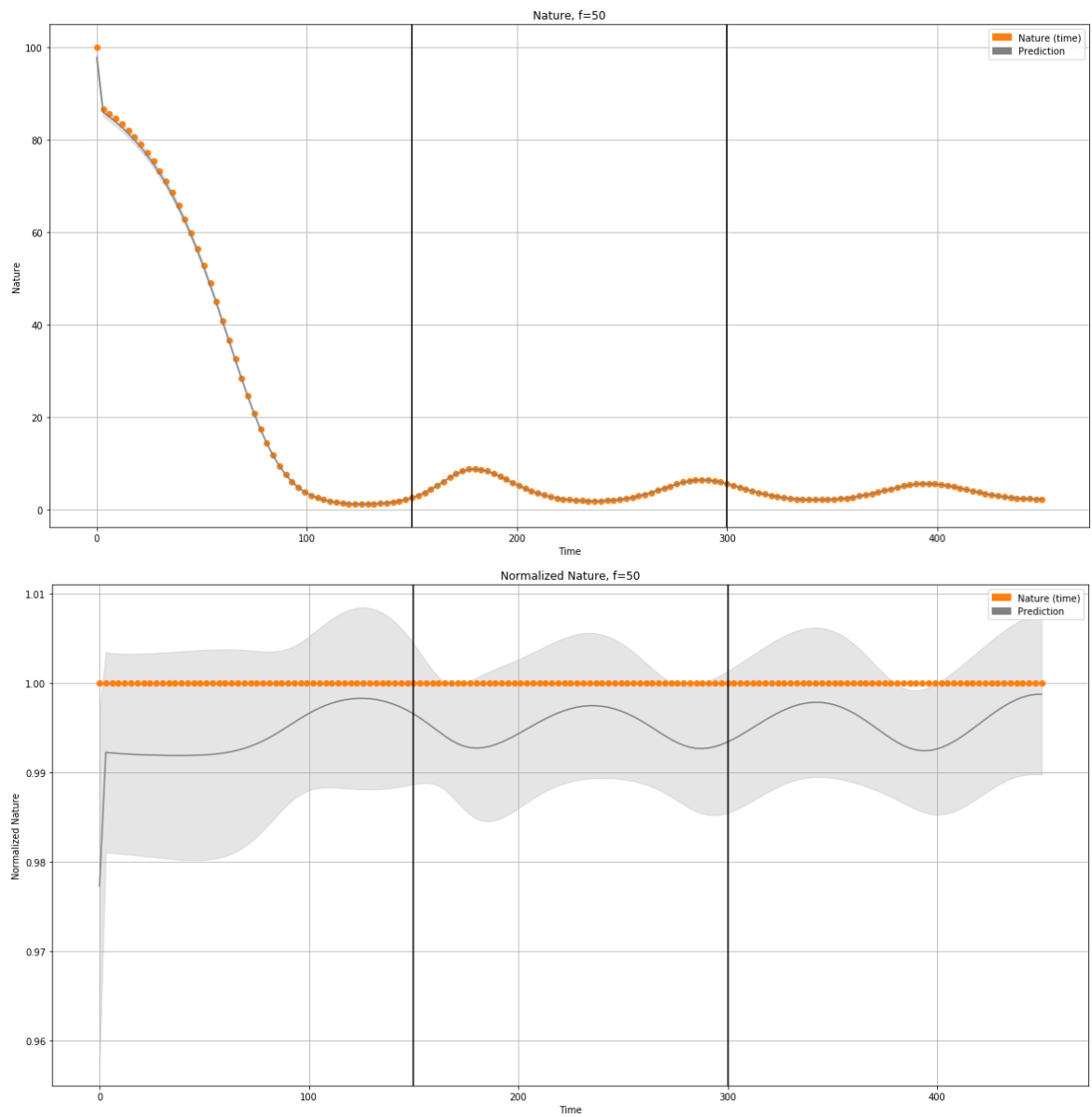

Fig. A.31 Supermodeling with different submodels results - nature (both real and normalized), *f = 50*

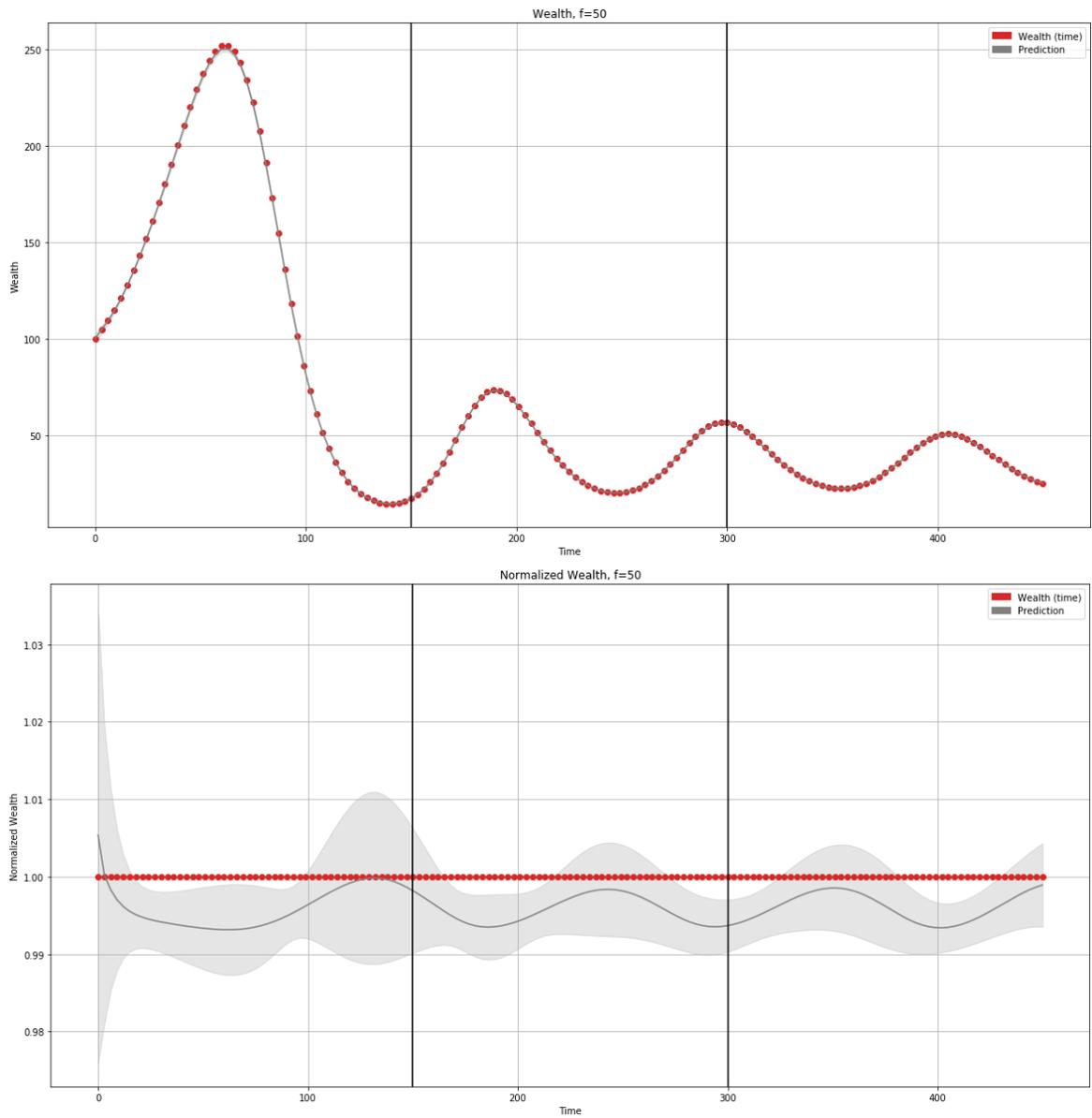

Fig. A.32 Supermodeling with different submodels results - wealth (both real and normalized), *f = 50*

# A.5 Supermodeling with Supermodel learned longer (600s)

## A.5.1 Sampling frequency *f = 15*

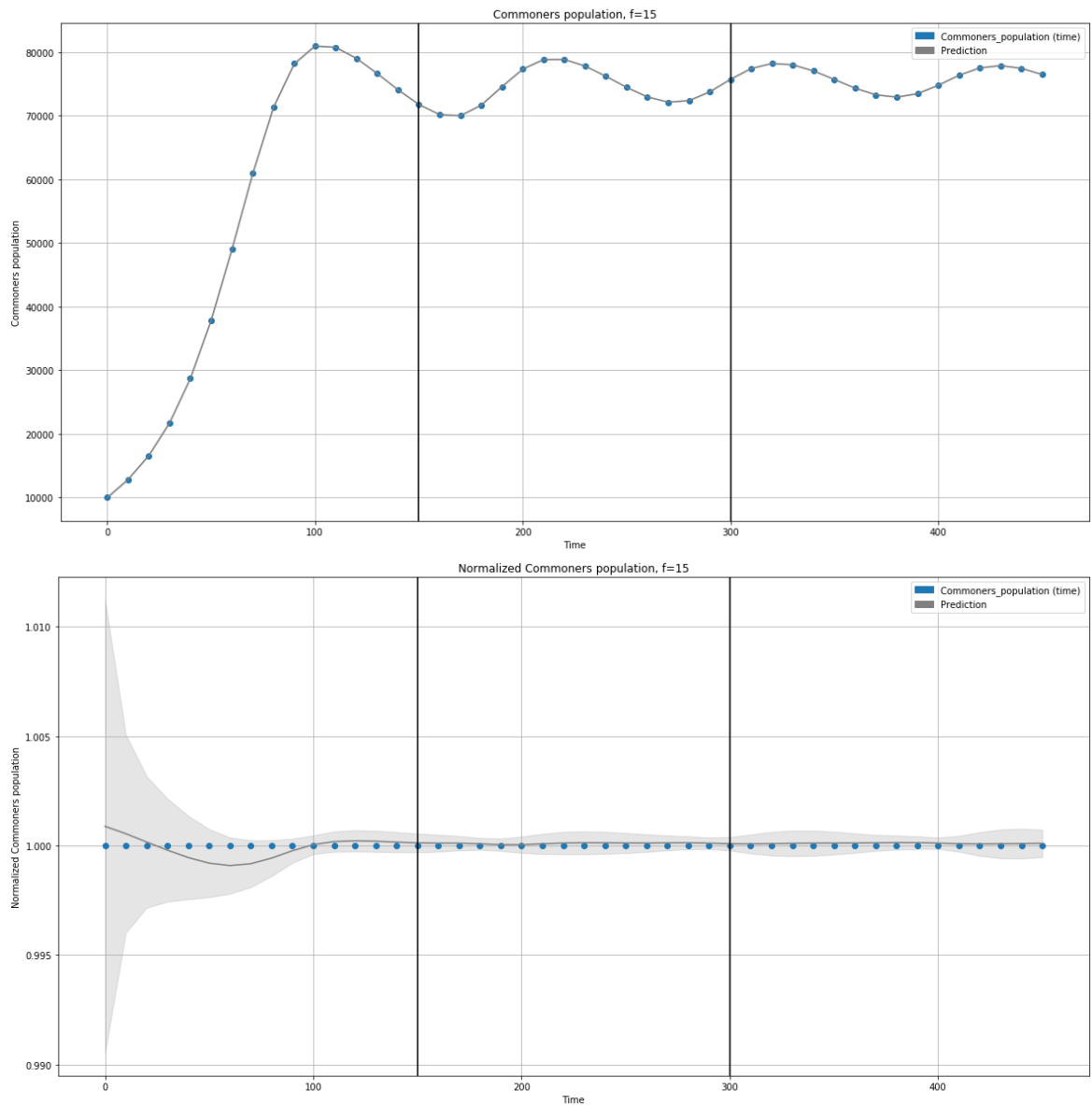

Fig. A.33 Supermodeling with Supermodel learned longer (600s) results - commoners population (both real and normalized), *f = 15*

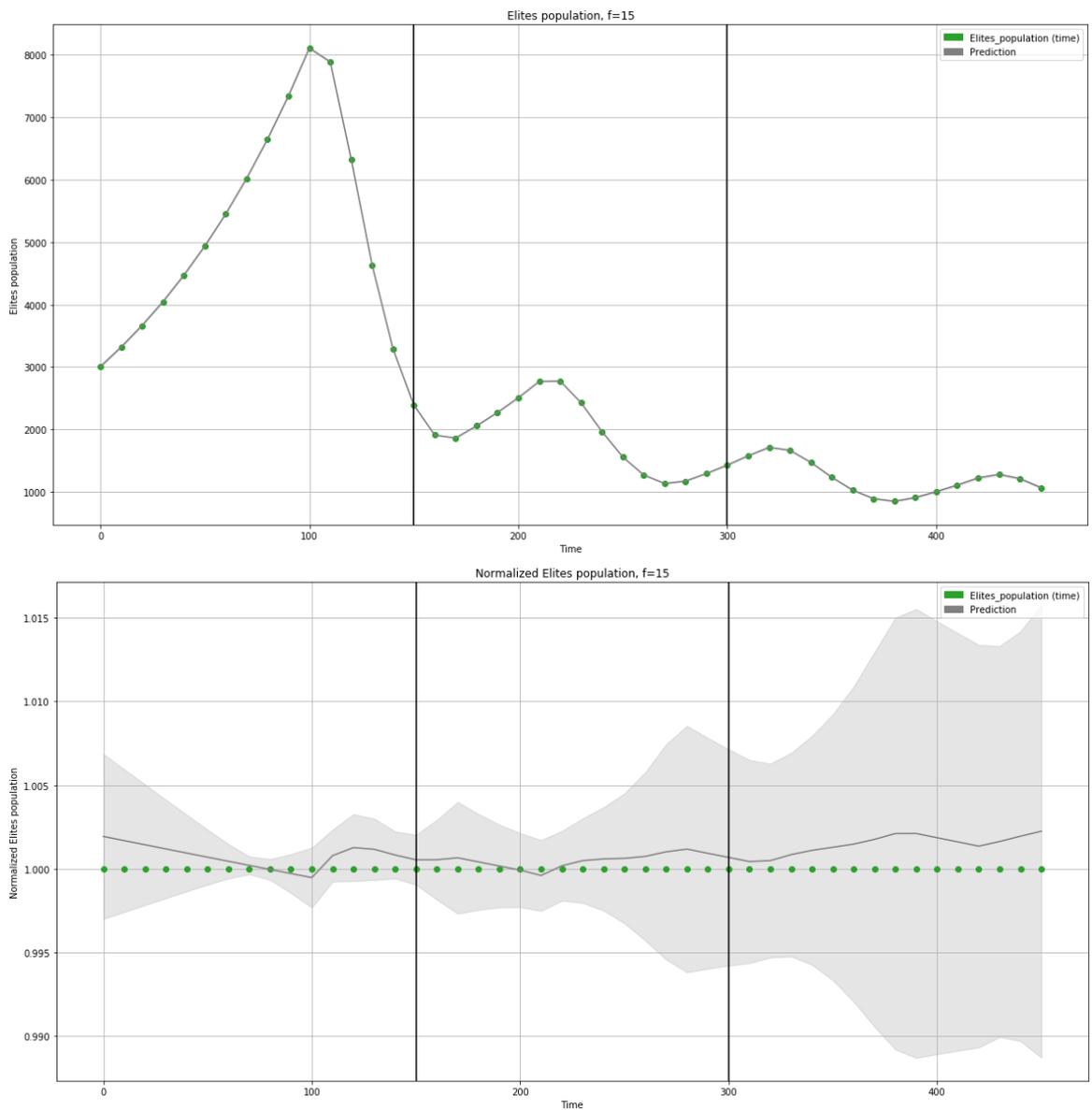

Fig. A.34 Supermodeling with Supermodel learned longer (600s) results - elites population (both real and normalized), *f = 15*

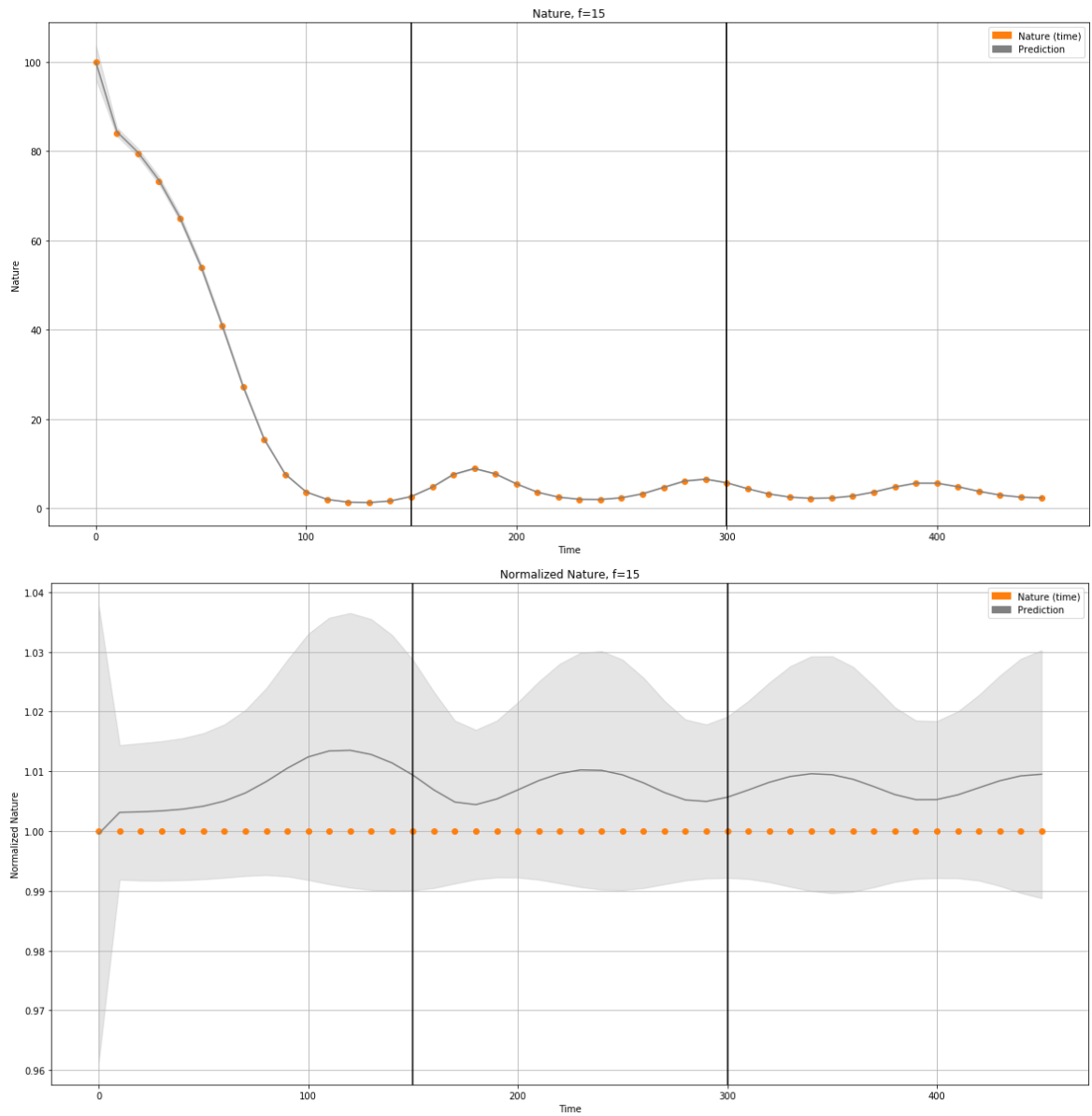

Fig. A.35 Supermodeling with Supermodel learned longer (600s) results - nature (both real and normalized), *f = 15*

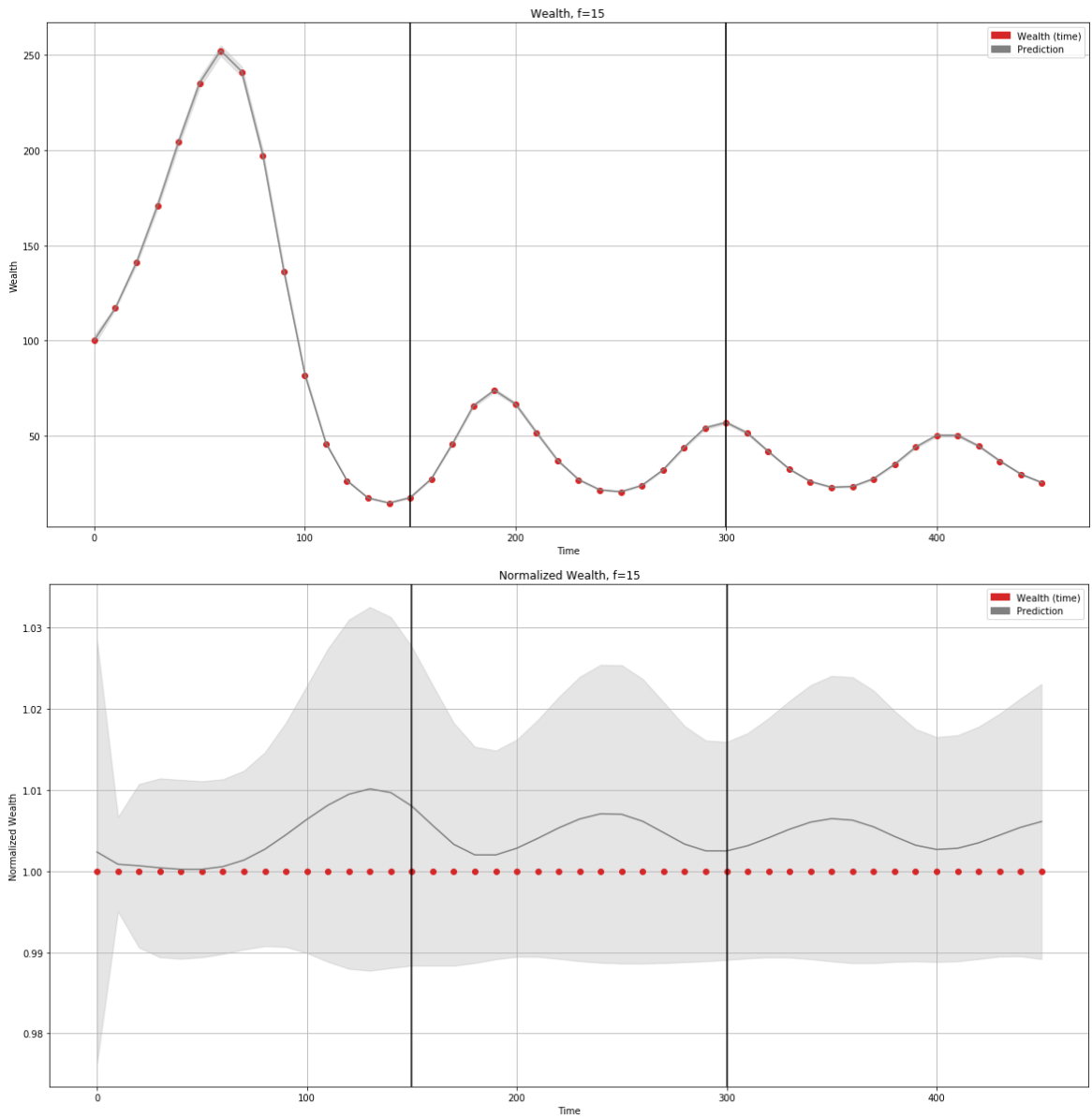

Fig. A.36 Supermodeling with Supermodel learned longer (600s) results - wealth (both real and normalized), *f = 15*

## A.5.2 Sampling frequency *f = 50*

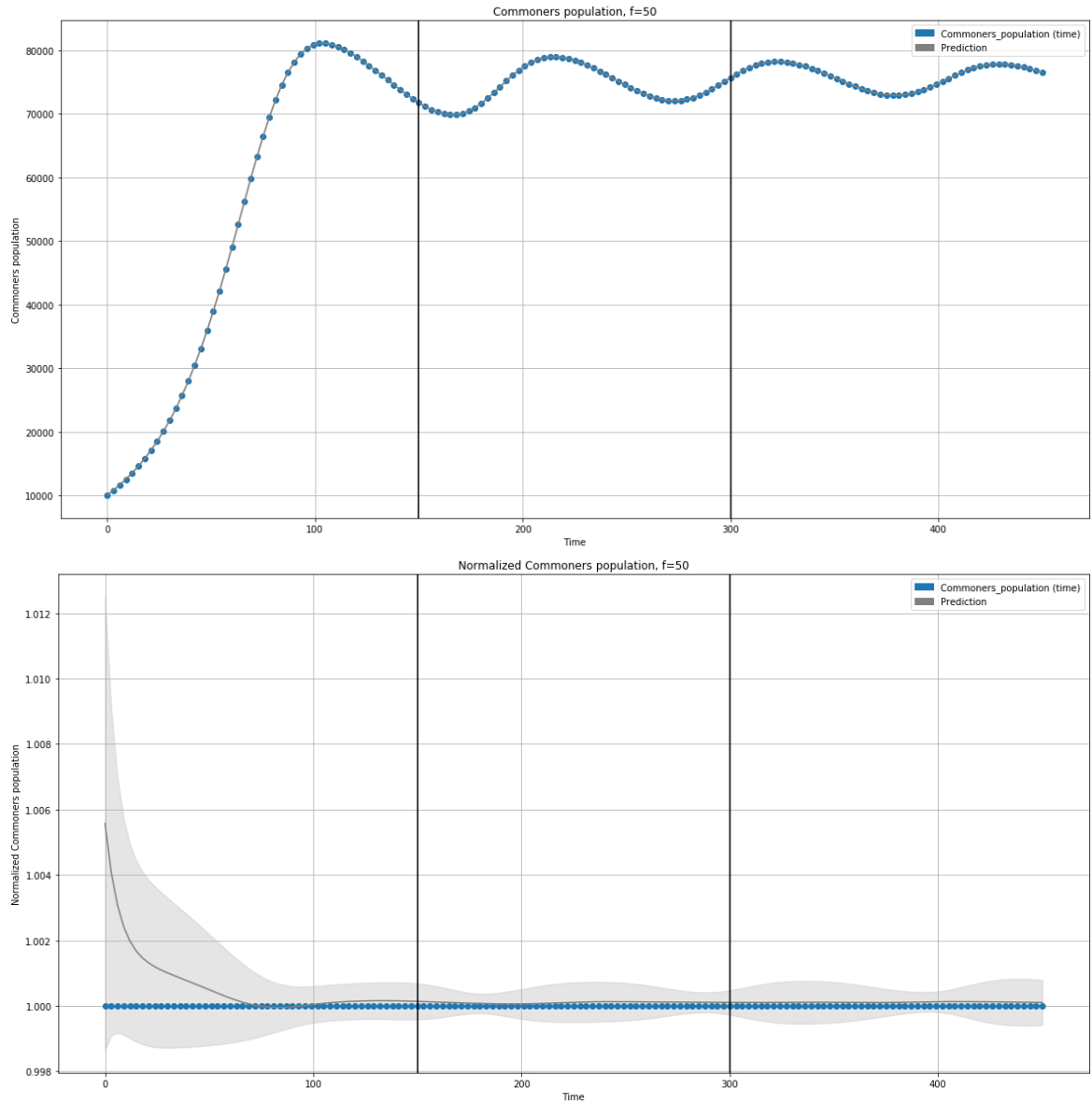

Fig. A.37 Supermodeling with Supermodel learned longer (600s) results - commoners population (both real and normalized), *f = 50*

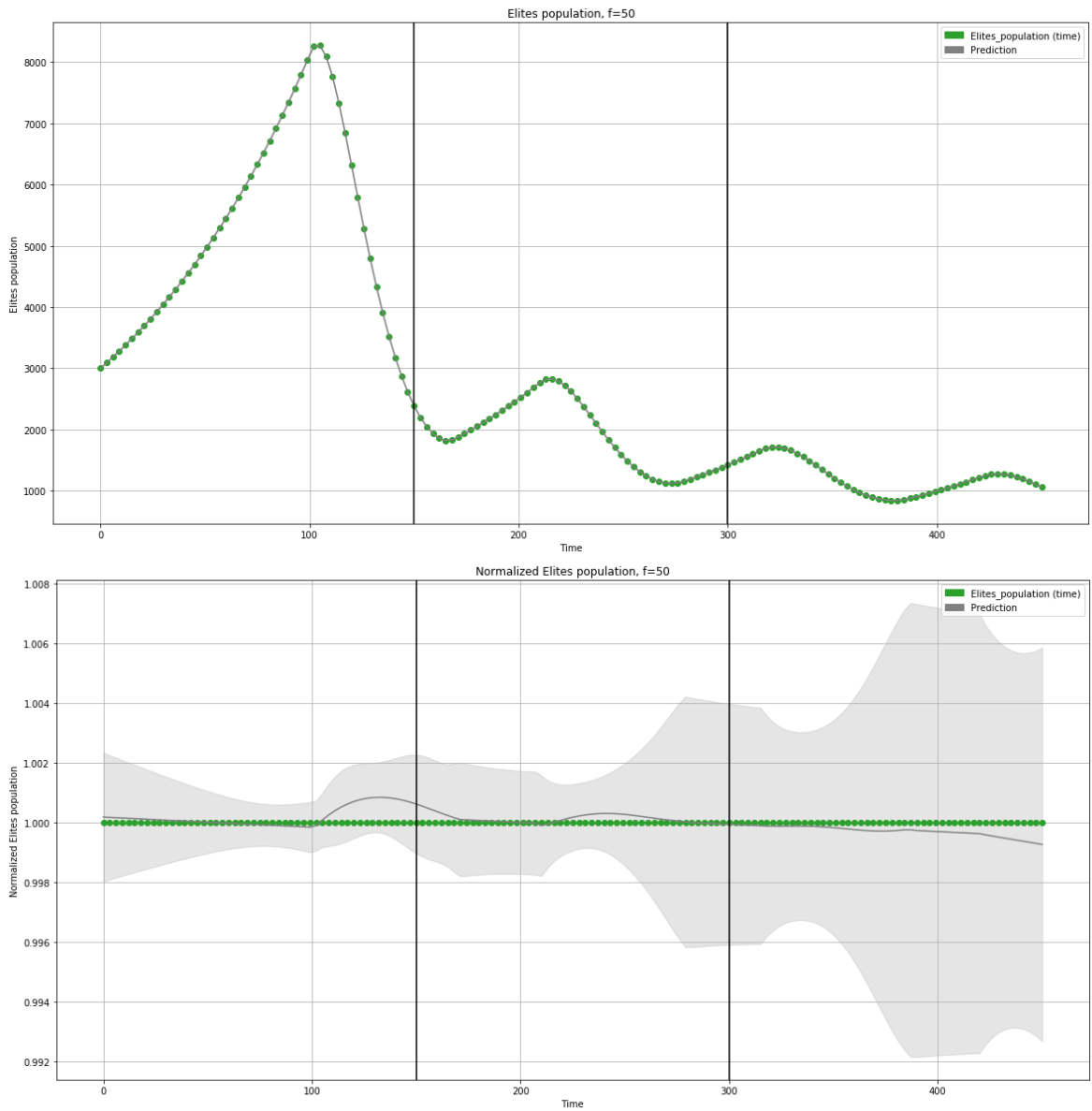

Fig. A.38 Supermodeling with Supermodel learned longer (600s) results - elites population (both real and normalized), *f = 50*

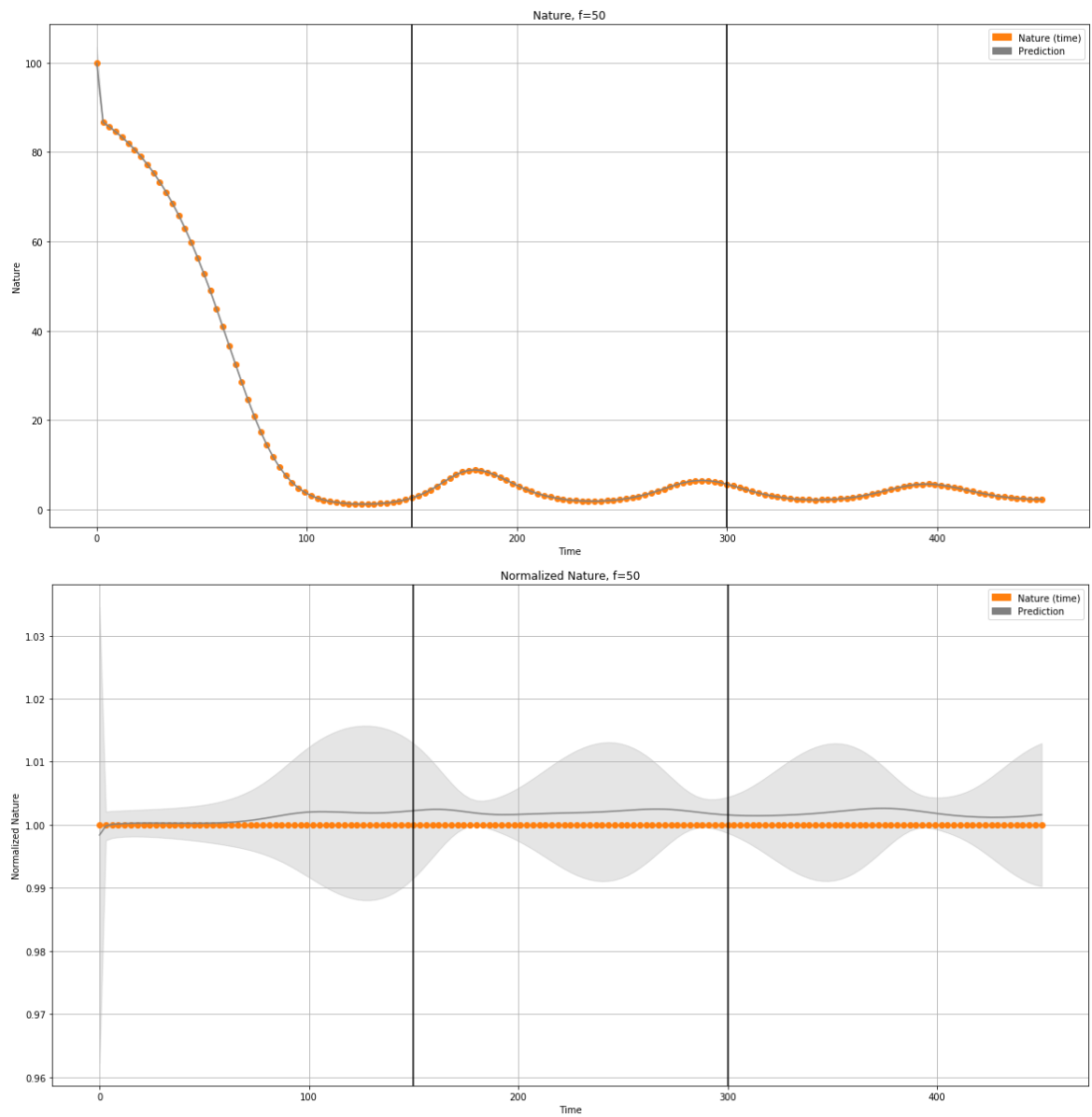

Fig. A.39 Supermodeling with Supermodel learned longer (600s) results - nature (both real and normalized), *f = 50*

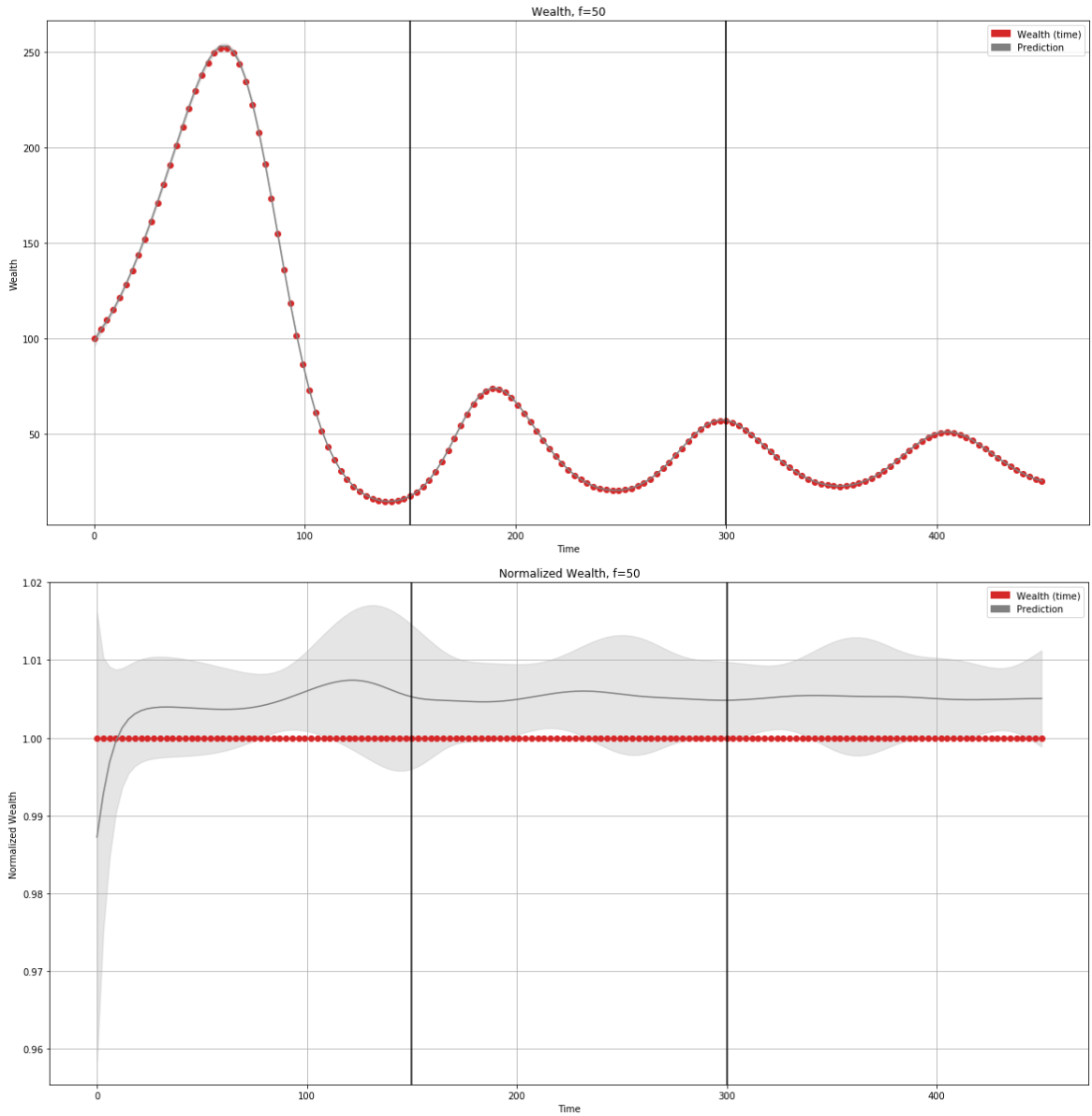

Fig. A.40 Supermodeling with Supermodel learned longer (600s) results - wealth (both real and normalized), *f = 50*

## A.6    Supermodeling with lower RMSE submodels

### A.6.1    Sampling frequency *f = 15*

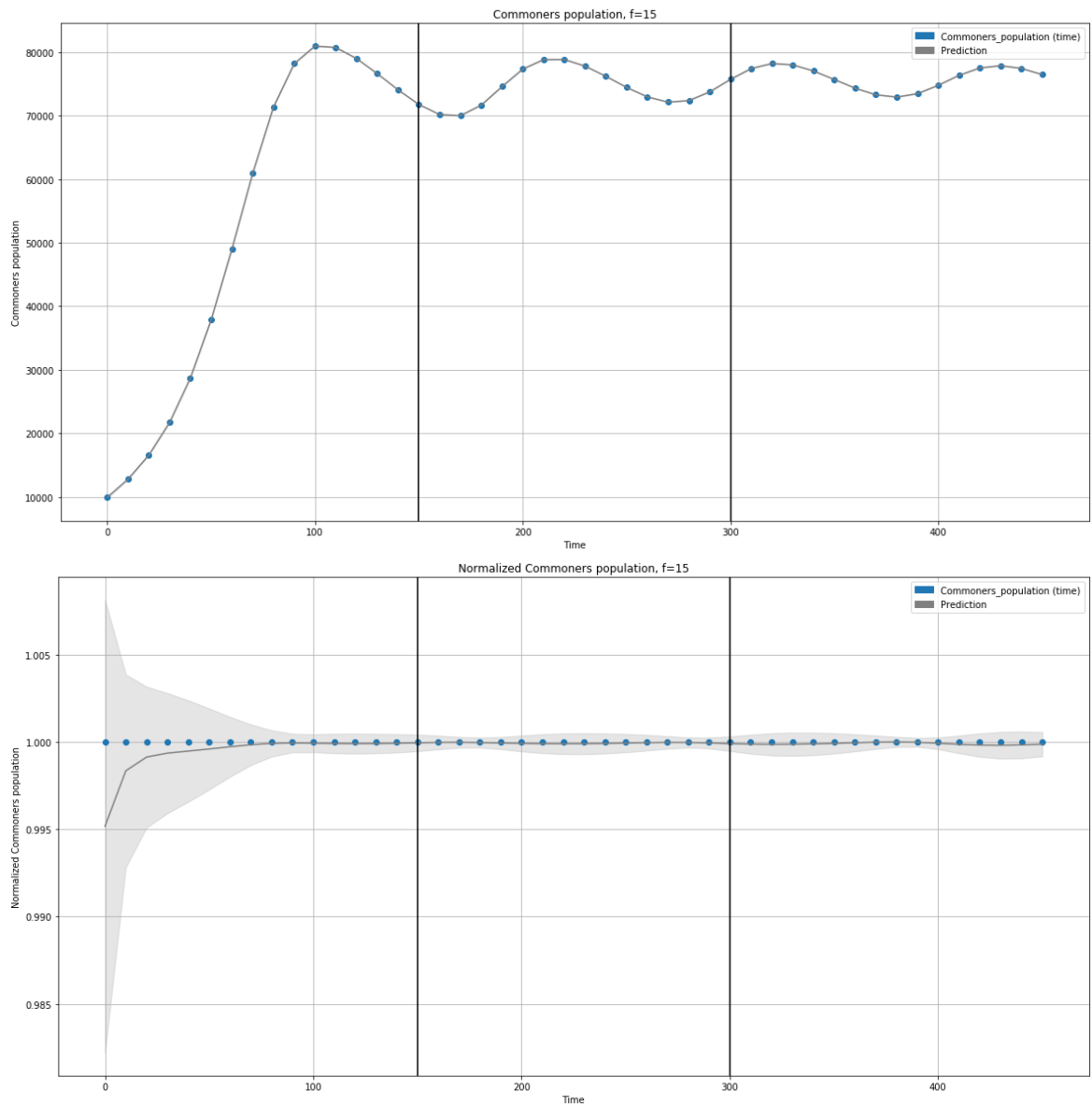

Fig. A.41 Supermodeling with with lower RMSE submodels results - commoners population (both real and normalized), *f = 15*

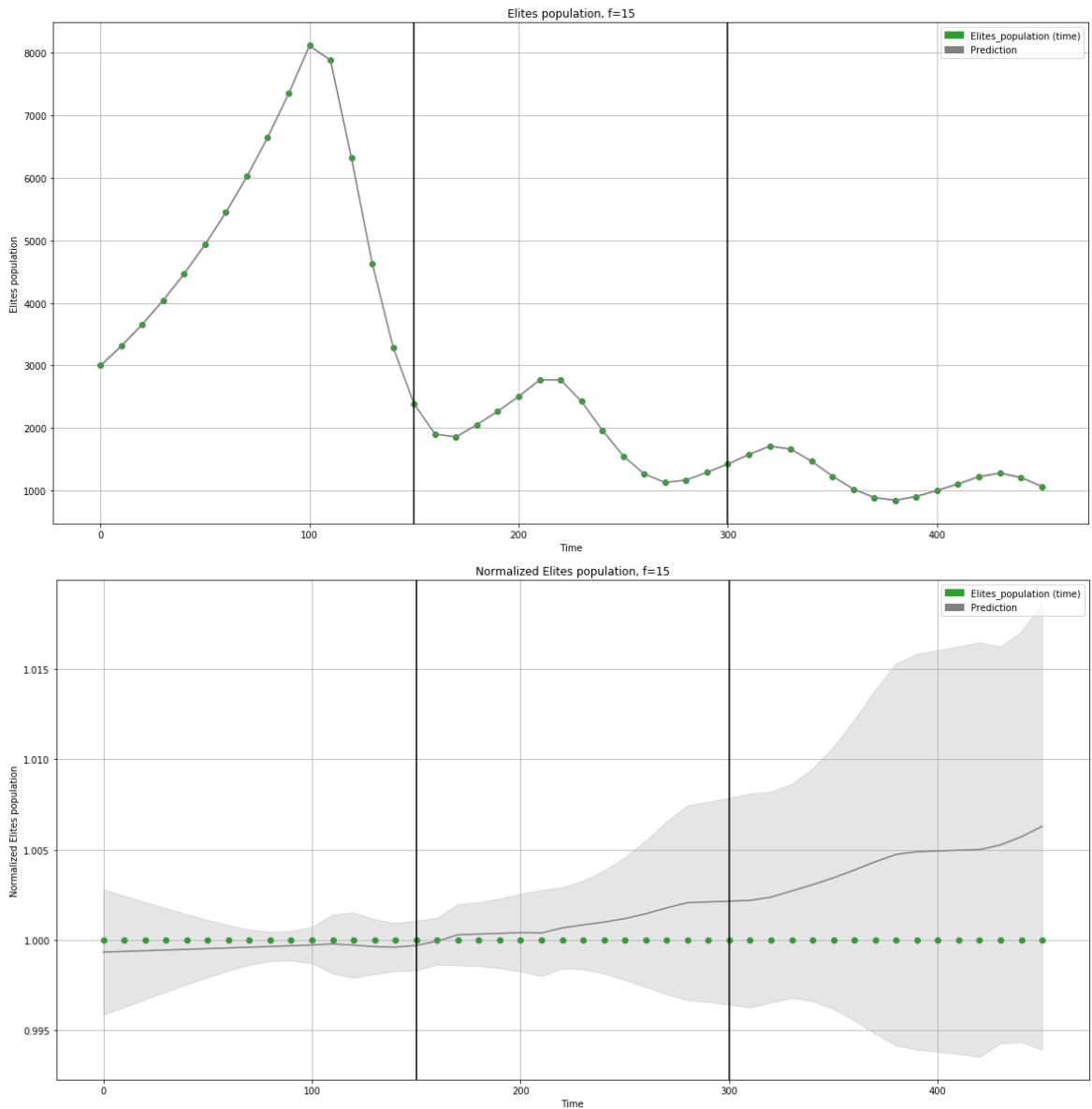

Fig. A.42 Supermodeling with with lower RMSE submodels results - elites population (both real and normalized), *f = 15*

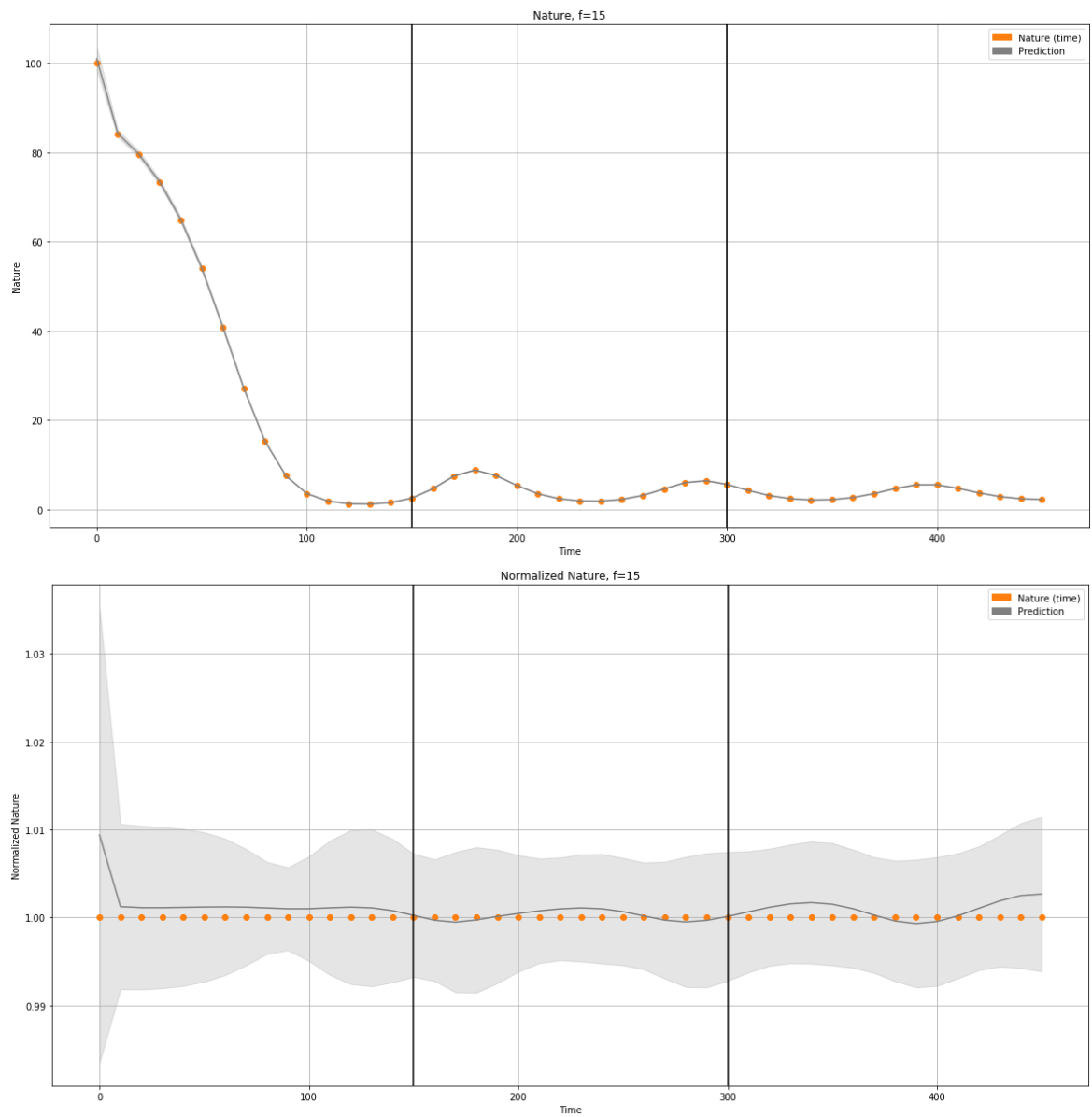

Fig. A.43 Supermodeling with with lower RMSE submodels results - nature (both real and normalized), *f = 15*

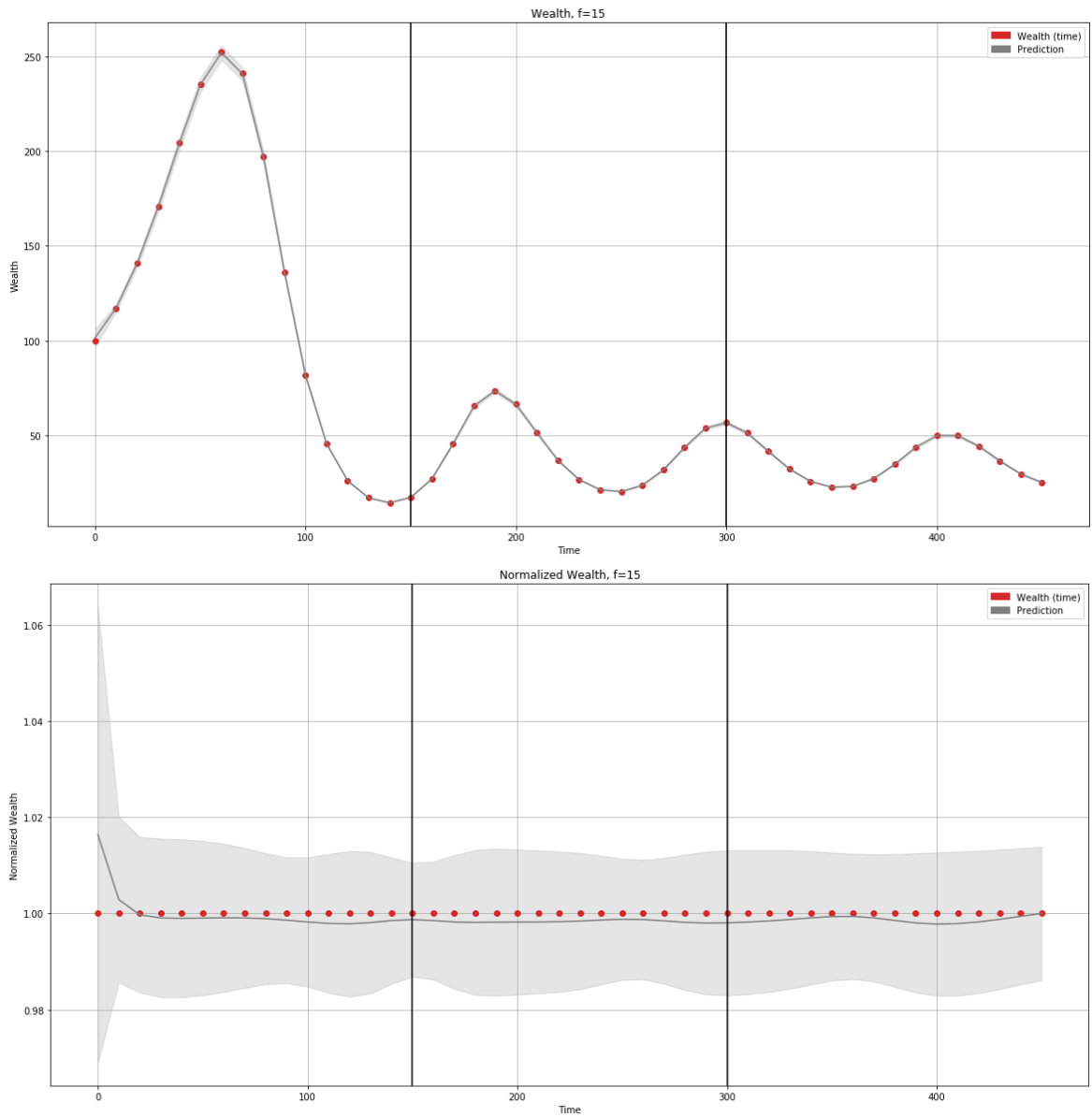

Fig. A.44 Supermodeling with with lower RMSE submodels results - wealth (both real and normalized), $f = 15$

## A.6.2    Sampling frequency *f = 50*

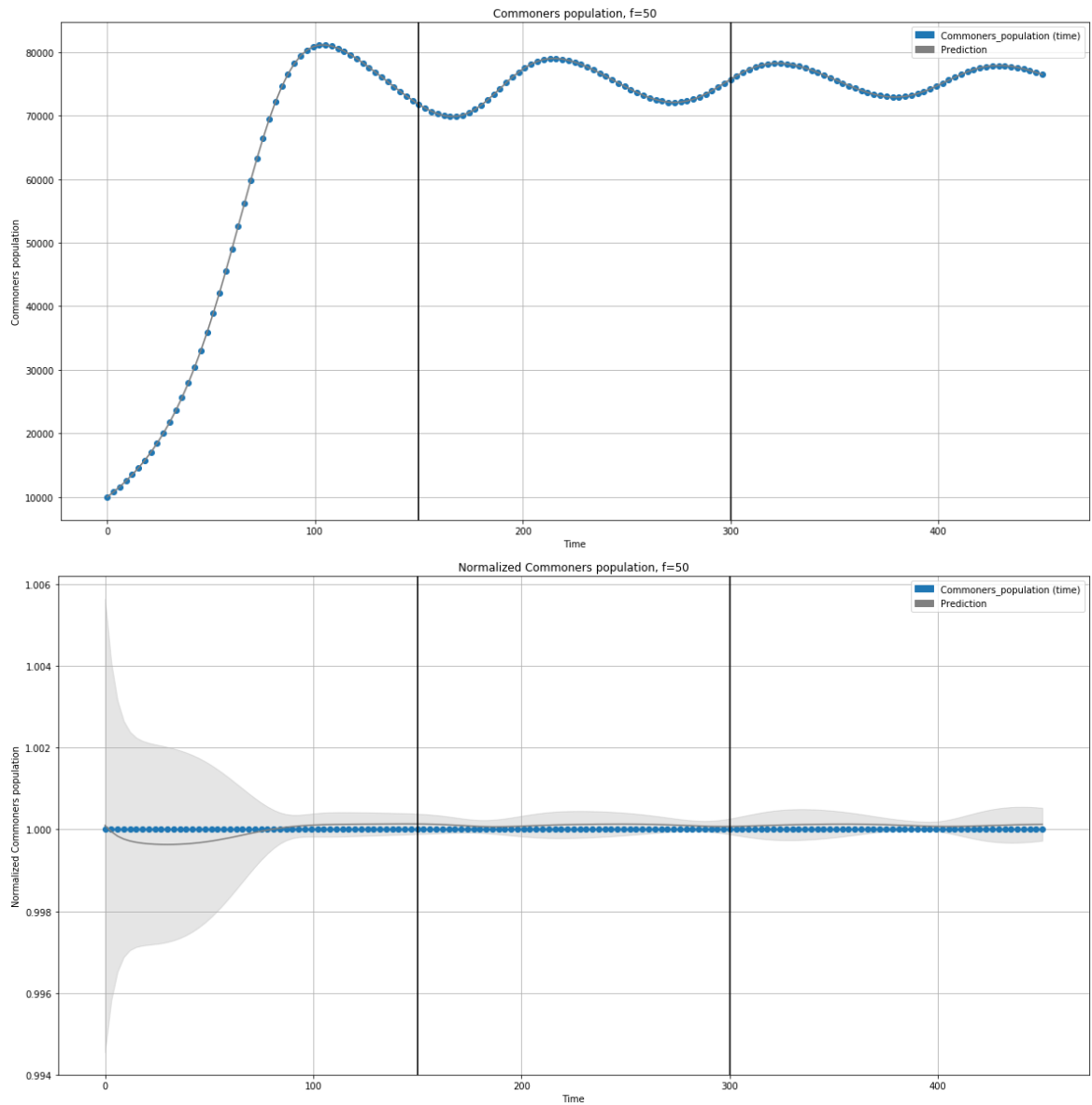

Fig. A.45 Supermodeling with with lower RMSE submodels results - commoners population (both real and normalized), *f = 50*

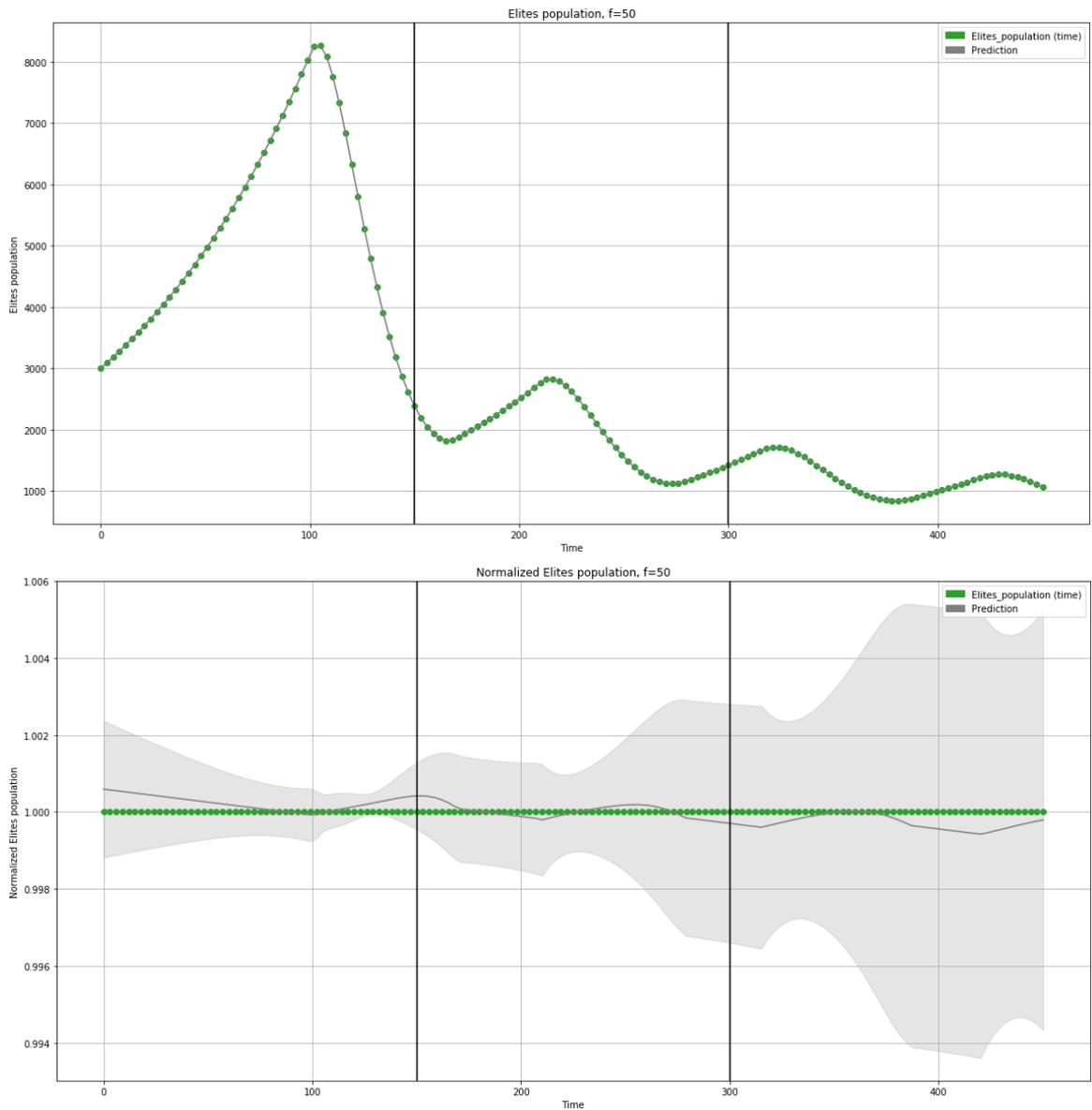

Fig. A.46 Supermodeling with with lower RMSE submodels results - elites population (both real and normalized), *f = 50*

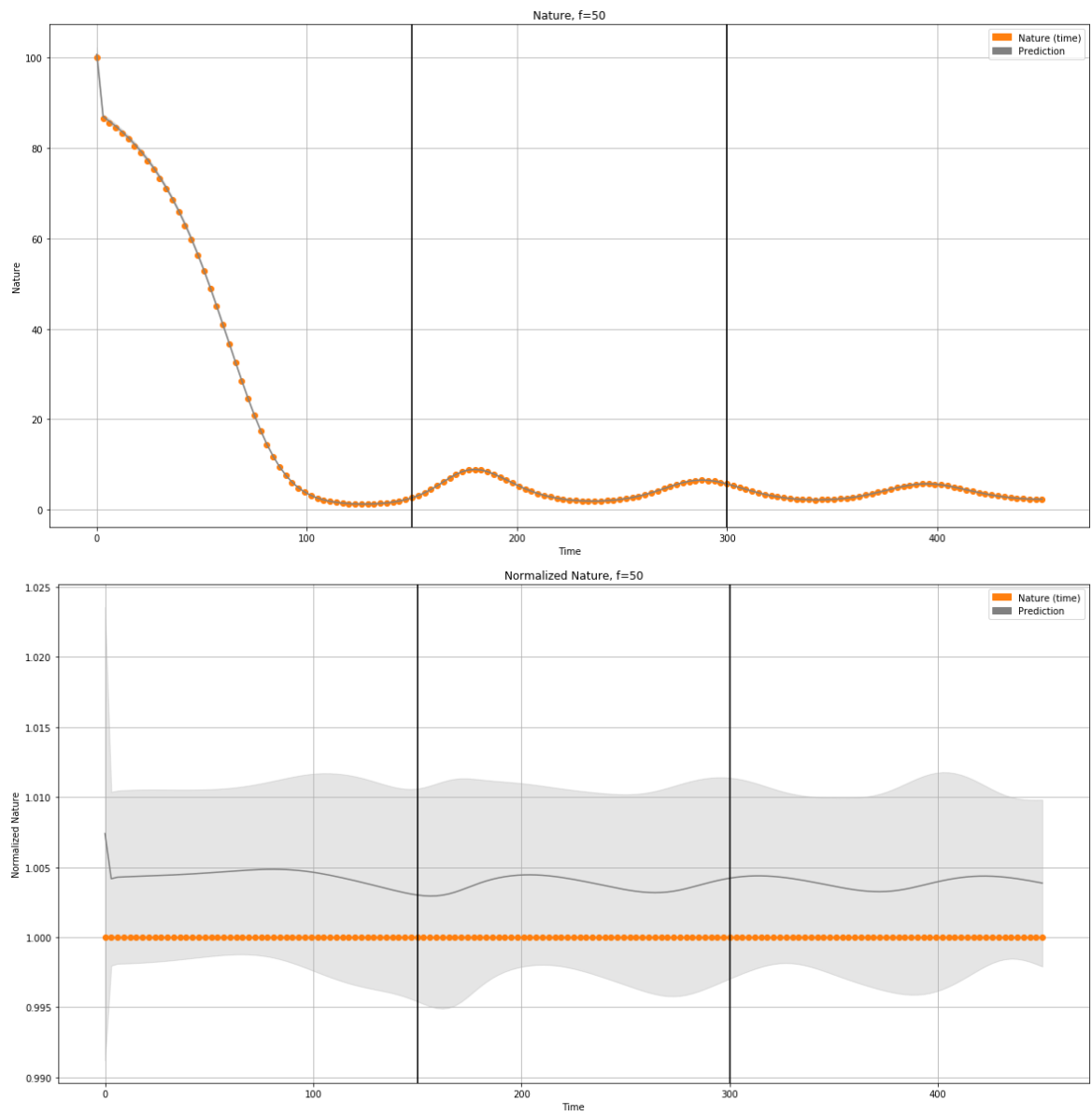

Fig. A.47 Supermodeling with with lower RMSE submodels results - nature (both real and normalized), *f = 50*

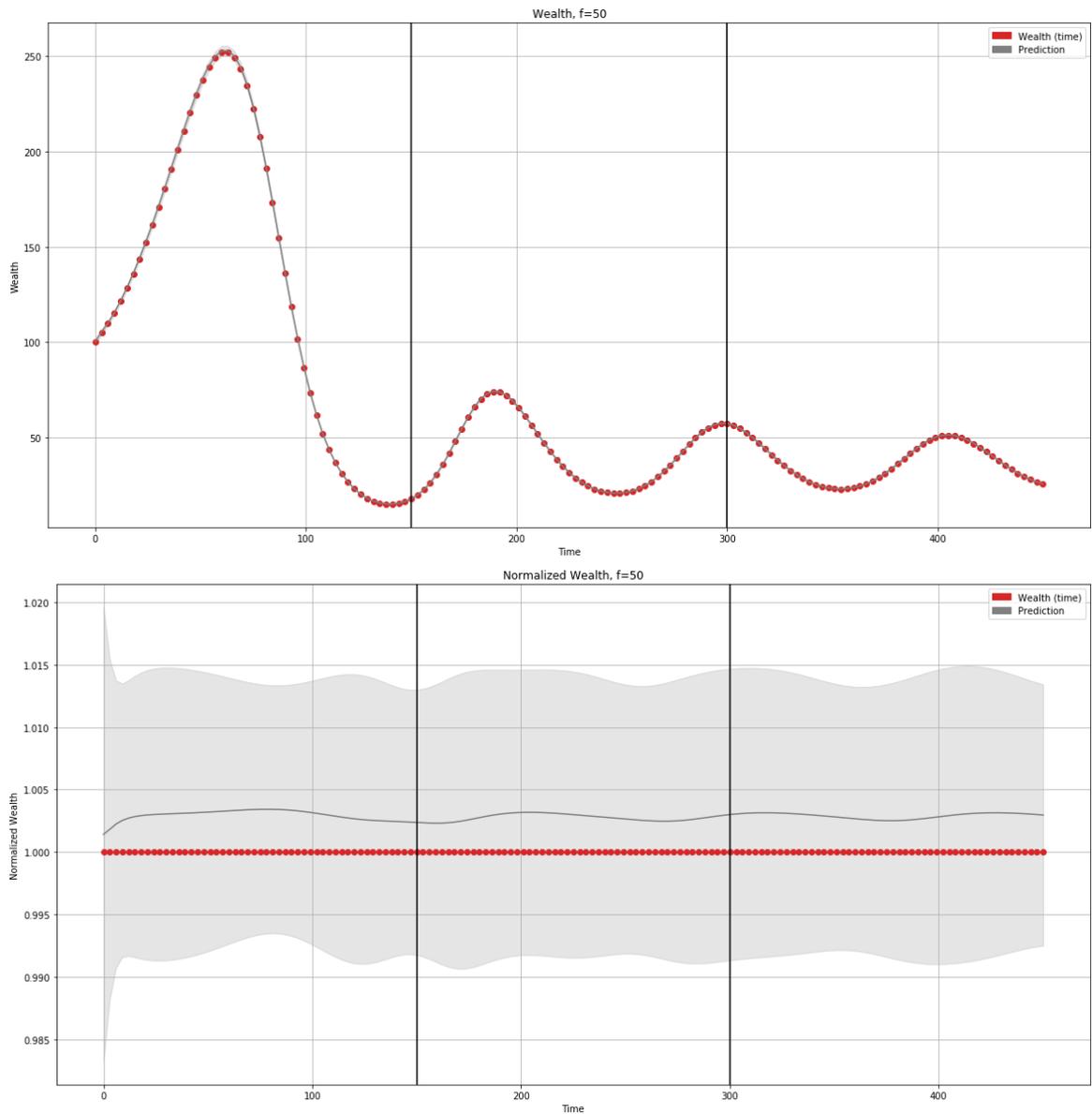

Fig. A.48 Supermodeling with with lower RMSE submodels results - wealth (both real and normalized), *f = 50*